\documentclass[11pt,a4paper]{article}

\usepackage[T1]{fontenc}
\usepackage[utf8]{inputenc}
\usepackage{lmodern}
\usepackage[a4paper,margin=2.5cm]{geometry}
\usepackage{microtype}
\usepackage{amsmath,amssymb,amsfonts,bm}

\usepackage[round,authoryear]{natbib}
\setcitestyle{citesep={;},aysep={,},yysep={;}}

\usepackage{xcolor}
\usepackage{url}
\usepackage{multirow}
\usepackage{booktabs}
\usepackage{makecell}
\usepackage{tabularx}
\usepackage{graphicx}
\usepackage{subcaption}
\usepackage{float}
\usepackage{comment}
\usepackage{xspace}
\usepackage[colorlinks=true,linkcolor=blue,citecolor=blue,urlcolor=blue]{hyperref}

\definecolor{draftblue}{RGB}{0,70,190}

\newcommand{\encNone}{w/o covariates\xspace}
\newcommand{\encOrdinal}{\texorpdfstring{\texttt{ordinal\textsubscript{h,W} + bin\textsubscript{H}}\xspace}{ordinal_h,W + bin_H}}
\newcommand{\encOnehot}{\texorpdfstring{\texttt{one-hot\textsubscript{h,W} + bin\textsubscript{H}}\xspace}{one-hot_h,W + bin_H}}
\newcommand{\encCyclic}{\texorpdfstring{\texttt{cyclic\textsubscript{h,W} + bin\textsubscript{H}}\xspace}{cyclic_h,W + bin_H}}
\newcommand{\encTcombo}{\texorpdfstring{$T_{h,W}(H=\mathrm{Sun})$\xspace}{T_h,W(H=Sun)}}
\newcommand{\encTholEight}{\texorpdfstring{$T_{h,W}(H=7)$\xspace}{T_h,W(H=7)}}
\newcommand{\encTholAll}{\texorpdfstring{$T_{h,W,H}$\xspace}{T_h,W,H}}
\newcommand{\encTholPartial}{\texorpdfstring{$T_{h,W} + T_{h,H}$\xspace}{T_h,W + T_h,H}}

\newcommand{\encThW}{\texorpdfstring{$T_{h,W}$\xspace}{T_h,W}}
\newcommand{\encThH}{\texorpdfstring{$T_{h,H}$\xspace}{T_h,H}}

\title{On the Encoding of Calendar Features in Covariate-Informed Time Series Foundation Models}

\author{Alexander Kreusel\thanks{Equal contribution.}, Maximilian Beichter\footnotemark[1], Alexandra Nikoltchovska, \\
Benjamin Schäfer, Ralf Mikut, Frederik Zahn, Veit Hagenmeyer \\
Institute for Automation and Applied Informatics \\
Karlsruhe Institute of Technology \\
Karlsruhe, Germany \\
\texttt{\{firstname.lastname\}@kit.edu}
}
\date{}

\begin{document}

\maketitle

\begin{abstract}
Time series foundation models (TSFMs) increasingly offer covariate support, with calendar information being the most common type of categorical covariate in practice, as it is available for every point in time. However, how these features are encoded is largely treated as an implementation detail rather than
as an object of study, despite encoding being one of the few design choices left for pretrained models at inference time. With the present study, we demonstrate that the way information is propagated into the model substantially affects forecast accuracy of TSFMs.
Therefore, we systematically compare seven calendar
encodings, including three conventional target-agnostic representations (ordinal, one-hot, and
cyclic) and four target encodings, across four state-of-the-art TSFMs (Chronos-2, TabPFN-TS-3, TiRex-2, TimesFM-3)\footnote{Results of $t_0$-beta \cite{meyer2026t0} are in preparation.} on nine
datasets from multiple domains in a zero-shot rolling evaluation. The nine datasets are influenced by social activities, such as weekends and holidays, making calendar information particularly relevant for forecasting. We find that encoding choice affects forecast accuracy across all models, with MASE differences of up to 9.9 \% on non-holidays and 83 \% on holidays within a single model. Effectiveness and utilization vary substantially across models. Improvements are concentrated around holidays, where calendar information provides signals unavailable from target history, while some model encoding combinations also improve non-holiday forecasts.
A SHAP-based attribution analysis locates the contribution of the covariates relative to the autoregressive component and explains why some model–encoding combinations degrade performance. Our findings suggest that current benchmarks for covariate-informed TSFMs can distort model comparisons by imposing fixed encoding strategies. Future benchmarks should therefore evaluate multiple encodings or provide model-specific encoding strategies aligned with each model's best practices. 
\end{abstract}

\section{Introduction}
Accurate time series forecasts are essential for downstream decisions in energy, transport, retail, finance, and other domains, where forecast errors directly affect operational costs. While forecasting has traditionally relied on dataset-specific models that require sufficient historical data and computing power, time series foundation models (TSFMs) have
emerged as a compelling alternative \citep{liang_2024}:
pretrained on large, heterogeneous datasets, TSFMs can generalize learned temporal patterns to new forecasting tasks without further task-specific training, and have demonstrated impressive performance in benchmarks such as \texttt{fev-bench} \citep{shchur_fev-bench_2026} or \texttt{GIFT-Eval} \citep{aksu_gift-eval_2024}. Given past observations of the target series, TSFMs can recognize the underlying structure from the context and implicitly activate the parts of the model that are best suited to it when similar series have been observed during pretraining. This zero-shot capability enables high scalability, allowing TSFMs to address forecasting tasks with limited historical data, while their input-driven predictions can improve robustness to distribution shifts. However, relying exclusively on target history has inherent limitations. It is most effective when the series exhibits strong autoregressive structure, but partially fails when the target series is affected by external factors such as public holidays, which recur at irregular intervals and can hardly be inferred from past observations alone. Recent work has therefore begun incorporating calendar information as categorical features into TSFMs with promising results, yet the impact of their encoding on forecast performance remains largely unexplored. This question is particularly relevant for pretrained models, since encoding is one of the few design choices available at inference time, and differences in pretraining strategies may lead to different requirements for representing additional categorical inputs. Furthermore, current benchmarks do not isolate covariate skill or test whether models use informative covariates while ignoring uninformative ones \citep{meyer2026t0}.
Our contributions in the present study are as follows:
\begin{itemize}
    \item We systematically analyze seven different encoding strategies for calendar information as covariates across four state-of-the-art TSFMs, and evaluate their effect on forecast accuracy on nine diverse datasets from multiple domains, that are influenced by social activities such as weekends and holidays.
    \item We employ Shapley additive explanations (SHAP) to examine how models utilize the available covariates and to assess the role of calendar information in forecasting performance.

\end{itemize}
This study is structured as follows. Section \ref{sec:relatedwork} reviews related work on TSFMs with covariate integration, with a particular focus on categorical feature encoding. Section \ref{sec:methodology} presents the proposed methodology. Section \ref{sec:ExperimentalSetup} describes the experimental design, while Section \ref{sec:results} and \ref{sec:discussion} present and discuss the results. Section \ref{sec:Conclusion_outlook} concludes the work with a summary and outlook.

\section{Related Work and Foundations}
\label{sec:relatedwork}
This section reviews related work on the integration of covariates into TSFMs and encoding strategies for calendar features.

\paragraph{Time Series Foundation Models and Covariates}
TSFMs have demonstrated impressive cross-domain performance, often matching or outperforming models trained from scratch \citep{aksu_gift-eval_2024, shchur_fev-bench_2026}. While early TSFMs, such as Chronos \citep{ansari_chronos_2024} or MOMENT \citep{goswami_moment_2024}, were mostly limited to univariate forecasting, recent models increasingly support the integration of past and future covariates. Two approaches can be distinguished. Adapters such as CoRa \citep{qin_cora_2025}, UniCA \citep{han_unica_2026} or TFMAdapter \citep{dange_tfmadapter_2025} are built on univariate backbones, whereas models like Chronos-2, TabPFN-TS-3 or TiRex-2 offer native covariate support. In studies across different domains such as electrical load forecasting \citep{hertel_nikoltchovska_2026a, kreusel_hertel_2026}, electricity price forecasting \citep{pan_evaluating_2026}, finance \citep{das_multivariate_2026} or photovoltaic generation \citep{longarini_time_2026}, covariate integration has demonstrated performance gains, highlighting the potential of covariate-informed TSFMs and motivating further research in this direction. However, gains are not universal, as \citet{meyer2026t0} report that covariates degrade the performance of $t_{0}-beta$  on 11 of 30 \texttt{fev}-tasks with known-future covariates. Whether these degradations stem from uninformative covariates themselves or from their representation
cannot be decided from the reported results.

\paragraph{Covariate Encoding}
Covariates in time series forecasting are commonly distinguished into continuous
and categorical features \citep{fahrmeir_regression_2021}. While continuous features are already numeric and, depending on the used model, require only standardization or normalization, categorical features must first be mapped to a numeric representation \citep{hancock_survey_2020}. For this, multiple encoding methods exist and can broadly be distinguished into target-agnostic and target-based approaches \citep{pargent_regularized_2022, micci-barreca_preprocessing_2001}. Target-agnostic methods, such as ordinal, one-hot, or cyclic encoding, derive representations exclusively from the categories themselves, whereas target-based methods additionally incorporate observed target values. In time series forecasting, categorical features are commonly calendar-derived, including temporal features such as hour, weekday, and month, as well as holiday information \citep{meisenbacher_review_2022}. While the former mainly differ in their temporal granularity, holiday features offer the possibility to further group holidays into different categories. \citet{ziel_2018a} compares 31 holiday encoding strategies on German load data, mainly distinguishing fixed-date and weekday-based holidays. Holiday modeling reduces forecast error by >80 \% on holidays and about 10 \% on ordinary days, while finer grouping is not consistently beneficial. \citet{cho_modeling_2013} group holidays by season, \citet{wagner_short-_2022} distinguish federal from regional holidays and introduce bridge days, while \citet{gaillard_additive_2016} additionally classify the days before and after public holidays, capturing their effects on electrical load.
In the context of TSFMs, categorical covariates have recently received closer attention, with several studies reporting improved forecast accuracy. \citet{kaas_probabilistic_2026} employ weekday and holiday features, \citet{kreusel_hertel_2026} and \citet{hertel_nikoltchovska_2026a} use holiday features, and \citet{grauer2026causality} use binary indicators for working hours in an office building. While all of the studies rely on different encoding schemes, the encoding itself is treated as an implementation detail rather than as an object of study, and a systematic comparison is missing. However, such a comparison is particularly relevant%
, as recent benchmarks such as \texttt{fev-bench} \citep{shchur_fev-bench_2026} and \texttt{FETS} \citep{obermeier_pruckner_2026} include covariate-informed tasks with calendar-derived features. 
In \texttt{fev-bench}, categorical covariates are heterogeneously encoded, with categories stored
as float32, string, or boolean. This heterogeneity is consequential, as no task declares a column as categorical and every model wrapper dispatches on the storage dtype alone. \texttt{FETS} sidesteps this problem by storing every covariate numerically: five sine–cosine pairs for the periodic calendar components and binary Int8 indicators for holidays and weekends.
While formulating generalizing statements across models is already difficult in the univariate setting due to data leakage \citep{meyer_kaltenpoth_2026}, 
the covariate setting adds further complexity, since each wrapper resolves the encoding differently. No prior work accounts for this additional factor, leaving its effect on TSFM performance unquantified.

\section{Methodology}
\label{sec:methodology}
This section introduces the two components used throughout this work. Section~\ref{sec:clustering_encodings} presents the calendar encodings under study.
Section~\ref{sec:shap_method} introduces the used SHAP approach. 

\subsection{Calendar Encodings}
\label{sec:clustering_encodings}
\begin{table}[h]
\centering
\footnotesize
\caption{Calendar encodings. All variants use hour (h), weekday (W), and holiday (H) information, differing in their representation and combination.}
\label{tab:calendar_encodings}
\small
\begin{tabularx}{\linewidth}{@{}l p{1.7cm} >{\raggedright\arraybackslash}X c >{\raggedright\arraybackslash}X@{}}
\toprule
\textbf{Encoding} &
\textbf{Holiday} &
\textbf{Representation} &
\textbf{Dim.} &
\textbf{Assumption} \\
\midrule

\encOrdinal
& binary flag
& ordinal (0--23; 0--6) $+$ flag
& 3
& -- \\

\encOnehot
& binary flag
& one-hot $(24{+}7)$ $+$ flag
& 32
& \rule{0pt}{4.5ex}-- \\

\encCyclic
& binary flag
& sin/cos $+$ flag
& 5
& \rule{0pt}{4.5ex}Hour and week follow fixed periodicity. \\

\encTcombo
& $\rightarrow$ Sunday
& joint target encoding%
& 1
& \rule{0pt}{2.25ex}Public holiday $\approx$ Sunday. \\

\encTholEight
& 8th weekday
& joint target encoding
& 1
& \rule{0pt}{4.5ex}Holidays are distinct from Sundays. \\

\encTholAll
& binary flag
& joint target encoding%
& 1
& Captures weekday-specific holiday effects. \\

\encTholPartial
& binary flag
& two separate target encodings%
& 2
& Keep weekday and holiday information separate. \\

\bottomrule
\end{tabularx}
\end{table}
Table \ref{tab:calendar_encodings} summarizes the calendar encodings applied in this study. We implement several different encodings, based on common approaches in the literature, roughly sorted by increasing complexity to cover most conceivable encoding strategies. All encodings use hour (h), weekday (W), and holiday (H) information, with the indices in the nomenclature indicating how each information is incorporated. The encodings can be divided into two categories. Conventional encodings, comprising ordinal, one-hot, and cyclic encodings, are based exclusively on calendar information and are further enriched with a binary holiday feature to account for the effect of public holidays. Furthermore, we employ simple target encoding strategies that compute the mean target value over the observations in the model context that share the same
combination of the incorporated categorical variables, leading to daily profiles on the scale of the target. 
Mathematically, the target encodings are described as follows: Let $\mathcal{X}_j$ denote the set of possible values of the $j$-th of $k$ categorical variables, let $x_t \in \mathcal{X}_1 \times \cdots \times \mathcal{X}_k$ denote the $k$-tuple of categorical values at time step $t$, and let ${x}^* \in \mathcal{X}_1 \times \cdots \times \mathcal{X}_k$ denote a specific tuple combination. Furthermore,
$\mathcal{C}=(1,\ldots,c)$ indexes the model context and
$\mathcal{F}=(c+1,\ldots,c+f)$ the forecast horizon. For a specific
combination ${x}^*$ we collect the context time steps that realize
it, $\mathcal{Z}_{{x}^*}:=\{t\in\mathcal{C}\mid x_t={x}^*\}$, and
define the associated target encoding as the mean target value over this
set,
\begin{equation}
T_{{x}^*} =
\frac{1}{|\mathcal{Z}_{{x}^*}|}
\sum_{t \in \mathcal{Z}_{{x}^*}} Y_t ,
\end{equation}
where $Y_t$ denotes the target at time step $t$. Every time step
$t\in\mathcal{C}\cup\mathcal{F}$ then receives the encoding $T_{x_t}$ of
its own combination, assuming that each combination that occurs in the horizon occurs at least once within the context. As the encodings are recomputed for every prediction window from the context alone, they
introduce no information beyond what the model already receives as input.\\
Detailed descriptions and visualizations of the encodings are part of Appendix \ref{sec:Appendix_Encodings}.

\subsection{SHAP-based Analysis for Identifying Contributions in Forecast}
\label{sec:shap_method}

To quantify the contributions of the calendar covariates relative to the past target, we adopt the sampling-free SHAP approach of \citet{hertel_nikoltchovska_2026a} and extend it from Chronos-2 and TabPFN-TS to TiRex-2 and TimesFM-3. 
SHAP \citep{lundberg_lee_2017} decomposes a forecast into additive contributions of feature groups $\mathcal{G} = \{g_1, \dots, g_M\}$, given by their Shapley values:
\begin{equation}
    \mathrm{SHAP}(g_i) = \sum_{S \subseteq \mathcal{G} \setminus \{g_i\}} \frac{|S|!\,(M - |S| - 1)!}{M!} \big( f(S \cup \{g_i\}) - f(S) \big),
\end{equation}
where $f(S)$ is the forecast given only the groups in coalition $S$. 
$\mathrm{SHAP}(g_i)$ is thus the average change in the forecast caused by adding $g_i$, averaged over all coalitions of the other groups.
Models with a fixed input dimension cannot evaluate $f(S)$ with features missing, so model-agnostic estimators such as KernelSHAP \citep{lundberg_lee_2017} replace absent features with samples from background data, which can create unrealistic feature combinations. 
TSFMs, in contrast, accept arbitrary context lengths and covariate sets, so absent groups can simply be withheld and $f(S)$ is evaluated directly. 
Following \citet{hertel_nikoltchovska_2026a}, the past target values are split into four temporal groups, and each covariate forms one additional group. We report the summed contribution of the temporal groups as a single past-target contribution. 
Details on grouping, masking, and model-specific adaptations are given in Appendix~\ref{sec:App_Shap}.

\section{Experimental Setup}
\label{sec:ExperimentalSetup}

Comparing encodings requires an evaluation that isolates the effect of the representation from the
effect of the model. We therefore evaluate every encoding on the same datasets, with the same
forecast windows, and with the same models. The following paragraphs summarize the models, the
datasets, the experiment design, and the metrics. A more detailed description is part of Appendix~\ref{sec:App_ExperimentalSetup}.

\paragraph{Models} We evaluate four state-of-the-art TSFMs, Chronos-2 \citep{ansari_shchur_2025a}, TabPFN-TS-3 \citep{grinsztajn_floge_2026}, TiRex-2 \citep{podest_pichler_2026}, and TimesFM-3 \citep{das_decoder-only_2024,timesfm3_2026}, alongside two baselines. These are a seasonal naive model with a 168-hour seasonal period and the pure target encoding \encTcombo{}, used without TSFMs and recomputed for each forecast window from the same context used for the TSFMs. For a more detailed description of all models and the model pipeline see Appendix \ref{sec:App_models}.

\paragraph{Datasets} We evaluate nine datasets across the domains energy, transportation, traffic and water. Each dataset provides exactly one year of test data at an hourly resolution. To our knowledge, none of the datasets are included in the pretraining corpora of the employed TSFMs. A detailed list of all datasets is provided in Appendix \ref{sec:App_datasets}. We enrich each dataset with holiday information derived via the \texttt{holidays} Python library, using its most precise known geographic location.

\paragraph{Evaluation Metrics} For consistency with \texttt{fev-bench}, we evaluate point and probabilistic forecasts using mean absolute scaled error (MASE) \citep{hyndman_koehler_2006} and scaled quantile loss (SQL) \citep{koenker_regression_1978,hyndman_koehler_2006}, as defined in Appendix \ref{sec:App_metrics}. Both metrics are scaled by the in-sample error of a seasonal-naive forecast with a 168-hour seasonal period, computed separately for each forecast window. Both metrics are implemented using the \texttt{fev-bench} evaluation library \citep{shchur_fev-bench_2026}.
\paragraph{Experiment Design}
We evaluate all models in a zero-shot setting, using a rolling evaluation across all datasets and encodings. Since longer context lengths have been shown to improve TSFM performance \citep{hertel_putz_2026b, kreusel_hertel_2026}, we use a context length of 8192%
, as it is the maximum context length of Chronos-2. Forecast horizon and stride are both set to 24 hours, and each forecast starts at 00:00. To isolate covariate effects, we report holiday and non-holiday results separately.

\paragraph{Hypotheses}
Within each model, encoding is the only factor varied across runs, with all other parameters held constant. Since all encodings contain identical calendar information and differ only in representation and covariate count, accuracy differences can be attributed to the encoding. Based on this setup, we formulate three
expectations. First, as the four TSFMs differ in their pretraining and in how covariates enter the
architecture, we expect the sensitivity to the representation to be model-specific. Second, we
expect gains to concentrate around holidays, which, unlike hour and weekday effects, cannot be
inferred from the target history. Third, we expect target encodings to be more robust, as they
condense target-scale information into one or two covariates, whereas conventional encodings add up
to 32 target-agnostic covariates that may dilute the autoregressive signal.
\section{Results}
\label{sec:results}

\begin{table}[H]
\centering
\footnotesize
\caption{Overall results across all datasets, split by day type (mean over 9 datasets).}
\label{tab:eval5-overall-split}
\resizebox{\textwidth}{!}{%
\begin{tabular}{ll rr rr rr}
\toprule
& & \multicolumn{2}{c}{\textbf{All days}} & \multicolumn{2}{c}{\textbf{Holidays only}} & \multicolumn{2}{c}{\textbf{Non-holidays only}} \\
\cmidrule(lr){3-4} \cmidrule(lr){5-6} \cmidrule(lr){7-8}
\textbf{Model} & \textbf{Encoding} & \textbf{MASE} & \textbf{SQL} & \textbf{MASE} & \textbf{SQL} & \textbf{MASE} & \textbf{SQL}
 \\
\midrule
\multirow{2}{*}{Baselines}
 & seasonal-naive (168h) & \underline{1.0362} & 0.9984 & 2.9358 & 2.6207 & \underline{0.9682} & 0.9404 \\
 & $T_{h,W}(H=Sun)_{naive}$    & 1.1050 & \underline{0.9468} & \underline{1.5369} & \underline{1.3182} & 1.0882 & \underline{0.9322} \\
\midrule
\multirow{8}{*}{Chronos-2}
 & \encNone              & 0.6903 & 0.5534 & 2.2299 & 1.9242 & 0.6353 & 0.5044 \\
 & \encOrdinal           & 0.6760 & 0.5381 & 1.5630 & 1.2854 & 0.6440 & 0.5111 \\
 & \encOnehot            & 0.7099 & 0.5690 & 2.0244 & 1.7294 & 0.6626 & 0.5272 \\
 & \encCyclic            & 0.6861 & 0.5469 & 1.6257 & 1.3509 & 0.6523 & 0.5179 \\
 & \encTcombo            & 0.6448 & 0.5125 & 1.2509 & 0.9878 & 0.6223 & 0.4949 \\
 & \encTholEight         & 0.6444 & 0.5121 & 1.2354 & \underline{0.9737} & 0.6224 & 0.4949 \\
 & \encTholAll           & 0.6523 & 0.5187 & 1.4161 & 1.1264 & 0.6237 & 0.4959 \\
 & \encTholPartial       & \underline{0.6435} & \underline{0.5120} & \underline{1.2285} & 0.9800 & \underline{0.6218} & \underline{0.4947} \\
\midrule
\multirow{8}{*}{TabPFN-TS-3}
 & \encNone              & 0.8441 & 0.6847 & 2.8198 & 2.3814 & 0.7727 & 0.6234 \\
 & \encOrdinal           & 0.7666 & 0.6091 & 1.3021 & 0.9963 & 0.7457 & 0.5940 \\
 & \encOnehot            & 0.7600 & 0.6041 & 1.3740 & 1.0654 & 0.7361 & 0.5861 \\
 & \encCyclic            & 0.7585 & 0.6050 & 1.2318 & 0.9516 & 0.7400 & 0.5914 \\
 & \encTcombo            & \underline{0.6968} & \underline{0.5570} & \underline{1.1628} & \underline{0.9327} & 0.6785 & \underline{0.5423} \\
 & \encTholEight         & 0.6988 & 0.5602 & 1.2228 & 1.0059 & \underline{0.6785} & 0.5429 \\
 & \encTholAll           & 0.7151 & 0.5741 & 1.6734 & 1.4072 & 0.6787 & 0.5425 \\
 & \encTholPartial       & 0.7046 & 0.5635 & 1.3610 & 1.0737 & 0.6793 & 0.5438 \\
\midrule
\multirow{8}{*}{TiRex-2}
 & \encNone              & 0.7842 & 0.6315 & 2.1557 & 1.8147 & 0.7360 & 0.5901 \\
 & \encOrdinal           & 0.8116 & 0.6496 & 2.0781 & 1.7190 & 0.7672 & 0.6122 \\
 & \encOnehot            & 0.7930 & 0.6403 & 2.2586 & 1.8944 & 0.7412 & 0.5961 \\
 & \encCyclic            & 0.7841 & 0.6319 & 2.2174 & 1.8607 & 0.7336 & 0.5887 \\
 & \encTcombo            & 0.7519 & 0.6027 & \underline{1.2318} & \underline{1.0217} & 0.7334 & 0.5864 \\
 & \encTholEight         & \underline{0.7518} & \underline{0.6019} & 1.2550 & 1.0321 & \underline{0.7324} & \underline{0.5852} \\
 & \encTholAll           & 0.7636 & 0.6124 & 1.5372 & 1.2922 & 0.7342 & 0.5864 \\
 & \encTholPartial       & 0.8034 & 0.6439 & 2.0996 & 1.7744 & 0.7565 & 0.6030 \\
\midrule
\multirow{8}{*}{TimesFM-3}
 & \encNone              & 0.6542 & 0.5269 & 2.1180 & 1.7979 & 0.6021 & 0.4816 \\
 & \encOrdinal           & 0.6306 & 0.5050 & 1.3840 & 1.1429 & 0.6032 & 0.4817 \\
 & \encOnehot            & 0.6370 & 0.5103 & 1.5529 & 1.2878 & 0.6036 & 0.4819 \\
 & \encCyclic            & \textbf{\underline{0.6263}} & 0.5019 & 1.3128 & 1.0888 & \textbf{\underline{0.6014}} & \textbf{\underline{0.4806}} \\
 & \encTcombo            & 0.6275 & \textbf{\underline{0.5011}} & \textbf{\underline{0.9969}} & \textbf{\underline{0.7913}} & 0.6134 & 0.4900 \\
 & \encTholEight         & 0.6302 & 0.5036 & 1.0566 & 0.8435 & 0.6139 & 0.4906 \\
 & \encTholAll           & 0.6397 & 0.5120 & 1.2499 & 1.0149 & 0.6165 & 0.4928 \\
 & \encTholPartial       & 0.6472 & 0.5187 & 1.0305 & 0.8179 & 0.6328 & 0.5074 \\
\bottomrule
\end{tabular}}
\end{table}

Table~\ref{tab:eval5-overall-split} reports the overall results averaged over the nine datasets, and the per-dataset tables are given in Appendix~\ref{sec:App_Results}. As all test sets are equally long and MASE and SQL are scale-independent, no dataset disproportionately influences the overall results.
We report results for all days, for holiday windows, and for non-holiday windows, and within a column, the best value across all models and
baselines is set in bold and the best value within a model group is underlined.
First, we analyze general effects across models, followed by the effects of the encodings within each model. 

\begin{figure}[H]
    \centering
    \includegraphics[width=1.0\linewidth]{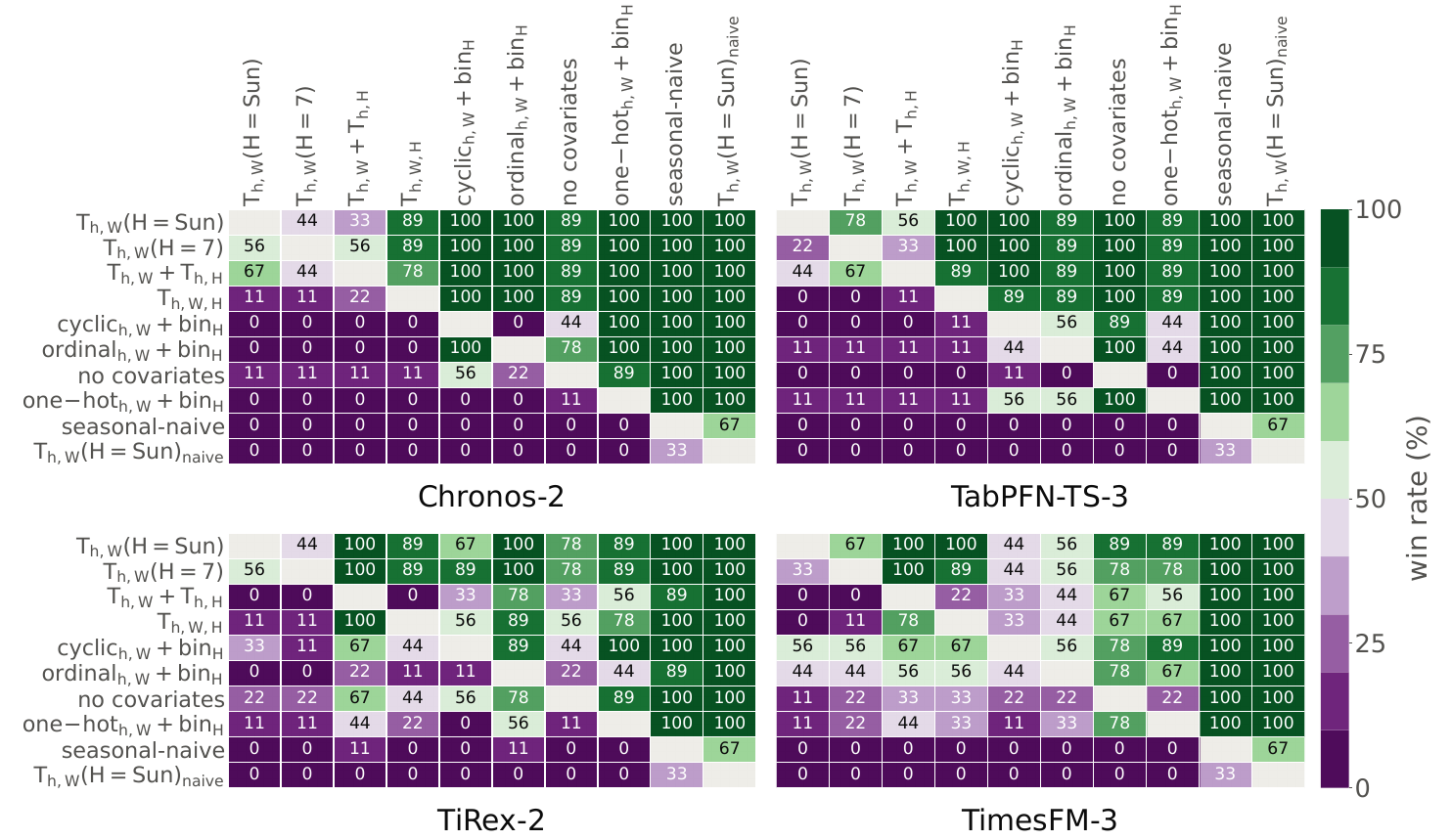}
        \caption{Pairwise win rate in MASE, as the share of the nine datasets won by the row encoding. }
    \label{fig:winrate}
\end{figure}

\paragraph{Overall results across all days}
TimesFM-3 achieves the best overall performance on model-level with the \encCyclic encoding, followed by Chronos-2 with the \encTholPartial encoding. Relative to the best performance of each model, no general pattern across encodings emerges, as can be seen in the pairwise win rate in Figure \ref{fig:winrate}. While Chronos-2, TabPFN-TS-3, and TiRex-2 perform best with target encodings, TimesFM-3 shows largely comparable performance across encodings, with its best result not obtained by a target encoding.

\paragraph{Results on holidays only}
TimesFM-3 achieves the overall best holiday performance with the \encTcombo encoding. On holiday windows covariates improve performance across 26 of the 28 model-encoding combinations, with TiRex-2 using \encOnehot and \encCyclic encoding as the only exceptions. Across models, target encodings clearly outperform conventional encodings on holidays, indicating that they capture substantially more information than the binary holiday representations of the conventional encodings.

\paragraph{Results on non-holidays only}
TimesFM-3 achieves
the best model-level performance with the \encCyclic encoding, which explains its overall advantage given that non-holidays constitute the vast majority of days. Encoding effects vary substantially across models, preventing a general conclusion. While Chronos-2, TiRex-2, and TimesFM-3 show only marginal gains or even deteriorations from covariates, TabPFN-TS-3 exhibits a clear improvement in mean forecast accuracy across all encodings.

The findings so far describe general effects across models. In the following, the focus shifts to the different effects of encodings within each model.

\begin{figure}[H]
    \centering    \includegraphics[width=1.0\linewidth]{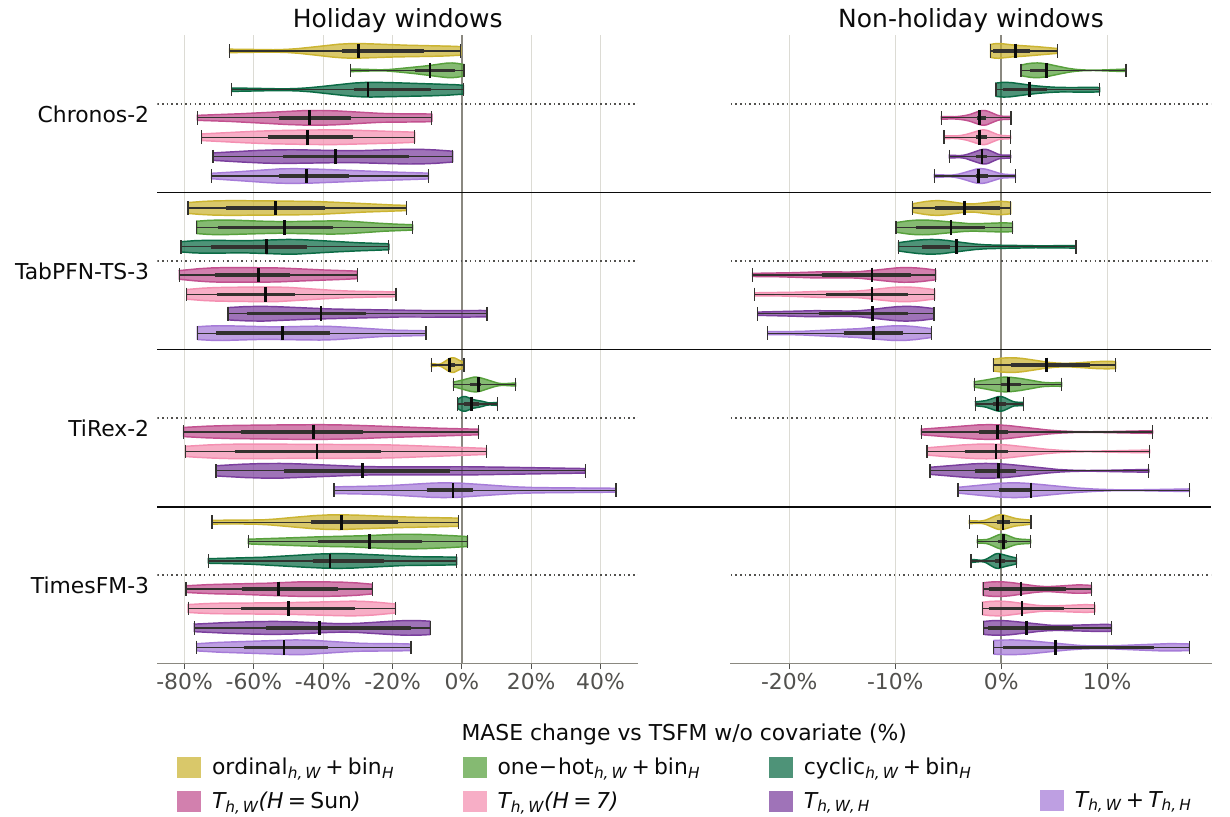}
    \caption{MASE change relative to the TSFM w/o covariates, averaged over nine datasets. Vertical lines indicate the mean, and horizontal lines the interquartile range (25th–75th percentile). Most models improve on holidays independently of the encoding, while results on non-holidays are mixed.}
    \label{fig:effect_overall}
\end{figure}

\paragraph{Chronos-2 benefits most from two separate target encodings}
Chronos-2 reaches its best results on all window types using target encodings, with the \encTholPartial{} encoding. While conventional encodings also improve performance on holidays, they decrease performance on non-holidays (cf. Figure \ref{fig:effect_overall}), indicating that the target scale representation of calendar information helps the model to capture autoregressive structure within the target series, while the pure calendar information of the conventional encoding dilutes the signal.
 
\paragraph{TabPFN-TS-3 benefits from every encoding strategy}
TabPFN-TS-3 is the only model that improves under all seven encodings and in all three window
types.
While improvements of target encodings and conventional encodings are on par on holidays, target encodings beat the conventional encodings on non-holidays (cf. Figure \ref{fig:effect_overall}). Still, conventional encodings improve the non-holiday MASE by up to 5.8 \%, indicating that the model relies on additional calendar information even when capturing the autoregressive structure of non-holiday periods.

\paragraph{TiRex-2 can benefit from covariates but is sensitive to multidimensional encodings}
Of the seven encodings, only the three joint target encodings improve the mean all-day MASE of TiRex-2.
All encodings that comprise two or more covariates show almost no, or a negative effect on the mean performance on both window types, indicating the incapability of the model to handle multiple covariates at once. 
A noteworthy observation is that TiRex-2 does not show substantial improvements on the \texttt{mta\_subway} and \texttt{mta\_bus} datasets with any encoding, especially in comparison to the other three models (cf. Appendix \ref{sec:App_results_mtasubway} and \ref{sec:App_results_mtabus}, Figures \ref{fig:effect_mta_subway_hourly} and \ref{fig:effect_mta_bus_hourly}).
\paragraph{TimesFM-3 benefits from covariates only on holidays} On holidays, TimesFM-3 substantially improves over the non-covariate model with all encoding strategies, and target encodings consistently outperform conventional encodings. In contrast, on non-holidays, covariates provide no meaningful improvement over the non-covariate baseline, and all target encodings reduce performance.

\section{Discussion}
\label{sec:discussion}
This section discusses generalizable observations across models based on the experiments in Section \ref{sec:results}. We use SHAP analyses to provide further insights into the observed effects. Beyond cross-model comparisons, SHAP analyses can also investigate encoding-dependent performance differences within models, helping to identify which encodings work, and in which scenarios they fail.

\begin{figure}[H]
    \centering
    \includegraphics[width=1.0\linewidth]{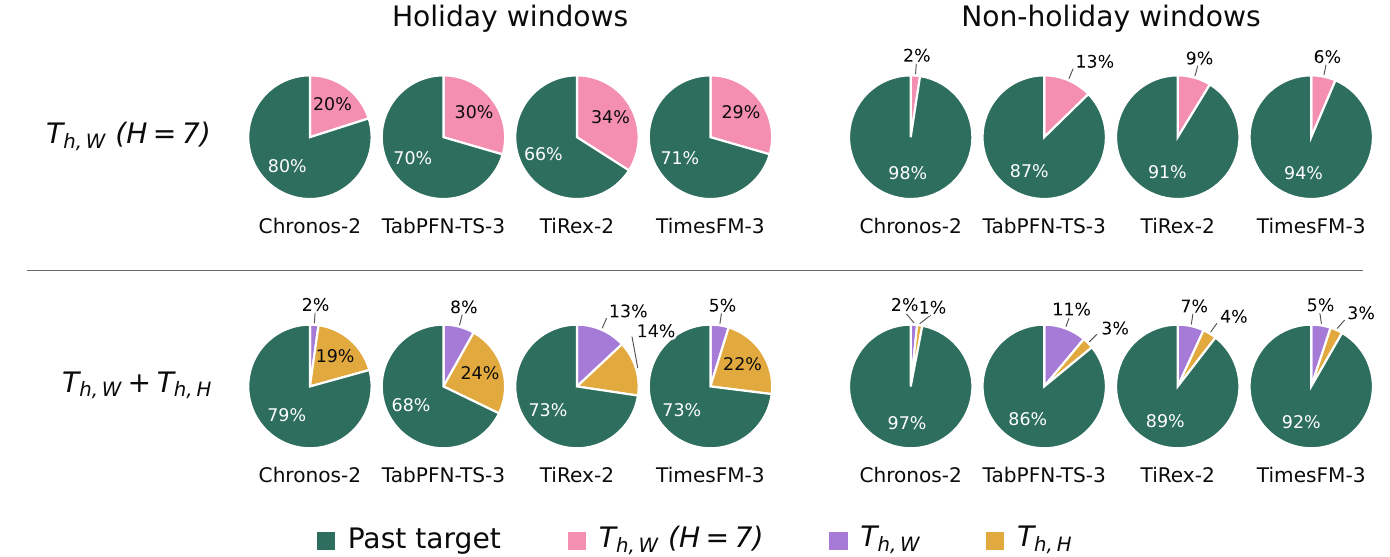}
    \caption{Share of the mean absolute SHAP value for the \encTholEight and \encTholPartial encodings, averaged across
    datasets. 
    Covariates contribute 20--34 \% on holidays but at most 14 \% on non-holidays. With \encTholPartial, most holiday contribution moves to \encThH, except for TiRex-2.}
    \label{fig:shap_pies}
\end{figure}

\paragraph{Encodings substantially impact performance across all models and data regimes}
The results demonstrate that the choice of encoding substantially affects model performance across all models, with effects differing between holidays and non-holidays. 
While the gap in MASE between the best and worst encoding within a model ranges from +4.8 \% (TiRex-2) to +9.9 \% (TabPFN-TS-3)
on non-holidays, the gap is between +44 \% (TabPFN-TS-3) to +83 \% (TiRex-2) on holidays. 
The large differences between holidays and non-holidays can be explained by the SHAP contributions shown in Figure~\ref{fig:shap_pies}, illustrated for the \encTholEight and \encTholPartial encodings. On holidays, all models assign substantially greater importance to the calendar features than on non-holidays. On non-holidays, Chronos-2 is less influenced by either of the two encodings than the other models,  instead relying primarily on the autoregressive structure of the series, explaining the only marginally changed performance compared to its non-covariate variant. In contrast, TimesFM-3 places more weight on the calendar features than Chronos-2, yet performs worse than its own non-covariate variant under both depicted encodings. This indicates that the model already captures the underlying autoregressive structure effectively, and that the covariate information
dilutes
the signal.

\begin{figure}[H]
    \centering
    \includegraphics[width=1.0\linewidth]{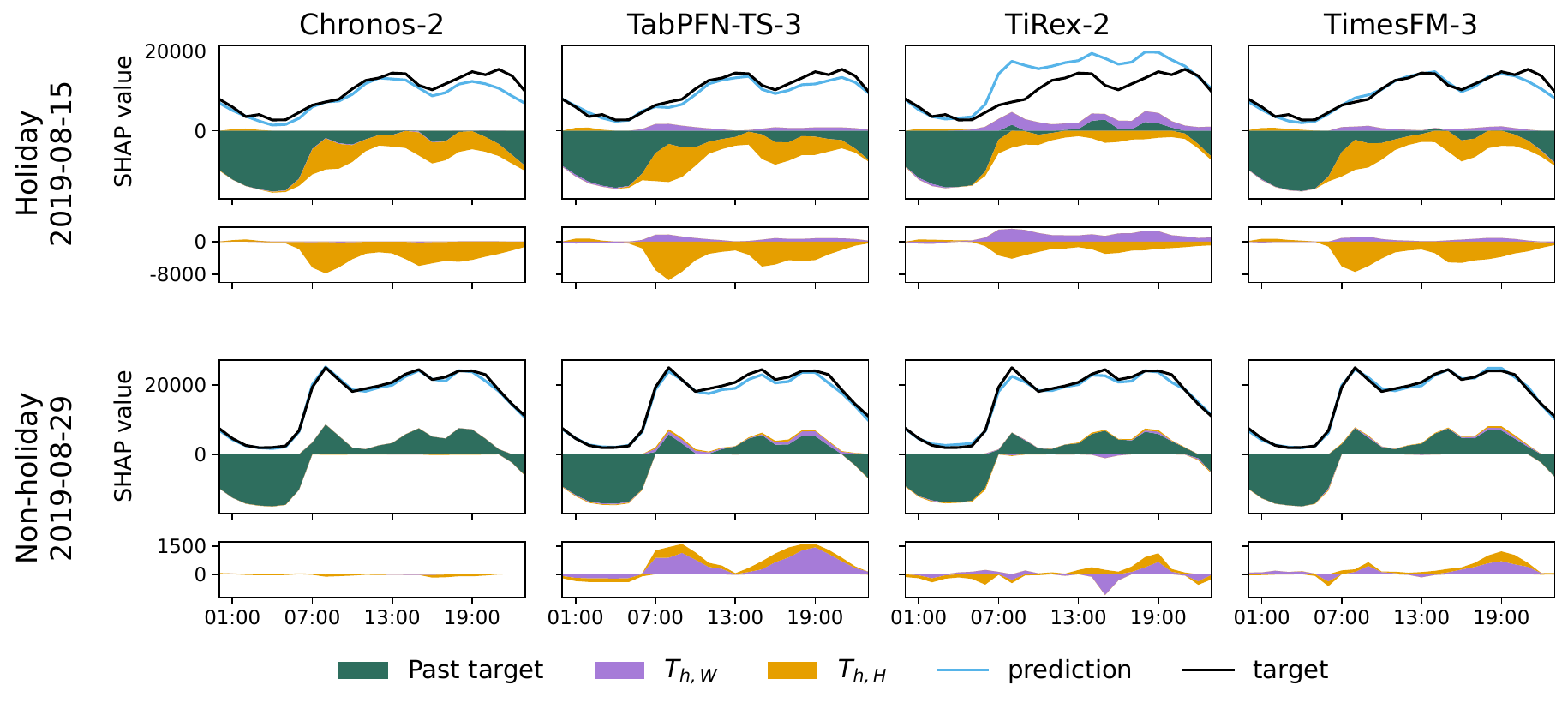}
    \caption{Hourly forecasts and SHAP values for a holiday
    and a non-holiday
    from \texttt{m\_dense} with \encTholPartial. Upper panels: median forecast, target, and SHAP contributions relative to the base prediction $f(\emptyset)$. Lower panels: \encThW and \encThH contributions on a separate scale. Units: vehicles per hour. On the holiday, \encThH lowers the forecast for all models except TiRex-2, which overpredicts.}
    \label{fig:Shap_holpartial}
\end{figure}

\paragraph{Encoding effectiveness and utilization vary across models}
While encoding choice affects performance across all models, the effects of specific encodings do not generalize across them. Figure~\ref{fig:Shap_holpartial} illustrates this exemplarily for a holiday and a non-holiday of the \texttt{m\_dense} dataset with the \encTholPartial encoding. While Chronos-2, TabPFN-TS-3, and TimesFM-3 on holidays correctly assign more weight to the holiday covariate \encThH than the weekday and hour information \encThW, the two covariate effects nearly cancel out for TiRex-2, producing an
overestimated forecast.
Such differences in how models handle specific encoding strategies likely stem from architectural and training differences. Chronos-2, for instance, applies target encoding during training but specifies neither which features are included nor whether their granularity adapts to the target frequency.
\paragraph{Encodings in the context of benchmarking}

The strong influence of encoding choice on model performance, combined with its varying effects across models, has direct implications for multivariate benchmarking. Current benchmarks such as \texttt{FETS} \citep{obermeier_pruckner_2026} and \texttt{fev-bench} \citep{shchur_fev-bench_2026} evaluate all models under a single fixed encoding, determined by data type of the dataset. Their conclusions are therefore limited to the specific model–encoding combinations tested and cannot be generalized to the full potential of a model, as alternative encodings may change the observed ranking. This can for example be observed in a second experimental run with context length 2048 (cf. Appendix \ref{sec:App_Results2048}).
Two complementary remedies are possible. Benchmarks could include multiple encoding strategies as an explicit evaluation factor, allowing each model to be assessed across different preprocessing choices rather than under a single fixed encoding. Alternatively, model developers could integrate model-specific categorical feature handling directly into the model, for example by accepting ordinal inputs and applying the internally optimal transformation, or publish a best-practice encoding recommendation against which their model should be evaluated. Both approaches 
shift the object of evaluation from the performance of a particular encoding to the actual potential of a model in multivariate forecasting.

\paragraph{Limitations}
All experiments use context length 8192, the maximum of Chronos-2, corresponding to roughly 49 weeks at hourly resolution. A second run with context 2048 (cf. Appendix \ref{sec:App_Results2048}) confirms our findings, but a context exceeding one year is of interest, as human behavior often follows an annual cycle that would then be fully contained in the context. Furthermore, all datasets are evaluated at hourly resolution, and additional resolutions would further validate the results. Finally, our study is restricted to settings where the target follows calendar effects such as holidays and weekends. For problems driven mainly by continuous exogenous drivers such as solar irradiance or temperature, no comparable benefit is expected. Incorporating such
covariates is left to future work.

\section{Summary and Outlook}
\label{sec:Conclusion_outlook}
    
Although covariates are increasingly used with TSFMs, their encoding is typically treated as an implementation detail rather than systematically investigated. Existing multivariate benchmarks commonly fix a single encoding across models and datasets, potentially conflating model capability with encoding choice. To demonstrate this, we systematically evaluate multiple calendar encodings across different TSFMs and datasets.
Our results can be summarized in three key findings:

\begin{itemize}
    \item Encodings impact performance across all models and data regimes.
    \item The effectiveness and utilization of specific encodings varies across models, indicating a strong interaction between model architecture and representation.
    \item Encoding effects should be considered in future multivariate benchmarks, as fixing a single encoding can influence and potentially distort model rankings.
\end{itemize}

Future work should revisit existing benchmarks by treating encoding as an explicit experimental factor. Beyond evaluation, our results suggest that improved encoding strategies offer an opportunity for future research. A promising direction is a fully data-driven approach, such as in-context clustering, where time-series-specific characteristics are identified from the context and used to derive or select suitable encodings. This could enable more adaptive representations of covariate information and lead to further improved TSFM forecasting performance.

\clearpage

\section*{Reproducibility Statement} All code required to reproduce our results will be made publicly available upon publication, including the evaluation pipeline, and the scripts that generate every table and figure. All experiments use publicly available benchmark data.

\section*{Generative AI Usage Disclosure}
In this work, we used generative AI tools for designing or providing feedback on
research methodology or experiments, implementing methods, assisting with translation,
cleaning and reformatting data sets, and supporting qualitative and thematic data
analysis. We have not used generative AI tools for proposing or refining hypotheses
and interpreting results, and generating synthetic data sets, helping develop
theoretical models or conceptual frameworks, formulating mathematical claims,
providing critical ingredients for proving mathematical claims, and assisting in the
writing of proofs are not applicable to this work. Additionally, we used generative AI
tools for creating or modifying scientific figures or images, creating or editing
software code, sourcing or searching for information, editing a research paper to
improve readability, identifying relevant literature, and formatting references. We
used the following generative AI tool: Claude Code Opus 5. We have reviewed all
AI-assisted work. All AI-assisted code was reviewed and tested for correctness by two
authors. AI-generated text was checked and edited by the authors against the
underlying results. Every AI-suggested reference was verified against the original
publication. We take responsibility for the final content of this work, including
text, claims or artifacts produced with the aid of generative AI.
\bibliography{bibliography}

@article{meyer2026t0,
  title   = {$t_0$: A Time-Series Foundation Model for Forecasting with Context},
  author  = {Meyer, Lucas and Sole, Claudio and Xiang, Huikan and Li, Nicolas
             and Franceschino, Lucas and Quera-Bofarull, Arnau and Scholl, Maarten P.
             and Fainberg, Joachim and N{\'e}giar, Geoffrey},
  journal = {arXiv preprint arXiv:2609.24559},
  url={https://arxiv.org/pdf/2609.24559v1},
  year    = {2026}
}

@misc{aksu_gift-eval_2024,
    title = {{GIFT}-{Eval}: {A} {Benchmark} {For} {General} {Time} {Series} {Forecasting} {Model} {Evaluation}},
    shorttitle = {{GIFT}-{Eval}},
    url = {http://arxiv.org/abs/2410.10393},
    doi = {10.48550/arXiv.2410.10393},
    urldate = {2025-12-08},
    publisher = {arXiv},
    author = {Aksu, Taha and Woo, Gerald and Liu, Juncheng and Liu, Xu and Liu, Chenghao and Savarese, Silvio and Xiong, Caiming and Sahoo, Doyen},
    month = nov,
    year = {2024},
}

@misc{ansari_chronos_2024,
    title = {Chronos: {Learning} the {Language} of {Time} {Series}},
    url = {https://arxiv.org/abs/2403.07815},
    doi = {10.48550/arXiv.2403.07815},
    author = {Ansari, Abdul Fatir and Stella, Lorenzo and Turkmen, Caner and Zhang, Xiyuan and Mercado, Pedro and Shen, Huibin and Shchur, Oleksandr and Rangapuram, Syama Sundar and Arango, Sebastian Pineda and Kapoor, Shubham and Zschiegner, Jasper and Maddix, Danielle C. and Wang, Hao and Mahoney, Michael W. and Torkkola, Kari and Wilson, Andrew Gordon and Bohlke-Schneider, Michael and Wang, Yuyang},
    month = nov,
    year = {2024},
}

@misc{ansari_shchur_2025a,
	title = {Chronos-2: {From} {Univariate} to {Universal} {Forecasting}},
	shorttitle = {Chronos-2},
	url = {http://arxiv.org/abs/2510.15821},
	doi = {10.48550/arXiv.2510.15821},
	urldate = {2025-12-08},
	publisher = {arXiv},
	author = {Ansari, Abdul Fatir and Shchur, Oleksandr and Küken, Jaris and Auer, Andreas and Han, Boran and Mercado, Pedro and Rangapuram, Syama Sundar and Shen, Huibin and Stella, Lorenzo and Zhang, Xiyuan and Goswami, Mononito and Kapoor, Shubham and Maddix, Danielle C. and Guerron, Pablo and Hu, Tony and Yin, Junming and Erickson, Nick and Desai, Prateek Mutalik and Wang, Hao and Rangwala, Huzefa and Karypis, George and Wang, Yuyang and Bohlke-Schneider, Michael},
	month = oct,
	year = {2025},
}

@inproceedings{auer_tirex_2025,
    title = {{TiRex}: Zero-Shot Forecasting Across Long and Short Horizons with Enhanced In-Context Learning},
    shorttitle = {{TiRex}},
    booktitle = {Advances in Neural Information Processing Systems 38},
    author = {Auer, Andreas and Podest, Patrick and Klotz, Daniel and Böck, Sebastian and Klambauer, Günter and Hochreiter, Sepp},
    year = {2025},
    pages = {64179--64230},
    publisher = {Neural Information Processing Systems Foundation},
    url = {https://doi.org/10.52202/085713-1929},
}

@article{cho_modeling_2013,
    title = {Modeling and {Forecasting} {Daily} {Electricity} {Load} {Curves}: {A} {Hybrid} {Approach}},
    volume = {108},
    shorttitle = {Modeling and {Forecasting} {Daily} {Electricity} {Load} {Curves}},
    url = {https://doi.org/10.1080/01621459.2012.722900},
    number = {501},
    urldate = {2026-08-12},
    journal = {Journal of the American Statistical Association},
    publisher = {Taylor \& Francis},
    author = {Cho, Haeran and Goude, Yannig and Brossat, Xavier and Yao, Qiwei},
    month = mar,
    year = {2013},
    pages = {7--21},
}

@misc{daiad_alicante_swm_2026,
  author       = {{DAIAD Project}},
  title        = {Household Water Consumption (Average)},
  howpublished = {HARDMIN Research Data Repository (HEAL-Link), Helix},
  year         = {2017},
  doi          = {10.26255/healyinb-b1xl},
  url          = {https://hardmin.heal-link.gr/dataset/700cde6f-ef69-41f0-8fec-9c3f74ae38d4},
  urldate      = {2026-08-24},
}

@inproceedings{dange_tfmadapter_2025,
    title = {{TFMAdapter}: {Lightweight} {Instance}-{Level} {Adaptation} of {Foundation} {Models} for {Forecasting} with {Covariates}},
    shorttitle = {{TFMAdapter}},
    url = {http://arxiv.org/abs/2509.13906},
    doi = {10.1145/3746252.3761272},
    urldate = {2025-12-23},
    booktitle = {Proceedings of the 34th {ACM} {International} {Conference} on {Information} and {Knowledge} {Management}},
    author = {Dange, Afrin and Sarawagi, Sunita},
    month = nov,
    year = {2025},
    pages = {498--507},
}

@inproceedings{das_decoder-only_2024,
    title = {A decoder-only foundation model for time-series forecasting},
    booktitle = {Proceedings of the 41st International Conference on Machine Learning},
    series = {Proceedings of Machine Learning Research},
    author = {Das, Abhimanyu and Kong, Weihao and Sen, Rajat and Zhou, Yichen},
    year = {2024},
    volume = {235},
    pages = {10148--10167},
    publisher = {PMLR},
    url = {https://proceedings.mlr.press/v235/das24c.html},
}

@misc{das_multivariate_2026,
    title = {Multivariate {Financial} {Forecasting} using the {Chronos} {Time} {Series} {Foundation} {Models}},
    url = {http://arxiv.org/abs/2605.21504},
    doi = {10.48550/arXiv.2605.21504},
    urldate = {2026-08-14},
    publisher = {arXiv},
    author = {Das, Sanjiv R. and Goyal, Tarang and Yadav, Mohini},
    month = may,
    year = {2026},
}

@misc{entsoe2026transparency,
  author       = {{ENTSO-E}},
  title        = {Transparency Platform},
  year         = {2026},
  howpublished = {\url{https://transparency.entsoe.eu}},
  note         = {Accessed: 2026-08-06}
}

@book{fahrmeir_regression_2021,
    address = {Berlin, Heidelberg},
    title = {Regression: {Models}, {Methods} and {Applications}},
    copyright = {https://www.springer.com/tdm},
    shorttitle = {Regression},
    url = {https://doi.org/10.1007/978-3-662-63882-8},
    language = {en},
    urldate = {2026-08-17},
    publisher = {Springer},
    author = {Fahrmeir, Ludwig and Kneib, Thomas and Lang, Stefan and Marx, Brian D.},
    year = {2021},
}

@misc{fanaee-t_2013,
    title = {Bike {Sharing}},
    url = {https://archive.ics.uci.edu/dataset/275},
    doi = {10.24432/C5W894},
    urldate = {2026-08-24},
    publisher = {UCI Machine Learning Repository},
    author = {Fanaee-T, Hadi},
    year = {2013},
}

@article{gaillard_additive_2016,
    title = {Additive models and robust aggregation for {GEFCom2014} probabilistic electric load and electricity price forecasting},
    volume = {32},
    url = {https://doi.org/10.1016/j.ijforecast.2015.12.001},
    language = {en},
    number = {3},
    urldate = {2026-08-17},
    journal = {International Journal of Forecasting},
    author = {Gaillard, Pierre and Goude, Yannig and Nedellec, Raphaël},
    month = jul,
    year = {2016},
    pages = {1038--1050},
}

@misc{goswami_moment_2024,
    title = {{MOMENT}: {A} {Family} of {Open} {Time}-series {Foundation} {Models}},
    shorttitle = {{MOMENT}},
    url = {http://arxiv.org/abs/2402.03885},
    doi = {10.48550/arXiv.2402.03885},
    urldate = {2026-08-14},
    publisher = {arXiv},
    author = {Goswami, Mononito and Szafer, Konrad and Choudhry, Arjun and Cai, Yifu and Li, Shuo and Dubrawski, Artur},
    month = oct,
    year = {2024},
}

@inproceedings{grauer2026causality,
  author    = {Grauer, Dennis and Ploennigs, Joern},
  title     = {Causality-Informed Time-Series Foundation Models for Building Performance Prediction},
  booktitle = {Proceedings of the 2026 European Conference on Computing in Construction (EC$^3$)},
  year      = {2026},
  address   = {Corfu, Greece},
  doi       = {10.35490/ec3.2026.327},
  url       = {https://ec-3.org/wp-content/uploads/2026/08/EC32026_327.pdf},
}

@misc{grinsztajn_floge_2026,
	title = {{TabPFN}-3: {Technical} {Report}},
	shorttitle = {{TabPFN}-3},
	url = {http://arxiv.org/abs/2605.13986},
	doi = {10.48550/arXiv.2605.13986},
	urldate = {2026-08-04},
	publisher = {arXiv},
	author = {Grinsztajn, Léo and Flöge, Klemens and Key, Oscar and Birkel, Felix and Jund, Philipp and Roof, Brendan and Manium, Mihir and Hoo, Shi Bin and Bühler, Magnus and Garg, Anurag and Safaric, Dominik and Robertson, Jake and Jäger, Benjamin and Alessi, Simone and Hayler, Adrian and Moroshan, Vladyslav and Purucker, Lennart and Singer, Philipp and Arazi, Alan and Siems, Julien and Metzen, Jan Hendrik and Grab, Georg and Erickson, Nick and Guo, Siyuan and Kalfon, Eliott and Bing, Simon and Salinas, David and Cornu, Clara and Wehrhahn, Lilly Charlotte and Kriuchkova, Diana and Kaya, Kursat and Sidhoum, Lydia and Salmon, Marie and Chen, Jerry and Hulsebos, Madelon and LeCun, Yann and Müller, Samuel and Schölkopf, Bernhard and Gambhir, Sauraj and Hollmann, Noah and Hutter, Frank},
	month = may,
	year = {2026},
}

@inproceedings{han_unica_2026,
    title = {{UniCA}: Unified Covariate Adaptation for Time Series Foundation Model},
    booktitle = {International Conference on Learning Representations ({ICLR})},
    author = {Han, Lu and Liu, Yu and Li, Lan and Deng, Qiwen and Jiang, Jian and Sun, Yinbo and Yu, Zhe and Wang, Binfeng and Lu, Xingyu and Ma, Lintao and Ye, Han-Jia and Zhan, De-Chuan},
    year = {2026},
    url = {https://arxiv.org/abs/2506.22039},
}

@article{hancock_survey_2020,
    title = {Survey on categorical data for neural networks},
    volume = {7},
    url = {https://doi.org/10.1186/s40537-020-00305-w},
    language = {en},
    number = {1},
    urldate = {2026-08-17},
    journal = {Journal of Big Data},
    author = {Hancock, John T. and Khoshgoftaar, Taghi M.},
    month = dec,
    year = {2020},
    pages = {28},
}

@inproceedings{hertel_nikoltchovska_2026a,
	address = {New York, NY, USA},
	series = {E-{Energy} '26},
	title = {Explainable {Load} {Forecasting} with {Covariate}-{Informed} {Time} {Series} {Foundation} {Models}},
	url = {https://doi.org/10.1145/3744255.3811724},
	urldate = {2026-08-04},
	booktitle = {Proceedings of the 17th {ACM} {International} {Conference} on {Future} and {Sustainable} {Energy} {Systems}},
	publisher = {Association for Computing Machinery},
	author = {Hertel, Matthias and Nikoltchovska, Alexandra and Pütz, Sebastian and Schäfer, Benjamin and Mikut, Ralf and Hagenmeyer, Veit},
	month = jun,
	year = {2026},
	pages = {612--626},
}

@misc{hertel_putz_2026b,
	title = {A {Benchmark} for {Electrical} {Load} {Forecasting} {Across} {Grid} {Levels}: {Time}-{Series} {Transformers} {Outperform} {Established} {Methods}},
	shorttitle = {A {Benchmark} for {Electrical} {Load} {Forecasting} {Across} {Grid} {Levels}},
	url = {http://arxiv.org/abs/2607.15705},
	doi = {10.48550/arXiv.2607.15705},
	urldate = {2026-08-04},
	publisher = {arXiv},
	author = {Hertel, Matthias and Pütz, Sebastian and Kolar, Jonathan and Schäfer, Benjamin and Mikut, Ralf and Hagenmeyer, Veit},
	month = jul,
	year = {2026},
}

@misc{hoo_muller_2026a,
	title = {From {Tables} to {Time}: {Extending} {TabPFN}-v2 to {Time} {Series} {Forecasting}},
	shorttitle = {From {Tables} to {Time}},
	url = {http://arxiv.org/abs/2501.02945},
	doi = {10.48550/arXiv.2501.02945},
	urldate = {2026-08-10},
	publisher = {arXiv},
	author = {Hoo, Shi Bin and Müller, Samuel and Salinas, David and Hutter, Frank},
	month = jan,
	year = {2026},
}

@article{hyndman_koehler_2006,
	title = {Another look at measures of forecast accuracy},
	volume = {22},
	copyright = {https://www.elsevier.com/tdm/userlicense/1.0/},
	url = {https://doi.org/10.1016/j.ijforecast.2006.03.001},
	language = {en},
	number = {4},
	urldate = {2026-08-03},
	journal = {International Journal of Forecasting},
	author = {Hyndman, Rob J. and Koehler, Anne B.},
	month = oct,
	year = {2006},
	pages = {679--688},
}

@misc{jiang_libcity_2024,
    title = {{LibCity}: {A} {Unified} {Library} {Towards} {Efficient} and {Comprehensive} {Urban} {Spatial}-{Temporal} {Prediction}},
    shorttitle = {{LibCity}},
    url = {http://arxiv.org/abs/2304.14343},
    doi = {10.48550/arXiv.2304.14343},
    urldate = {2026-08-06},
    publisher = {arXiv},
    author = {Jiang, Jiawei and Han, Chengkai and Jiang, Wenjun and Zhao, Wayne Xin and Wang, Jingyuan},
    month = mar,
    year = {2024},
}

@misc{kaas_probabilistic_2026,
    title = {Probabilistic {Low}-{Voltage} {Peak} {Load} {Forecasting} with {Time} {Series} {Foundation} {Models} {Evaluated} on {Application}-{Oriented} {Metrics}},
    url = {http://arxiv.org/abs/2607.01966},
    doi = {10.48550/arXiv.2607.01966},
    urldate = {2026-08-18},
    publisher = {arXiv},
    author = {Kaas, Benedikt and Treutlein, Manuel and Gerber, Hannes Benedikt and Neumann, Oliver and Phatthanakhuha, Cheewan and Resch, Oliver and Mikut, Ralf and Hagenmeyer, Veit},
    month = jul,
    year = {2026},
}

@article{koenker_regression_1978,
    title = {Regression {Quantiles}},
    volume = {46},
    url = {https://doi.org/10.2307/1913643},
    number = {1},
    urldate = {2026-08-07},
    journal = {Econometrica},
    publisher = {[Wiley, Econometric Society]},
    author = {Koenker, Roger and Bassett, Gilbert},
    year = {1978},
    pages = {33--50},
}

@inproceedings{kreusel_hertel_2026,
	address = {New York, NY, USA},
	series = {{ACM} {Sustainability} {Week} '26},
	title = {Covariates {Are} the {Key} to {Accurate} {Probabilistic} {Building} {Energy} {Forecasting} with {Time} {Series} {Foundation} {Models}},
	url = {https://doi.org/10.1145/3765611.3815507},
	urldate = {2026-08-04},
	booktitle = {Proceedings of the 2026 {ACM} {Sustainability} {Week}},
	publisher = {Association for Computing Machinery},
	author = {Kreusel, Alexander and Hertel, Matthias and Noskiewicz, Moritz and Zahn, Frederik and Maaß, Heiko and Mikut, Ralf and Hagenmeyer, Veit},
	month = jun,
	year = {2026},
	pages = {172--182},
}

@misc{longarini_time_2026,
    title = {Time series {Foundation} {Models} based on {Physics}-{Informed} {Synthetic} {Histories} for {Cold}-{Start} {Photovoltaic} {Forecasting}},
    url = {http://arxiv.org/abs/2606.07457},
    doi = {10.48550/arXiv.2606.07457},
    urldate = {2026-08-18},
    publisher = {arXiv},
    author = {Longarini, Lorenzo and Rongoni, Alessandro and Silenzi, Simone and Frontoni, Emanuele and Rosati, Riccardo},
    month = jun,
    year = {2026},
}

@inproceedings{lundberg_lee_2017,
	address = {Red Hook, NY, USA},
	series = {{NIPS}'17},
	title = {A unified approach to interpreting model predictions},
	url = {https://doi.org/10.48550/arXiv.1705.07874},
	urldate = {2025-08-29},
	booktitle = {Proceedings of the 31st {International} {Conference} on {Neural} {Information} {Processing} {Systems}},
	publisher = {Curran Associates Inc.},
	author = {Lundberg, Scott M. and Lee, Su-In},
	month = dec,
	year = {2017},
	pages = {4768--4777},
}

@article{meisenbacher_review_2022,
    title = {Review of automated time series forecasting pipelines},
    volume = {12},
    copyright = {© 2022 The Authors. WIREs Data Mining and Knowledge Discovery published by Wiley Periodicals LLC.},
    url = {https://doi.org/10.1002/widm.1475},
    language = {en},
    number = {6},
    urldate = {2026-08-14},
    journal = {WIREs Data Mining and Knowledge Discovery},
    author = {Meisenbacher, Stefan and Turowski, Marian and Phipps, Kaleb and Rätz, Martin and Müller, Dirk and Hagenmeyer, Veit and Mikut, Ralf},
    year = {2022},
    pages = {e1475},
}

@misc{meyer_kaltenpoth_2026,
    title = {Rethinking {Evaluation} in the {Era} of {Time} {Series} {Foundation} {Models}: ({Un})known {Information} {Leakage} {Challenges}},
    shorttitle = {Rethinking {Evaluation} in the {Era} of {Time} {Series} {Foundation} {Models}},
    url = {http://arxiv.org/abs/2510.13654},
    doi = {10.48550/arXiv.2510.13654},
    urldate = {2026-08-14},
    publisher = {arXiv},
    author = {Meyer, Marcel and Kaltenpoth, Sascha and Zalipski, Kevin and Müller, Oliver},
    month = feb,
    year = {2026},
}

@article{micci-barreca_preprocessing_2001,
    title = {A preprocessing scheme for high-cardinality categorical attributes in classification and prediction problems},
    volume = {3},
    url = {https://doi.org/10.1145/507533.507538},
    language = {en},
    number = {1},
    urldate = {2026-08-17},
    journal = {ACM SIGKDD Explorations Newsletter},
    author = {Micci-Barreca, Daniele},
    month = jul,
    year = {2001},
    pages = {27--32},
}

@dataset{mta_subway_hourly_ridership_2020_2024,
  author       = {{Metropolitan Transportation Authority}},
  title        = {{MTA Subway Hourly Ridership: 2020--2024} [Dataset wujg-7c2s]},
  year         = {2024},
  publisher    = {NY Open Data},
  url          = {https://data.ny.gov/Transportation/MTA-Subway-Hourly-Ridership-2020-2024/wujg-7c2s/about_data},
  note         = {Retrieved via Socrata API}
}

@dataset{mta_subway_hourly_ridership_2025,
  author       = {{Metropolitan Transportation Authority}},
  title        = {{MTA Subway Hourly Ridership: Beginning 2025 [Dataset 5wq4-mkjj]}},
  year         = {2025},
  publisher    = {NY Open Data},
  url          = {https://data.ny.gov/Transportation/MTA-Subway-Hourly-Ridership-Beginning-2025/5wq4-mkjj/about_data},
  note         = {Retrieved via Socrata API}
}

@article{obermeier_pruckner_2026,
    title = {{FETS} Benchmark: Foundation Models Enable Scalable and Generalizable Energy Time Series Forecasting},
    shorttitle = {{FETS} Benchmark},
    journal = {Energy and AI},
    author = {Obermeier, Marco and Pruckner, Marco and Haselbeck, Florian and Zeiselmair, Andreas},
    year = {2026},
    volume = {26},
    pages = {100867},
    url = {https://doi.org/10.1016/j.egyai.2026.100867},
}

@misc{pan_evaluating_2026,
    title = {Evaluating {Time} {Series} {Foundation} {Models} for {Electricity} {Price} {Forecasting}: {Contamination} {Risk}, {Distributional} {Shifts}, and {Covariate} {Dependence}},
    shorttitle = {Evaluating {Time} {Series} {Foundation} {Models} for {Electricity} {Price} {Forecasting}},
    url = {http://arxiv.org/abs/2607.02623},
    doi = {10.48550/arXiv.2607.02623},
    urldate = {2026-08-14},
    publisher = {arXiv},
    author = {Pan, Zhenghua and Ezzat, Ahmed Aziz},
    month = jul,
    year = {2026},
}

@article{pargent_regularized_2022,
    title = {Regularized target encoding outperforms traditional methods in supervised machine learning with high cardinality features},
    volume = {37},
    url = {http://arxiv.org/abs/2104.00629},
    doi = {10.1007/s00180-022-01207-6},
    number = {5},
    urldate = {2026-08-17},
    journal = {Computational Statistics},
    author = {Pargent, Florian and Pfisterer, Florian and Thomas, Janek and Bischl, Bernd},
    month = nov,
    year = {2022},
    pages = {2671--2692},
}

@misc{podest_pichler_2026,
	title = {{TiRex}-2: {Generalizing} {TiRex} to {Multivariate} {Data} and {Streaming}},
	shorttitle = {{TiRex}-2},
	url = {http://arxiv.org/abs/2607.01204},
	doi = {10.48550/arXiv.2607.01204},
	urldate = {2026-08-04},
	publisher = {arXiv},
	author = {Podest, Patrick and Pichler, Marco and Bürger, Elias and Zólyomi, Levente and Voggenberger, Bernhard and Berghammer, Wilhelm and Klotz, Daniel and Böck, Sebastian and Klambauer, Günter and Hochreiter, Sepp},
	month = jul,
	year = {2026},
}

@misc{qin_cora_2025,
    title = {{CoRA}: {Covariate}-{Aware} {Adaptation} of {Time} {Series} {Foundation} {Models}},
    shorttitle = {{CoRA}},
    url = {http://arxiv.org/abs/2510.12681},
    doi = {10.48550/arXiv.2510.12681},
    urldate = {2025-12-23},
    publisher = {arXiv},
    author = {Qin, Guo and Chen, Zhi and Liu, Yong and Shi, Zhiyuan and Liu, Haixuan and Huang, Xiangdong and Wang, Jianmin and Long, Mingsheng},
    month = oct,
    year = {2025},
}

@misc{shchur_fev-bench_2026,
    title = {fev-bench: {A} {Realistic} {Benchmark} for {Time} {Series} {Forecasting}},
    shorttitle = {fev-bench},
    url = {http://arxiv.org/abs/2509.26468},
    doi = {10.48550/arXiv.2509.26468},
    urldate = {2026-08-04},
    publisher = {arXiv},
    author = {Shchur, Oleksandr and Ansari, Abdul Fatir and Turkmen, Caner and Stella, Lorenzo and Erickson, Nick and Guerron, Pablo and Bohlke-Schneider, Michael and Wang, Yuyang},
    month = jun,
    year = {2026},
}

@software{timesfm3_2026,
  author = {{Google Research}},
  title = {{TimesFM 3.0: A Zero-Shot Foundation Model for Multivariate Forecasting}},
  month = {August},
  year = {2026},
  publisher = {Hugging Face / GitHub},
  url = {https://github.com/google-research/timesfm}
}

@article{treutlein_real-world_2026,
    title = {Real-world energy data of 200 feeders from low-voltage grids with metadata in {Germany} over two years},
    journal = {Scientific Data},
    author = {Treutlein, Manuel and Bothe, Pascal and Schmidt, Marc and Hahn, Roman and Neumann, Oliver and Gielnik, Frederik and Mikut, Ralf and Hagenmeyer, Veit},
    year = {2026},
    volume = {13},
    number = {1},
    pages = {1283},
    url = {https://doi.org/10.1038/s41597-026-08152-9},
}

@article{wagner_short-_2022,
    title = {Short- and long-term forecasting of electricity prices using embedding of calendar information in neural networks},
    volume = {28},
    url = {https://doi.org/10.1016/j.jcomm.2022.100246},
    language = {en},
    urldate = {2026-08-14},
    journal = {Journal of Commodity Markets},
    author = {Wagner, Andreas and Ramentol, Enislay and Schirra, Florian and Michaeli, Hendrik},
    month = dec,
    year = {2022},
    pages = {100246},
}

@article{ziel_2018a,
    title = {Modeling public holidays in load forecasting: a {German} case study},
    volume = {6},
    shorttitle = {Modeling public holidays in load forecasting},
    url = {https://doi.org/10.1007/s40565-018-0385-5},
    journal = {Journal of Modern Power Systems and Clean Energy},
    author = {Ziel, Florian},
    month = feb,
    year = {2018},
    pages = {191--207},
}

@inproceedings{liang_2024, author = {Liang, Yuxuan and Wen, Haomin and Nie, Yuqi and Jiang, Yushan and Jin, Ming and Song, Dongjin and Pan, Shirui and Wen, Qingsong}, title = {{Foundation Models for Time Series Analysis: A Tutorial and Survey}}, year = {2024}, isbn = {9798400704901}, publisher = {Association for Computing Machinery}, address = {New York, NY, USA}, url = {https://doi.org/10.1145/3637528.3671451}, booktitle = {Proceedings of the 30th ACM SIGKDD Conference on Knowledge Discovery and Data Mining}, pages = {6555–6565}, numpages = {11}, location = {Barcelona, Spain}, series = {KDD '24} }

@dataset{mta_bus_hourly_ridership_2020_2024,
  author       = {{Metropolitan Transportation Authority}},
  title        = {{MTA Bus Hourly Ridership: 2020--2024} [Dataset wujg-7c2s]},
  year         = {2024},
  publisher    = {NY Open Data},
  url          = {https://data.ny.gov/Transportation/MTA-Bus-Hourly-Ridership-2020-2024/kv7t-n8in/about_data},
  note         = {Retrieved via Socrata API}
}

@dataset{mta_bus_hourly_ridership_2025,
  author       = {{Metropolitan Transportation Authority}},
  title        = {{MTA Bus Hourly Ridership: Beginning 2025 [Dataset 5wq4-mkjj]}},
  year         = {2025},
  publisher    = {NY Open Data},
  url          = {https://data.ny.gov/Transportation/MTA-Bus-Hourly-Ridership-Beginning-2025/gxb3-akrn/about_data},
  note         = {Retrieved via Socrata API}
}
\bibliographystyle{plainnat}

\clearpage

\appendix
\section{Appendix}
    This section provides further information on our methodology, experimental design and the results. It is structured as follows: Section \ref{sec:Appendix_Encodings} explains and visualizes all encoding strategies, and provides further details on the applied SHAP method. Section \ref{sec:App_ExperimentalSetup} contains detailed information on the datasets, models, evaluation metrics, and the hardware and software used in this study. Section \ref{sec:App_Results} contains the dataset specific results of the experiments of the main part of the study, as well as the results of a second experiment run with context length 2048.
\subsection{Methodology}
\subsubsection{Encodings}
\label{sec:Appendix_Encodings}
\paragraph{\encOrdinal}
\begin{figure}[H]
    \centering
    \includegraphics[width=0.9\linewidth]{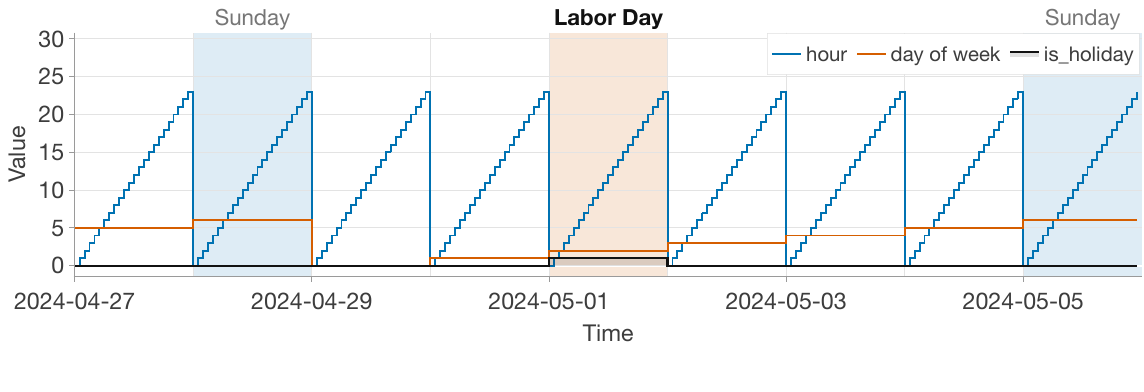}
    \caption{The \encOrdinal{} encoding with its three resulting covariates.}
    \label{fig:Ordinal_holbin}
\end{figure}
The \encOrdinal{} encoding represents hours using values from 0 to 23 and weekdays using values from 0 to 6. Holidays are binary encoded, resulting in three covariates in total. Figure \ref{fig:Ordinal_holbin} illustrates all three covariates and their behavior during weekdays and holidays.
\paragraph{\encOnehot}
\begin{figure}[H]
    \centering
    \includegraphics[width=0.9\linewidth]{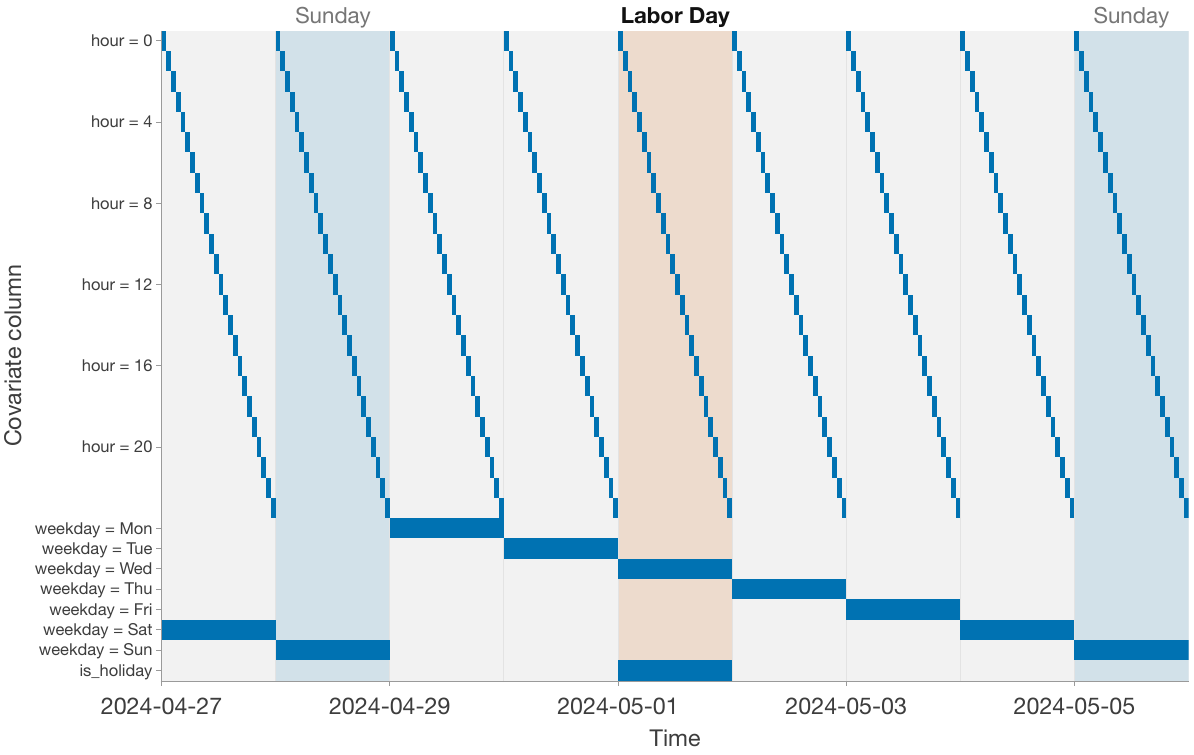}
    \caption{The \encOnehot{} encoding with its 32 resulting covariates.}
    \label{fig:onehot_holbin}
\end{figure}

The \encOnehot{} encoding represents hour and weekday using a one-hot encoding. Holidays are binary encoded, resulting in 32 covariates in total. Figure \ref{fig:onehot_holbin} illustrates all 32 covariates and their behavior during weekdays and holidays.

\paragraph{\encCyclic}
\begin{figure}[H]
    \centering
    \includegraphics[width=1.0\linewidth]{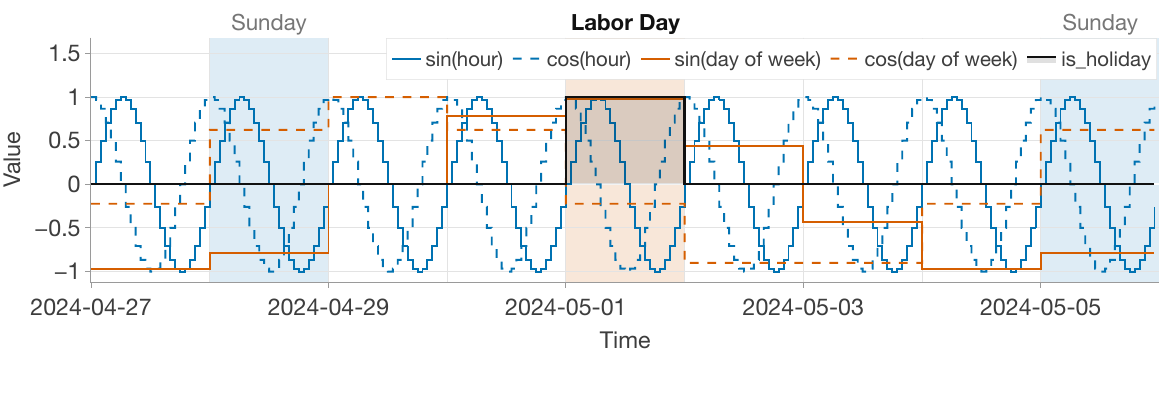}
    \caption{The \encCyclic{} encoding with its five resulting covariates.}
    \label{fig:cyclic_holbin}
\end{figure}

The \encCyclic{} encoding represents hour and weekday using sine and cosine functions, while holidays are binary encoded, resulting in five covariates in total. Figure \ref{fig:cyclic_holbin} illustrates all five covariates and their behavior on holidays. This encoding strategy is very similar to the fixed encoding applied in the \texttt{FETS} benchmark \citep{obermeier_pruckner_2026}, where minute, hour, weekday, day, and month are encoded with sine and cosine functions, and holidays are encoded binary. 

\paragraph{\encTcombo}
\begin{figure}[H]
    \centering
    \includegraphics[width=1.0\linewidth]{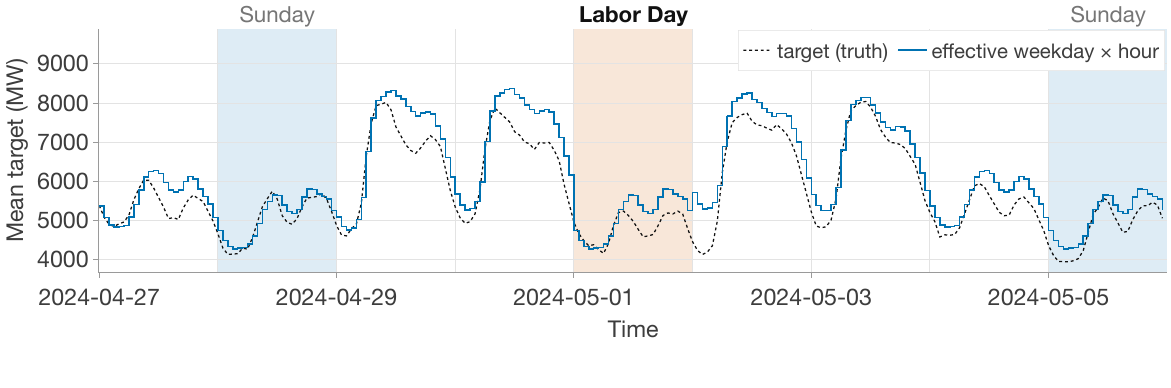}
    \caption{The \encTcombo{} target encoding using the categories hour and weekday, with holidays mapped to Sunday.}
    \label{fig:target_combo}
\end{figure}

The \encTcombo{} encoding is a target encoding with the categories hour (0--23) and weekday (0--6), where holidays are mapped to Sunday. Accordingly, the resulting encoding for Sundays and holidays in Figure \ref{fig:target_combo} is identical.
\clearpage

\paragraph{\encTholEight}
\begin{figure}[H]
    \centering
    \includegraphics[width=1.0\linewidth]{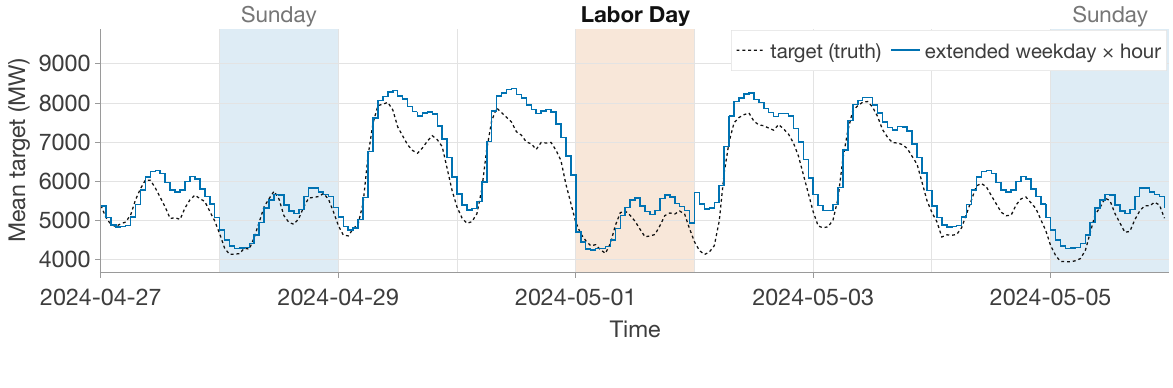}
    \caption{The \encTholEight{} target encoding using the categories hour and weekday, with holidays treated as a separate weekday.}
    \label{fig:target_hol8}
\end{figure}

The \encTholEight{} encoding is a target encoding with the categories hour and weekday. While hour is encoded as 0–23, as in \encTcombo{}, holidays are treated as a separate weekday category, so weekday is encoded as 0 to 7. Accordingly, different encoding profiles are formed for Sundays and holidays, as can be seen in Figure \ref{fig:target_hol8}.

\paragraph{\encTholAll}
\begin{figure}[H]
    \centering
    \includegraphics[width=1.0\linewidth]{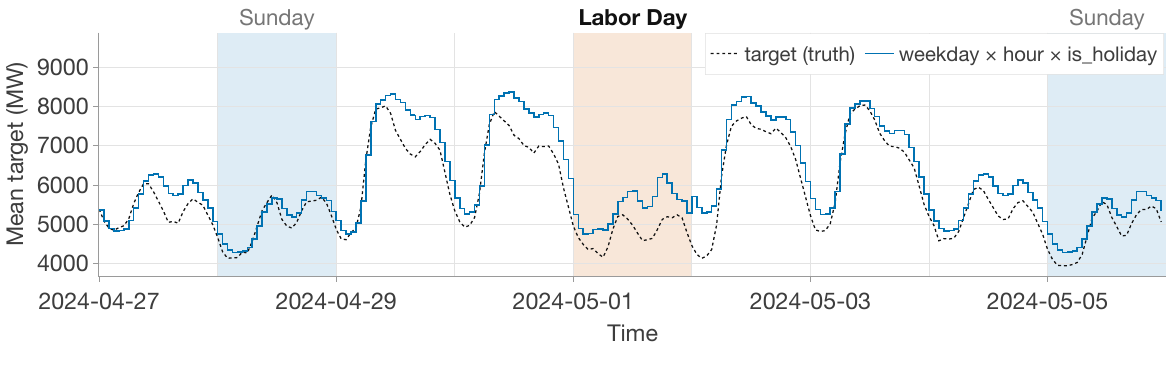}
    \caption{The \encTholAll{} target encoding using the categories hour, weekday and holiday.}
    \label{fig:target_holall}
\end{figure}

The \encTholAll{} encoding (cf. Figure \ref{fig:target_holall}) is a target encoding with the categories hour (0--23), weekday (0--6), and holiday (0 or 1). Treating holidays as a separate category leads to separate holiday encoding profiles for each weekday, e.g., the encoding for a holiday that falls on a Monday is exclusively calculated from past holidays that also fell on a Monday. This target encoding can potentially suffer from data scarcity, since a given weekday–holiday combination may occur only a few times, or not at all in the context.
\clearpage
\paragraph{\encTholPartial}
\begin{figure}[H]
    \centering
    \includegraphics[width=1.0\linewidth]{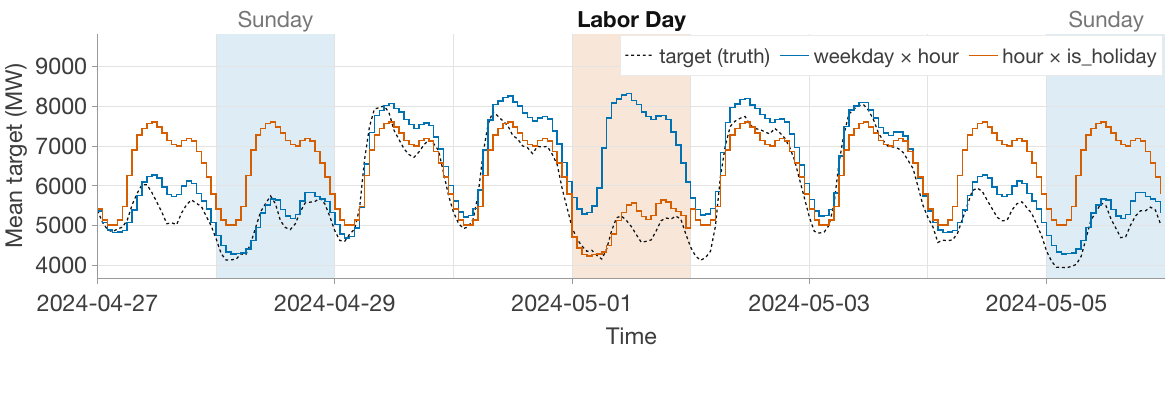}
    \caption{The \encTholPartial{} encoding consisting of two separate target encodings.}
    \label{fig:target_holpartial}
\end{figure}

The \encTholPartial{} encoding consists of two separate target encodings, both of which are used as model inputs. \encThW{} uses the categories hour (0--23) and weekday (0--6), while \encThH{} uses the categories hour (0--23) and holiday (0 or 1). Figure~\ref{fig:target_holpartial} shows the resulting encodings. By combining both covariates, the \encTholPartial encoding contains the same weekday and holiday information as all the other encoding strategies.

\subsubsection{SHAP Methodology}
\label{sec:App_Shap}

SHAP values are computed for the $T_{h,W}(H{=}7)$ and $T_{h,W} + T_{h,H}$ encodings, for the median point forecast of each horizon step and forecast window. Figure~\ref{fig:Shap_holpartial} shows these per-step values. Figure~\ref{fig:shap_pies} reports mean absolute SHAP values averaged over horizon steps, forecast windows, and datasets, separately for holidays and non-holidays. For the SHAP analyses, we directly call the native model APIs, bypassing \texttt{fev.ForecastingModel} to enable feature-subset masking.

\paragraph{Feature groups}
As in \citet{hertel_nikoltchovska_2026a}, the past target is split into four temporal groups. Together with the covariate groups, this yields $M=5$ groups for $T_{h,W}(H{=}7)$ and $M=6$ for $T_{h,W} + T_{h,H}$. All $2^M$ coalitions, i.e., 32 or 64 per forecast, are evaluated, so we obtain exact SHAP values. For clarity, the figures show the summed contribution of the temporal groups, which reduces them to two or three groups.

\paragraph{Masking}
For each coalition, the context starts at the earliest active temporal group, inactive target values are set to NaN, and inactive covariates are removed. Without any active temporal group, the context mean serves as the base prediction $f(\emptyset)$. The target encodings are computed once from the full context and are not recomputed under masking. Their Shapley values therefore capture the full contribution of the encoding as provided to the model, including the target statistics it summarizes. This is the quantity of interest, as it shows to what extent a model relies on the encoded calendar profile instead of the past target values.

\paragraph{Model-specific masking}
Chronos-2 and TabPFN-TS-3 are handled as in \citet{hertel_nikoltchovska_2026a}. Since TiRex-2 also handles NaN values natively, it is treated in the same way as Chronos-2. TimesFM-3 would fill NaN values with zeros or interpolate across them, which leaks masked information. We therefore pass the masked positions to its internal target mask.

\clearpage
\subsection{Experimental Setup}
\label{sec:App_ExperimentalSetup}

\subsubsection{Models and Pipeline}
\label{sec:App_models}
\paragraph{Chronos-2}
\citep{ansari_shchur_2025a} is a 120M-parameter TSFM based on an encoder-only Transformer architecture. It is pretrained on large-scale real and synthetic time-series data and models temporal and cross-series dependencies through temporal and group attention. It produces multi-step quantile forecasts, with the median used as point forecast.
\paragraph{TabPFN-TS-3}
\citep{grinsztajn_floge_2026} is a 58M-parameter checkpoint of TabPFN-3, trained exclusively on synthetic data. Following \citet{hoo_muller_2026a}, it represents each time step as a row of temporal features, with optional future known covariates. It predicts the full forecast horizon in a single pass, outputting a discretized predictive distribution from which quantiles are derived, with the median used as point forecast.
\paragraph{TiRex-2}
\citep{podest_pichler_2026} is an 82M-parameter TSFM based on a recurrent xLSTM architecture that extends TiRex \citep{auer_tirex_2025} to multivariate targets with past and future known covariates. It is pretrained on real and synthetic univariate data and models temporal and cross-variate dependencies through alternating time and variate mixing blocks. It produces multi-step quantile forecasts, with the median used as point forecast.

\paragraph{TimesFM-3} \citep{timesfm3_2026} is a 330M-parameter TSFM based on a decoder-only transformer architecture that supports multivariate targets with past and future known covariates. It is pretrained on a corpus of real and synthetic time series comprising more than 1T time points. Inputs are split into patches of 32 time steps, and future known covariates use lookahead tokens that concatenate the current patch with future patches. The model captures temporal and cross-variate dependencies through alternating causal time attention and full variate attention layers. The model predicts the full forecast horizon in a single pass and outputs quantiles, with the median used as point forecast.

\paragraph{Baseline Models}
We use two baseline models to assess the forecasting performance of the TSFMs. The first is a non-holiday-aware seasonal naive model with a seasonal period of 168 hours. The second is the target encoding \encTcombo{} itself (cf. Table~\ref{tab:calendar_encodings}), recomputed for each forecast window based on a context of 8192 observations. Both baselines produce quantile forecasts as a Gaussian band around their point forecast, whose width is given by the standard deviation of residuals over all context timesteps. For the seasonal naive model, these are the residuals of each observation against the observation exactly one season (168 hours) earlier. For the target-encoding baseline, they are the residuals of each observation against its timestep-matched encoding value, pooled across all day-type/hour cells present in the context.

\paragraph{Pipeline}
All covariates are provided in numeric form (int64 or float64) as \texttt{fev}-dataset. No column is passed as a string, boolean or categorical dtype, so no model applies a dtype-triggered encoding of its own. To the best of the authors knowledge, two models nevertheless branch on the covariate values: TabPFN-TS-3 infers categorical features from cardinality and re-encodes them internally, and the TiRex-2 checkpoint used here detects binary variates and bypasses its instance scaling for them. Both branches are triggered by exactly the binary indicator columns, i.e. the one-hot encoding and the binary holiday flags; the ordinal, cyclic and target encodings are unaffected in all four models.

\subsubsection{Datasets%
}
\label{sec:App_datasets}
\begin{table*}[h!]
\centering
\footnotesize
\caption{Overview of the benchmark datasets used in this work.}
\vspace{\baselineskip}

\begin{tabular}{p{1.6cm}p{1.1cm}p{4.2cm}lp{3.5cm}}
\toprule
\textbf{Name} & \textbf{Category} & \textbf{Description} & \textbf{Test Period} & \textbf{Reference} \\
\midrule

\texttt{m\_dense} &
Traffic &
Madrid traffic intensity aggregated over 30 sensors. &
\makecell{2019-01 --\\2019-12} & \cite{jiang_libcity_2024}
\\

\texttt{mta\_subway} &
Transport &
New York City subway ridership aggregated over all stations. &
\makecell{2025-01 --\\ 2025-12} & \cite{mta_subway_hourly_ridership_2020_2024, mta_subway_hourly_ridership_2025}
\\

\texttt{mta\_bus} &
Transport & 
New York City bus ridership aggregated over all stations. &
\makecell{2025-01 --\\ 2025-12} & \cite{mta_bus_hourly_ridership_2020_2024, mta_bus_hourly_ridership_2025}
\\

\texttt{TransnetBW} &
Energy &
Total load of the TransnetBW control area in Germany. &
\makecell{2025-01 --\\ 2025-12} & \cite{entsoe2026transparency}
\\

\texttt{FeederBW} &
Energy &
Active power of three individual low-voltage feeders (\#033, \#035 and \#091) from German distribution system operator NetzeBW.&
\makecell{2024-04 --\\2025-03} & \cite{treutlein_real-world_2026}

\\
\texttt{Alicante} &
Water &
Average water consumption of 1,007 consumers in Alicante.&
\makecell{2016-01 --\\2016-12} & \cite{daiad_alicante_swm_2026}

\\
\texttt{uci\_bike} &
Transport &
Aggregated bikesharing rental counts in Washington, DC.&
\makecell{2012-01 --\\2012-12} & \cite{fanaee-t_2013}
\\
\bottomrule
\end{tabular}

\label{tab:datasets}
\end{table*}

Table~\ref{tab:datasets} summarizes all datasets used in this study. The test period of each dataset spans exactly one year%
. While the \texttt{m\_dense}, \texttt{mta\_subway} and \texttt{mta\_bus} datasets natively have an hourly resolution, the \texttt{TransnetBW} and \texttt{FeederBW} sets originally have higher frequencies
and are resampled to hourly resolution using the mean. From the \texttt{FeederBW} dataset, the three selected series are evaluated separately and referred to as \texttt{FeederBW\_33}, \texttt{FeederBW\_35}, and \texttt{FeederBW\_91}. None of the datasets used in this study are included in the pretraining corpora of the employed TSFMs, excluding train-test leakage \citep{meyer_kaltenpoth_2026}.
We evaluate all models in zero-shot setting, using a rolling evaluation across all datasets and encoding strategies. 

\subsubsection{Evaluation Metrics}
\label{sec:App_metrics}
MASE and SQL are implemented in the \texttt{fev-bench} evaluation library \citep{shchur_fev-bench_2026}, but are outlined here consistent with our notation.

\paragraph{Mean Absolute Scaled Error}

\begin{equation}
\operatorname{MASE}
=
\frac{1}{N}
\sum_{i=1}^{N}
\left(
\frac{
\dfrac{1}{f}
\sum_{h=1}^{f}
\left|
Y_{i,c+h}
-
\hat{Y}_{i,c+h}
\right|
}{
\dfrac{1}{c-m}
\sum_{s=m+1}^{c}
\left|
Y_{i,s}
-
Y_{i,s-m}
\right|
}
\right).
\label{eq:MASE}
\end{equation}
The mean absolute scaled error (MASE) is a point forecast metric that scales
forecast errors by the in-sample MAE of a seasonal naïve forecast, making it
scale-independent. MASE is defined as in Equation~\ref{eq:MASE}, where $N$
denotes the number of prediction windows, $f$ the length of the forecast
horizon $\mathcal{F}=(c+1,\ldots,c+f)$, $Y_{i,c+h}$ the observed value at
forecast step $h \in \{1,\ldots,f\}$ of prediction window $i$, and
$\hat{Y}_{i,c+h}$ the corresponding forecast. Furthermore, $c$ denotes the
length of the context window $\mathcal{C}=(1,\ldots,c)$, $m$ the seasonal
period, and $Y_{i,s}$ the observed value at context position $s \in \mathcal{C}$
of window $i$. In this study, the seasonal period is set to $m = 168$ hours,
and the seasonal error is re-computed for each context window \citep{hyndman_koehler_2006}. 

\paragraph{Scaled Quantile Loss}

\begin{equation}
\operatorname{SQL}
=
\frac{1}{N}
\sum_{i=1}^{N}
\left(
\frac{
\dfrac{1}{|\mathcal{Q}|\,f}
\sum_{\tau \in \mathcal{Q}}
\sum_{h=1}^{f}
\rho_{\tau}
\!\left(
Y_{i,c+h}
-
\hat{Y}_{i,c+h}^{(\tau)}
\right)
}{
\dfrac{1}{c-m}
\sum_{s=m+1}^{c}
\left|
Y_{i,s}
-
Y_{i,s-m}
\right|
}
\right).
\label{eq:SQL}
\end{equation}

We report the scaled quantile loss (SQL), which averages the pinball loss
\citep{koenker_regression_1978} over the quantile grid
$\mathcal{Q} = \{0.1, 0.2, \ldots, 0.9\}$ and scales it by the in-sample
seasonal-naive error \citep{hyndman_koehler_2006}, making it the probabilistic extension of MASE. SQL is defined as in Equation~\ref{eq:SQL}, where $\rho_{\tau}$ denotes the pinball loss at quantile level $\tau$.
As for MASE, we use a seasonal period of $m=168$ hours, and the seasonal-naive error is computed separately for each context window.

\subsubsection{Hardware and Software}
\paragraph{Compute environment.}
All experiments were run on a single workstation with one NVIDIA
GeForce~RTX~3090 GPU. The benchmark is implemented in
Python~3.11.15 on top of the \texttt{fev}~0.10.0 evaluation framework; all models
are wrapped as \texttt{fev.ForecastingModel} implementations so that data
loading, window construction (fixed context of 8192 hourly steps for every model, horizon 24, stride 24) and metric computation are identical across models.
The foundation models run on the single GPU using PyTorch~2.9.1;
the naive baselines are CPU-only. No distributed execution is used.

\clearpage
\subsection{Results}
\label{sec:App_Results}
This section contains further results of the experiments. Sections \ref{sec:App_results_m_dense} to \ref{sec:App_results_uci_bike} contain the dataset specific results from the experiments presented in Section \ref{sec:results}. Section \ref{sec:oracle_results} presents overall results of all models when provided with ground truth of the target as known covariate.
Section \ref{sec:App_Results2048} presents the overall results of a second experimental run conducted with context length 2048. 
\subsubsection{m\_dense}
\label{sec:App_results_m_dense}
\begin{table}[H]
\centering
\footnotesize
\caption{Results for the \texttt{m\_dense} dataset, split by date-type.}
\label{tab:eval5-m-dense}
\resizebox{\textwidth}{!}{%
\begin{tabular}{ll rr rr rr}
\toprule
& & \multicolumn{2}{c}{\textbf{All days}} & \multicolumn{2}{c}{\textbf{Holidays only}} & \multicolumn{2}{c}{\textbf{Non-holidays only}} \\
\cmidrule(lr){3-4} \cmidrule(lr){5-6} \cmidrule(lr){7-8}
\textbf{Model} & \textbf{Encoding} & \textbf{MASE} & \textbf{SQL} & \textbf{MASE} & \textbf{SQL} & \textbf{MASE} & \textbf{SQL}
 \\
\midrule
\multirow{2}{*}{Baselines}
 & seasonal-naive (168h) & 1.0106 & 0.9876 & 3.8760 & 3.4173 & \underline{0.9132} & 0.9050 \\
 & $T_{h,W}(H=Sun)_{naive}$    & \underline{0.9404} & \underline{0.8174} & \underline{1.4057} & \underline{1.1195} & 0.9246 & \underline{0.8072} \\
\midrule
\multirow{8}{*}{Chronos-2}
 & \encNone              & 0.5667 & 0.4661 & 3.2231 & 2.8733 & 0.4764 & 0.3843 \\
 & \encOrdinal           & 0.5385 & 0.4364 & 2.0874 & 1.7720 & 0.4858 & 0.3910 \\
 & \encOnehot            & 0.5715 & 0.4689 & 2.7856 & 2.5112 & 0.4962 & 0.3995 \\
 & \encCyclic            & 0.5515 & 0.4463 & 2.1623 & 1.8558 & 0.4967 & 0.3984 \\
 & \encTcombo            & 0.5042 & 0.4071 & 1.5240 & 1.2165 & 0.4695 & 0.3796 \\
 & \encTholEight         & 0.5017 & 0.4054 & \underline{1.4231} & \underline{1.1317} & 0.4704 & 0.3807 \\
 & \encTholAll           & 0.5076 & 0.4093 & 1.5639 & 1.2207 & 0.4717 & 0.3817 \\
 & \encTholPartial       & \textbf{\underline{0.4998}} & \textbf{\underline{0.4050}} & 1.5269 & 1.2333 & \textbf{\underline{0.4649}} & \textbf{\underline{0.3769}} \\
\midrule
\multirow{8}{*}{TabPFN-TS-3}
 & \encNone              & 0.8632 & 0.7158 & 3.8713 & 3.2631 & 0.7609 & 0.6292 \\
 & \encOrdinal           & 0.7757 & 0.6278 & 1.2380 & 0.9696 & 0.7600 & 0.6162 \\
 & \encOnehot            & 0.7814 & 0.6327 & 1.1471 & 0.9211 & 0.7690 & 0.6228 \\
 & \encCyclic            & 0.7352 & 0.5969 & 1.0662 & 0.8703 & 0.7240 & 0.5876 \\
 & \encTcombo            & 0.7078 & 0.5744 & \underline{1.0535} & \underline{0.8636} & 0.6960 & 0.5646 \\
 & \encTholEight         & 0.7014 & 0.5719 & 1.0815 & 0.9072 & 0.6884 & 0.5605 \\
 & \encTholAll           & 0.7188 & 0.5852 & 1.4575 & 1.2477 & 0.6937 & 0.5627 \\
 & \encTholPartial       & \underline{0.6636} & \underline{0.5421} & 1.1292 & 0.9276 & \underline{0.6477} & \underline{0.5290} \\
\midrule
\multirow{8}{*}{TiRex-2}
 & \encNone              & \underline{0.6973} & 0.5749 & 3.2580 & 2.8495 & \underline{0.6102} & \underline{0.4976} \\
 & \encOrdinal           & 0.7374 & 0.6042 & 2.9714 & 2.5300 & 0.6615 & 0.5387 \\
 & \encOnehot            & 0.7283 & 0.6018 & 3.1792 & 2.7644 & 0.6450 & 0.5283 \\
 & \encCyclic            & 0.7101 & 0.5857 & 3.2760 & 2.8551 & 0.6229 & 0.5086 \\
 & \encTcombo            & 0.7133 & 0.5618 & 1.1817 & 0.9287 & 0.6974 & 0.5494 \\
 & \encTholEight         & 0.7064 & \underline{0.5574} & \underline{1.0202} & \underline{0.8115} & 0.6958 & 0.5487 \\
 & \encTholAll           & 0.7153 & 0.5661 & 1.3063 & 1.0720 & 0.6952 & 0.5489 \\
 & \encTholPartial       & 0.7666 & 0.6073 & 2.1815 & 1.8493 & 0.7185 & 0.5651 \\
\midrule
\multirow{8}{*}{TimesFM-3}
 & \encNone              & 0.5747 & 0.4714 & 3.2189 & 2.7659 & 0.4848 & 0.3934 \\
 & \encOrdinal           & \underline{0.5219} & \underline{0.4244} & 2.0393 & 1.7248 & \underline{0.4703} & \underline{0.3802} \\
 & \encOnehot            & 0.5363 & 0.4352 & 2.3697 & 1.9911 & 0.4739 & 0.3823 \\
 & \encCyclic            & 0.5262 & 0.4289 & 2.1471 & 1.8497 & 0.4711 & 0.3806 \\
 & \encTcombo            & 0.5301 & 0.4292 & 0.9858 & 0.7779 & 0.5146 & 0.4173 \\
 & \encTholEight         & 0.5282 & 0.4287 & \textbf{\underline{0.9520}} & \textbf{\underline{0.7716}} & 0.5138 & 0.4170 \\
 & \encTholAll           & 0.5354 & 0.4349 & 1.0633 & 0.8410 & 0.5175 & 0.4211 \\
 & \encTholPartial       & 0.5720 & 0.4717 & 1.0845 & 0.8693 & 0.5546 & 0.4582 \\
\bottomrule
\end{tabular}}
\end{table}

\clearpage
\begin{figure}[H]
    \centering
    \includegraphics[width=1.0\linewidth]{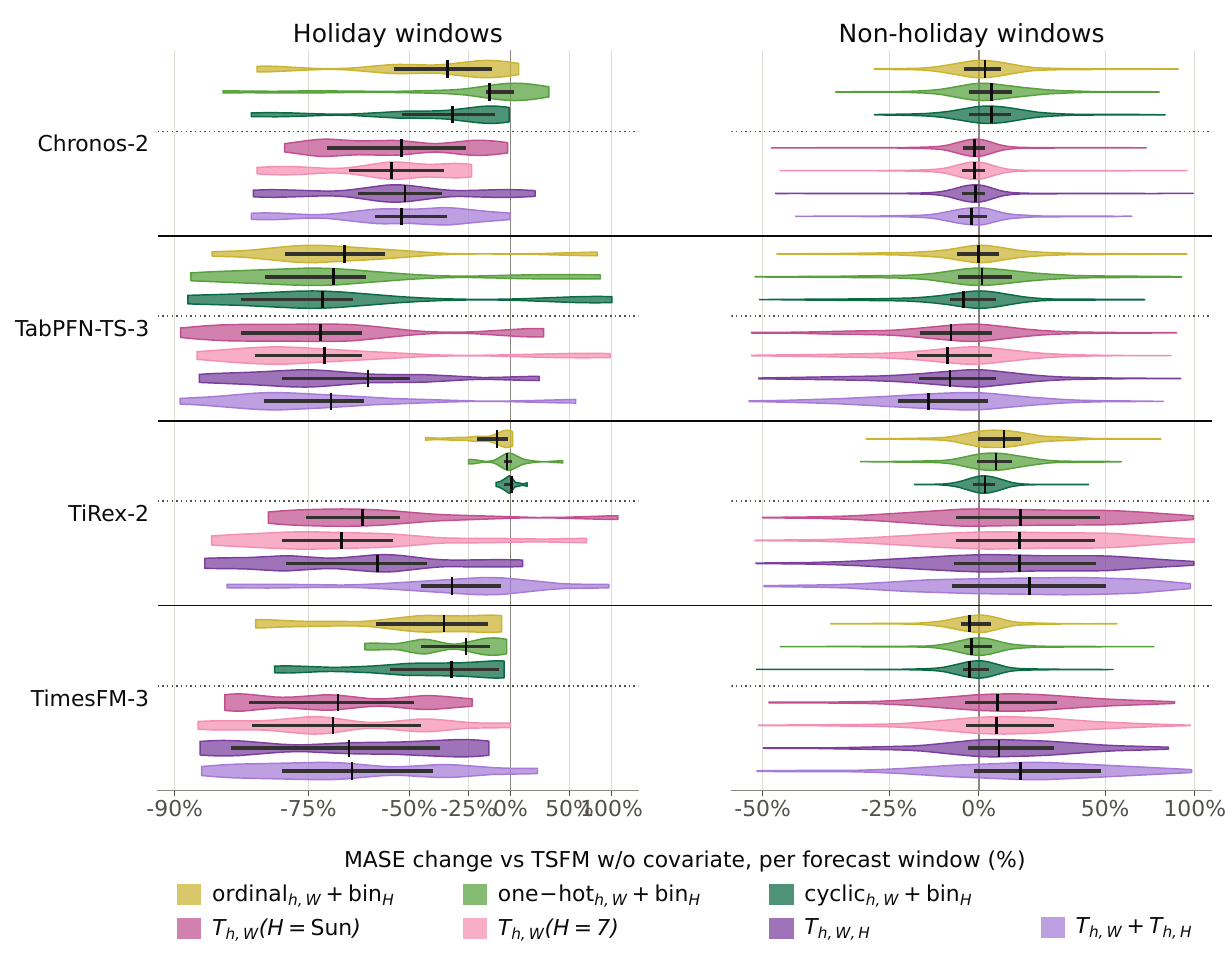}
    \caption{MASE change against the model w/o covariates on the \texttt{m\_dense} dataset.}
    \label{fig:effect_m_dense}
\end{figure}

\begin{figure}[H]
    \centering
    \includegraphics[width=1.0\linewidth]{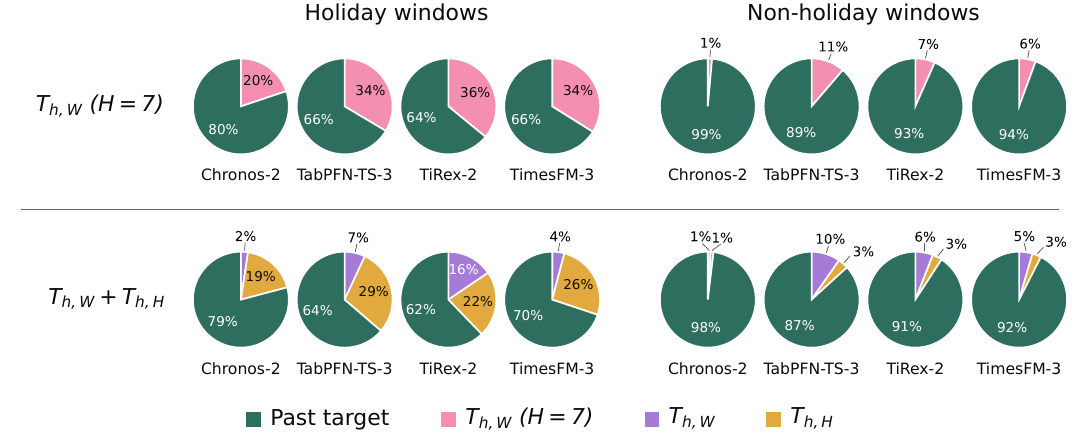}
    \caption{SHAP contributions for all models on the \texttt{m\_dense} dataset, exemplarily depicted for the \encTholEight and \encTholPartial encodings.}
    \label{fig:placeholder}
\end{figure}
\clearpage

\subsubsection{mta\_subway}
\label{sec:App_results_mtasubway}
\begin{table}[ht]
\centering
\caption{Results for the \texttt{mta\_subway} dataset, split by date-type.}
\label{tab:eval5-mta-subway-hourly}
\resizebox{\textwidth}{!}{%
\begin{tabular}{ll rr rr rr}
\toprule
& & \multicolumn{2}{c}{\textbf{All days}} & \multicolumn{2}{c}{\textbf{Holidays only}} & \multicolumn{2}{c}{\textbf{Non-holidays only}} \\
\cmidrule(lr){3-4} \cmidrule(lr){5-6} \cmidrule(lr){7-8}
\textbf{Model} & \textbf{Encoding} & \textbf{MASE} & \textbf{SQL} & \textbf{MASE} & \textbf{SQL} & \textbf{MASE} & \textbf{SQL}
 \\
\midrule
\multirow{2}{*}{Baselines}
 & seasonal-naive (168h) & \underline{1.0399} & 1.1594 & 4.3610 & 3.9833 & \underline{0.9075} & 1.0468 \\
 & $T_{h,W}(H=Sun)_{naive}$    & 1.0595 & \underline{0.9450} & \underline{2.8182} & \underline{2.5015} & 0.9893 & \underline{0.8829} \\
\midrule
\multirow{8}{*}{Chronos-2}
 & \encNone              & 0.7549 & 0.6086 & 3.2626 & 2.8593 & 0.6549 & 0.5188 \\
 & \encOrdinal           & 0.7616 & 0.6089 & 2.5726 & 2.2009 & 0.6894 & 0.5454 \\
 & \encOnehot            & 0.8271 & 0.6689 & 3.2127 & 2.8555 & 0.7320 & 0.5817 \\
 & \encCyclic            & 0.7918 & 0.6341 & 2.6988 & 2.3076 & 0.7157 & 0.5673 \\
 & \encTcombo            & 0.6907 & 0.5496 & 2.1615 & 1.6968 & \underline{0.6320} & 0.5038 \\
 & \encTholEight         & 0.6941 & 0.5516 & 2.2295 & 1.7492 & 0.6328 & \underline{0.5038} \\
 & \encTholAll           & 0.7179 & 0.5726 & 2.7617 & 2.2296 & 0.6364 & 0.5065 \\
 & \encTholPartial       & \underline{0.6886} & \underline{0.5486} & \underline{2.0881} & \underline{1.6656} & 0.6328 & 0.5041 \\
\midrule
\multirow{8}{*}{TabPFN-TS-3}
 & \encNone              & 0.9943 & 0.8066 & 4.8747 & 4.2726 & 0.8395 & 0.6683 \\
 & \encOrdinal           & 0.9069 & 0.7089 & 2.9485 & 2.1927 & 0.8255 & 0.6497 \\
 & \encOnehot            & 0.9120 & 0.7220 & 3.0588 & 2.3942 & 0.8264 & 0.6553 \\
 & \encCyclic            & 0.8554 & 0.6785 & 2.6945 & \underline{2.0599} & 0.7821 & 0.6234 \\
 & \encTcombo            & \underline{0.7623} & \underline{0.6213} & \underline{2.4610} & 2.0660 & \underline{0.6945} & 0.5636 \\
 & \encTholEight         & 0.7658 & 0.6279 & 2.5234 & 2.1769 & 0.6957 & 0.5661 \\
 & \encTholAll           & 0.8035 & 0.6562 & 3.5265 & 2.9862 & 0.6949 & \underline{0.5632} \\
 & \encTholPartial       & 0.7895 & 0.6341 & 3.0241 & 2.3532 & 0.7004 & 0.5655 \\
\midrule
\multirow{8}{*}{TiRex-2}
 & \encNone              & 0.9415 & 0.7718 & 2.5285 & 2.0535 & 0.8782 & 0.7207 \\
 & \encOrdinal           & 1.0268 & 0.8296 & \underline{2.4657} & \underline{1.9664} & 0.9694 & 0.7842 \\
 & \encOnehot            & 0.9474 & 0.7822 & 2.9214 & 2.3684 & 0.8686 & 0.7189 \\
 & \encCyclic            & 0.9446 & 0.7764 & 2.7866 & 2.2705 & 0.8711 & 0.7168 \\
 & \encTcombo            & 0.9288 & 0.7695 & 2.6489 & 2.3140 & 0.8602 & 0.7079 \\
 & \encTholEight         & \underline{0.9193} & \underline{0.7590} & 2.7073 & 2.2959 & \underline{0.8480} & \underline{0.6977} \\
 & \encTholAll           & 0.9553 & 0.7908 & 3.4280 & 2.9991 & 0.8567 & 0.7028 \\
 & \encTholPartial       & 1.0064 & 0.8266 & 3.6526 & 3.1124 & 0.9009 & 0.7354 \\
\midrule
\multirow{8}{*}{TimesFM-3}
 & \encNone              & 0.6337 & 0.5195 & 2.8536 & 2.4981 & 0.5451 & 0.4406 \\
 & \encOrdinal           & 0.6166 & 0.5008 & 2.1742 & 1.8642 & 0.5545 & 0.4464 \\
 & \encOnehot            & 0.6298 & 0.5137 & 2.5228 & 2.1905 & 0.5543 & 0.4469 \\
 & \encCyclic            & \textbf{\underline{0.5978}} & \textbf{\underline{0.4855}} & 2.0097 & 1.7212 & \textbf{\underline{0.5415}} & \textbf{\underline{0.4362}} \\
 & \encTcombo            & 0.6251 & 0.5008 & 1.6005 & 1.2729 & 0.5862 & 0.4701 \\
 & \encTholEight         & 0.6390 & 0.5122 & 1.8520 & 1.4625 & 0.5906 & 0.4743 \\
 & \encTholAll           & 0.6721 & 0.5407 & 2.4339 & 1.9817 & 0.6018 & 0.4832 \\
 & \encTholPartial       & 0.6735 & 0.5414 & \textbf{\underline{1.4698}} & \textbf{\underline{1.1666}} & 0.6418 & 0.5165 \\
\bottomrule
\end{tabular}}
\end{table}
\clearpage
\begin{figure}[H]
    \centering
    \includegraphics[width=1.0\linewidth]{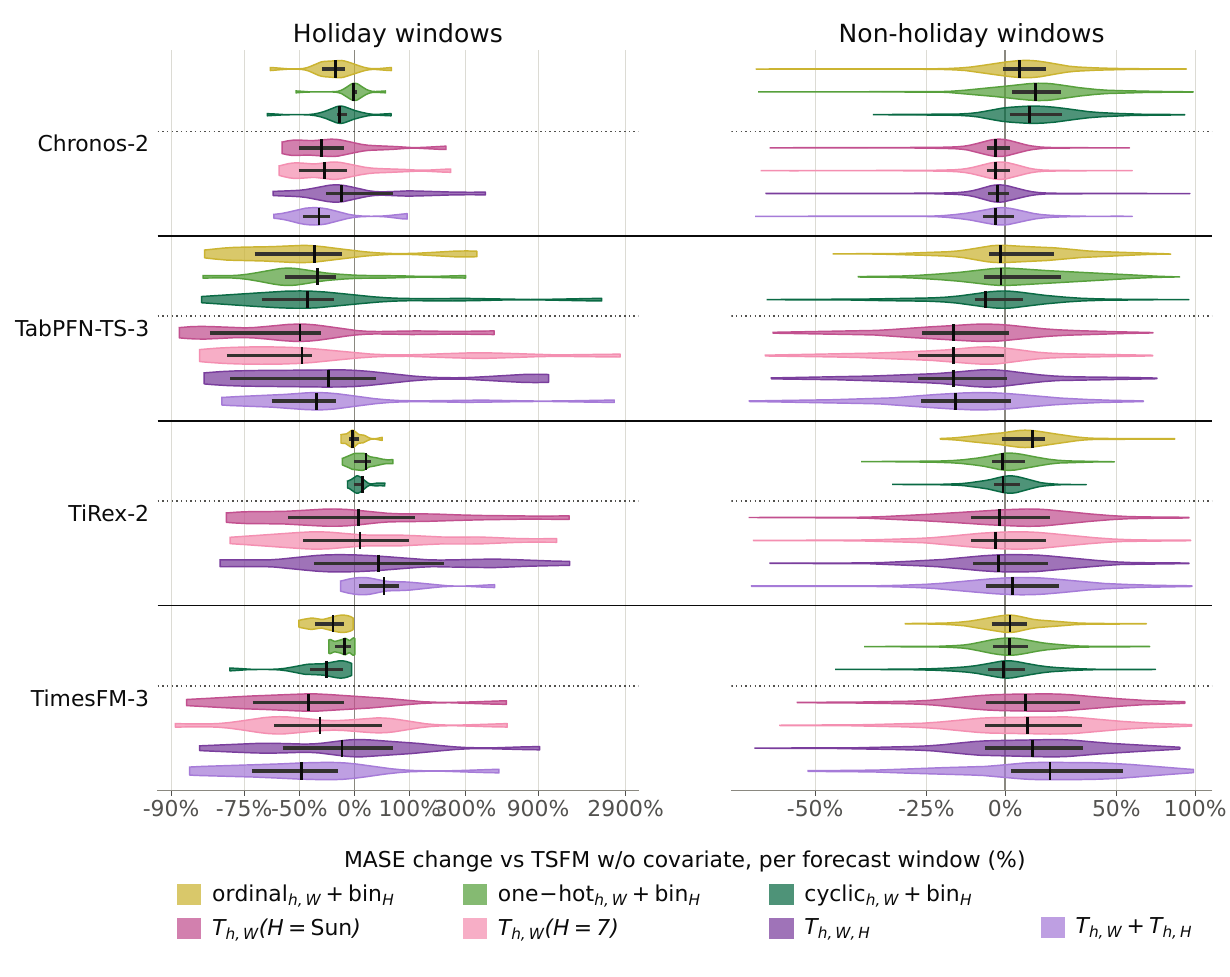}
    \caption{MASE change against the model w/o covariates on the \texttt{mta\_subway} dataset.}
    \label{fig:effect_mta_subway_hourly}
\end{figure}

\begin{figure}[H]
    \centering
    \includegraphics[width=1.0\linewidth]{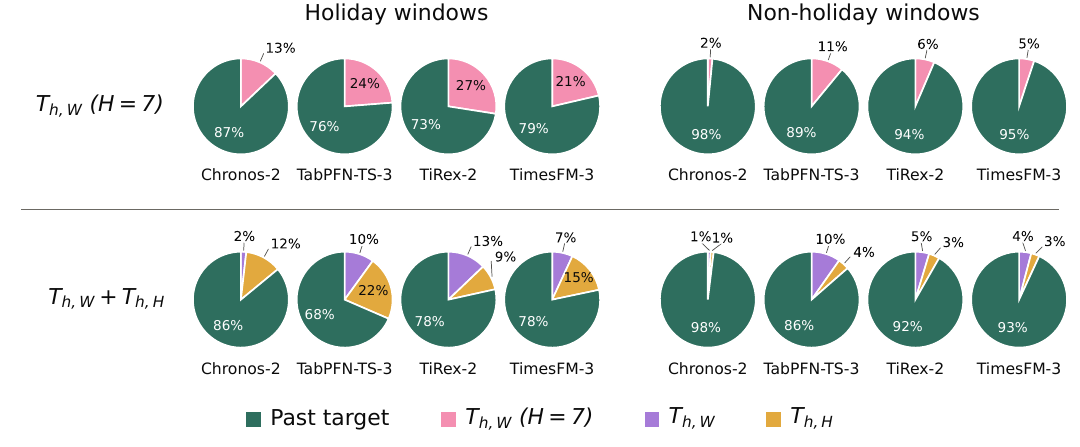}
    \caption{SHAP contributions for all models on the \texttt{mta\_subway} dataset, exemplarily depicted for the \encTholEight and \encTholPartial encodings.}
    \label{fig:placeholder}
\end{figure}
\clearpage
\subsubsection{mta\_bus}
\label{sec:App_results_mtabus}
\begin{table}[H]
\centering
\caption{Results for the \texttt{mta\_bus} dataset, split by date-type.}
\label{tab:eval5-mta-bus-hourly}
\resizebox{\textwidth}{!}{%
\begin{tabular}{ll rr rr rr}
\toprule
& & \multicolumn{2}{c}{\textbf{All days}} & \multicolumn{2}{c}{\textbf{Holidays only}} & \multicolumn{2}{c}{\textbf{Non-holidays only}} \\
\cmidrule(lr){3-4} \cmidrule(lr){5-6} \cmidrule(lr){7-8}
\textbf{Model} & \textbf{Encoding} & \textbf{MASE} & \textbf{SQL} & \textbf{MASE} & \textbf{SQL} & \textbf{MASE} & \textbf{SQL}
 \\
\midrule
\multirow{2}{*}{Baselines}
 & seasonal-naive (168h) & \underline{1.0352} & 1.1499 & 4.4528 & 4.0390 & \underline{0.8989} & 1.0346 \\
 & $T_{h,W}(H=Sun)_{naive}$    & 1.0809 & \underline{0.9837} & \underline{2.7476} & \underline{2.4613} & 1.0144 & \underline{0.9248} \\
\midrule
\multirow{8}{*}{Chronos-2}
 & \encNone              & 0.7804 & 0.6293 & 3.5664 & 3.1018 & 0.6693 & 0.5307 \\
 & \encOrdinal           & 0.7569 & 0.6019 & 2.3343 & 1.9043 & 0.6940 & 0.5499 \\
 & \encOnehot            & 0.8092 & 0.6540 & 3.4036 & 2.9105 & 0.7057 & 0.5640 \\
 & \encCyclic            & 0.7849 & 0.6278 & 2.4517 & 2.0471 & 0.7184 & 0.5712 \\
 & \encTcombo            & 0.6878 & 0.5451 & 2.1003 & 1.6142 & 0.6315 & 0.5025 \\
 & \encTholEight         & 0.6895 & 0.5464 & 2.1001 & 1.6265 & 0.6332 & 0.5033 \\
 & \encTholAll           & 0.7184 & 0.5713 & 2.7707 & 2.2042 & 0.6366 & 0.5062 \\
 & \encTholPartial       & \underline{0.6796} & \underline{0.5407} & \underline{1.9960} & \underline{1.5669} & \underline{0.6271} & \underline{0.4997} \\
\midrule
\multirow{8}{*}{TabPFN-TS-3}
 & \encNone              & 0.9435 & 0.8020 & 5.0247 & 4.2693 & 0.7807 & 0.6637 \\
 & \encOrdinal           & 0.8178 & 0.6478 & 2.6567 & 2.0060 & 0.7444 & 0.5936 \\
 & \encOnehot            & 0.8741 & 0.6854 & 3.0826 & 2.3104 & 0.7860 & 0.6205 \\
 & \encCyclic            & 0.8041 & 0.6365 & 2.5296 & \underline{1.8679} & 0.7353 & 0.5874 \\
 & \encTcombo            & \underline{0.7191} & \underline{0.5759} & \underline{2.4942} & 1.9174 & 0.6483 & 0.5224 \\
 & \encTholEight         & 0.7248 & 0.5886 & 2.5580 & 2.1131 & 0.6517 & 0.5278 \\
 & \encTholAll           & 0.7599 & 0.6211 & 3.6401 & 3.1589 & \underline{0.6451} & \underline{0.5199} \\
 & \encTholPartial       & 0.7899 & 0.6344 & 2.8972 & 2.2430 & 0.7058 & 0.5703 \\
\midrule
\multirow{8}{*}{TiRex-2}
 & \encNone              & 0.9631 & 0.7789 & 3.0336 & 2.4821 & 0.8805 & 0.7110 \\
 & \encOrdinal           & 1.0499 & 0.8321 & 2.9162 & 2.3247 & 0.9754 & 0.7726 \\
 & \encOnehot            & 0.9509 & 0.7793 & 3.2766 & 2.6941 & \underline{0.8581} & 0.7029 \\
 & \encCyclic            & \underline{0.9484} & 0.7703 & 3.1883 & 2.6192 & 0.8591 & \underline{0.6966} \\
 & \encTcombo            & 0.9519 & 0.7689 & \underline{2.5983} & 2.2903 & 0.8863 & 0.7082 \\
 & \encTholEight         & 0.9537 & \underline{0.7638} & 2.6418 & \underline{2.2633} & 0.8863 & 0.7040 \\
 & \encTholAll           & 0.9872 & 0.7945 & 3.3539 & 2.9642 & 0.8928 & 0.7079 \\
 & \encTholPartial       & 1.0447 & 0.8379 & 3.7789 & 3.2399 & 0.9356 & 0.7421 \\
\midrule
\multirow{8}{*}{TimesFM-3}
 & \encNone              & 0.6298 & 0.5188 & 3.1170 & 2.6992 & \textbf{\underline{0.5306}} & \textbf{\underline{0.4319}} \\
 & \encOrdinal           & 0.6049 & 0.4945 & 2.0956 & 1.7560 & 0.5455 & 0.4442 \\
 & \encOnehot            & 0.6210 & 0.5086 & 2.5206 & 2.1440 & 0.5452 & 0.4434 \\
 & \encCyclic            & \textbf{\underline{0.5880}} & \textbf{\underline{0.4798}} & \textbf{\underline{1.8362}} & \textbf{\underline{1.5389}} & 0.5382 & 0.4376 \\
 & \encTcombo            & 0.6303 & 0.5112 & 1.9995 & 1.5968 & 0.5757 & 0.4679 \\
 & \encTholEight         & 0.6395 & 0.5208 & 2.1952 & 1.7823 & 0.5774 & 0.4705 \\
 & \encTholAll           & 0.6690 & 0.5488 & 2.8314 & 2.4011 & 0.5827 & 0.4749 \\
 & \encTholPartial       & 0.6649 & 0.5416 & 1.9124 & 1.5393 & 0.6152 & 0.5018 \\
\bottomrule
\end{tabular}}
\end{table}
\clearpage
\begin{figure}[H]
    \centering
    \includegraphics[width=1.0\linewidth]{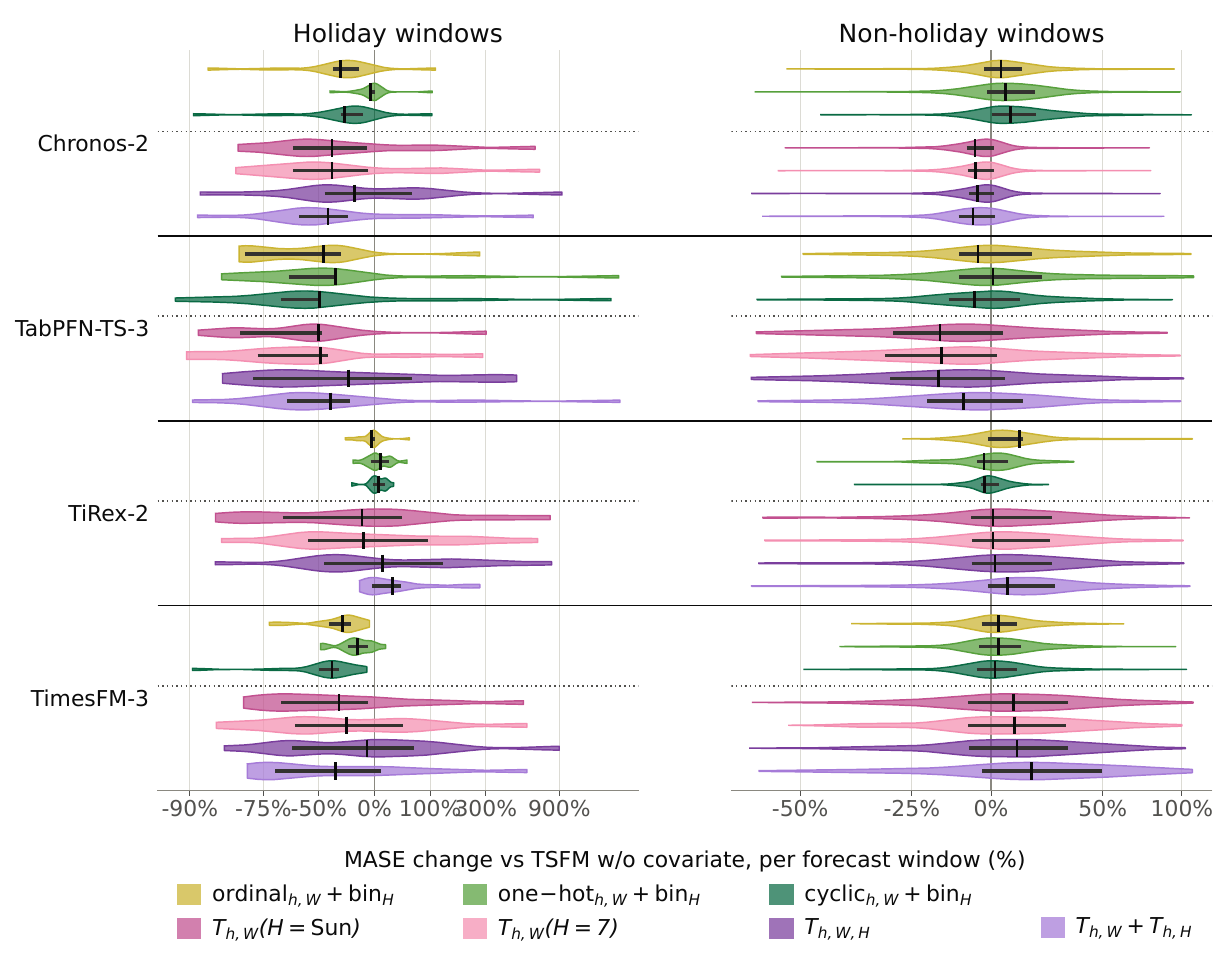}
    \caption{MASE change against the model w/o covariates on the \texttt{mta\_bus} dataset.}
    \label{fig:effect_mta_bus_hourly}
\end{figure}

\begin{figure}[H]
    \centering
    \includegraphics[width=1.0\linewidth]{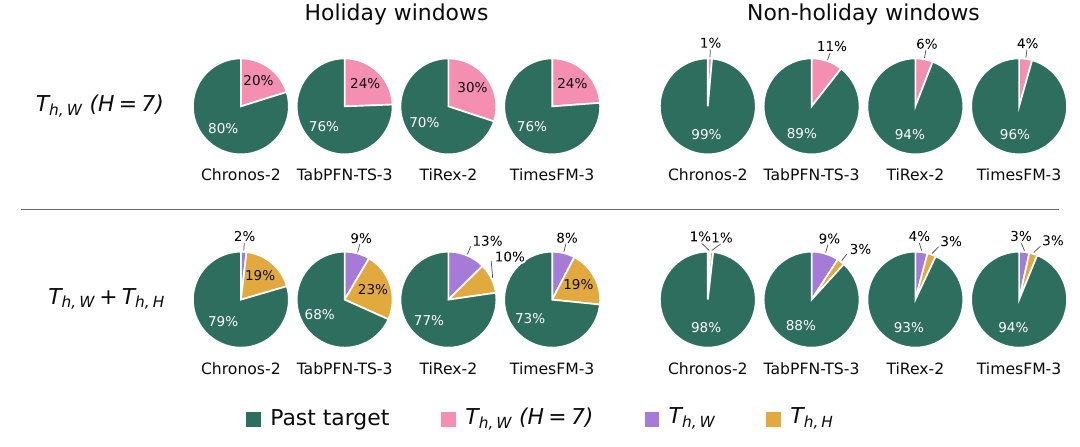}
    \caption{SHAP contributions for all models on the \texttt{mta\_bus} dataset, exemplarily depicted for the \encTholEight and \encTholPartial encodings.}
    \label{fig:placeholder}
\end{figure}

\clearpage
\subsubsection{TransnetBW}
\begin{table}[H]
\centering
\caption{Results for the \texttt{TransnetBW} dataset, split by date-type.}
\label{tab:eval5-entsoe-transnetbw-load}
\resizebox{\textwidth}{!}{%
\begin{tabular}{ll rr rr rr}
\toprule
& & \multicolumn{2}{c}{\textbf{All days}} & \multicolumn{2}{c}{\textbf{Holidays only}} & \multicolumn{2}{c}{\textbf{Non-holidays only}} \\
\cmidrule(lr){3-4} \cmidrule(lr){5-6} \cmidrule(lr){7-8}
\textbf{Model} & \textbf{Encoding} & \textbf{MASE} & \textbf{SQL} & \textbf{MASE} & \textbf{SQL} & \textbf{MASE} & \textbf{SQL}
 \\
\midrule
\multirow{2}{*}{Baselines}
 & seasonal-naive (168h) & \underline{1.0373} & \underline{0.8887} & 3.5117 & 2.9987 & \underline{0.9532} & \underline{0.8170} \\
 & $T_{h,W}(H=Sun)_{naive}$    & 1.4602 & 1.1108 & \underline{1.1270} & \underline{0.8393} & 1.4715 & 1.1200 \\
\midrule
\multirow{8}{*}{Chronos-2}
 & \encNone              & 0.4850 & 0.3885 & 2.3853 & 2.0609 & 0.4205 & 0.3316 \\
 & \encOrdinal           & 0.4284 & 0.3353 & 0.7864 & 0.6258 & 0.4162 & 0.3254 \\
 & \encOnehot            & 0.4765 & 0.3781 & 1.6213 & 1.3350 & 0.4376 & 0.3456 \\
 & \encCyclic            & 0.4310 & 0.3380 & 0.8024 & 0.6533 & 0.4184 & 0.3273 \\
 & \encTcombo            & \underline{0.4159} & \underline{0.3273} & \underline{0.5654} & \underline{0.4785} & 0.4109 & 0.3222 \\
 & \encTholEight         & 0.4165 & 0.3274 & 0.5954 & 0.4932 & \underline{0.4104} & \underline{0.3218} \\
 & \encTholAll           & 0.4192 & 0.3290 & 0.6724 & 0.5394 & 0.4106 & 0.3219 \\
 & \encTholPartial       & 0.4190 & 0.3296 & 0.6634 & 0.5384 & 0.4107 & 0.3225 \\
\midrule
\multirow{8}{*}{TabPFN-TS-3}
 & \encNone              & 0.7261 & 0.5740 & 2.8440 & 2.4972 & 0.6541 & 0.5086 \\
 & \encOrdinal           & 0.6126 & 0.4841 & 0.5964 & 0.4721 & 0.6131 & 0.4845 \\
 & \encOnehot            & 0.5918 & 0.4689 & 0.6689 & 0.5307 & 0.5891 & 0.4668 \\
 & \encCyclic            & 0.5891 & 0.4683 & 0.5394 & 0.4478 & 0.5908 & 0.4690 \\
 & \encTcombo            & \underline{0.5014} & \underline{0.3971} & \underline{0.5248} & \underline{0.4175} & \underline{0.5006} & \underline{0.3964} \\
 & \encTholEight         & 0.5048 & 0.3987 & 0.5875 & 0.4435 & 0.5020 & 0.3972 \\
 & \encTholAll           & 0.5177 & 0.4099 & 0.9258 & 0.7331 & 0.5038 & 0.3989 \\
 & \encTholPartial       & 0.5154 & 0.4087 & 0.6741 & 0.5220 & 0.5100 & 0.4049 \\
\midrule
\multirow{8}{*}{TiRex-2}
 & \encNone              & 0.5693 & 0.4578 & 2.4610 & 2.1697 & 0.5050 & 0.3996 \\
 & \encOrdinal           & 0.5641 & 0.4511 & 2.4111 & 2.1004 & 0.5013 & 0.3950 \\
 & \encOnehot            & 0.5790 & 0.4652 & 2.4781 & 2.1797 & 0.5144 & 0.4070 \\
 & \encCyclic            & 0.5614 & 0.4496 & 2.4319 & 2.1301 & 0.4978 & 0.3925 \\
 & \encTcombo            & \underline{0.4676} & \underline{0.3726} & \textbf{\underline{0.4855}} & \textbf{\underline{0.3819}} & \underline{0.4670} & \underline{0.3722} \\
 & \encTholEight         & 0.4705 & 0.3752 & 0.4987 & 0.3866 & 0.4696 & 0.3749 \\
 & \encTholAll           & 0.4790 & 0.3816 & 0.7147 & 0.5359 & 0.4710 & 0.3764 \\
 & \encTholPartial       & 0.5379 & 0.4324 & 1.5530 & 1.3258 & 0.5034 & 0.4020 \\
\midrule
\multirow{8}{*}{TimesFM-3}
 & \encNone              & 0.4649 & 0.3733 & 2.5292 & 2.1733 & 0.3947 & 0.3121 \\
 & \encOrdinal           & 0.4042 & 0.3208 & 0.7050 & 0.5865 & 0.3940 & 0.3118 \\
 & \encOnehot            & 0.4147 & 0.3264 & 0.9716 & 0.7456 & 0.3957 & 0.3121 \\
 & \encCyclic            & \textbf{\underline{0.4019}} & \textbf{\underline{0.3186}} & 0.6795 & 0.5774 & \textbf{\underline{0.3924}} & \textbf{\underline{0.3098}} \\
 & \encTcombo            & 0.4059 & 0.3213 & \underline{0.5161} & \underline{0.4259} & 0.4021 & 0.3177 \\
 & \encTholEight         & 0.4064 & 0.3215 & 0.5316 & 0.4313 & 0.4022 & 0.3178 \\
 & \encTholAll           & 0.4093 & 0.3237 & 0.5763 & 0.4617 & 0.4036 & 0.3190 \\
 & \encTholPartial       & 0.4117 & 0.3290 & 0.5944 & 0.4684 & 0.4055 & 0.3243 \\
\bottomrule
\end{tabular}}
\end{table}
\clearpage
\begin{figure}[H]
    \centering
    \includegraphics[width=1.0\linewidth]{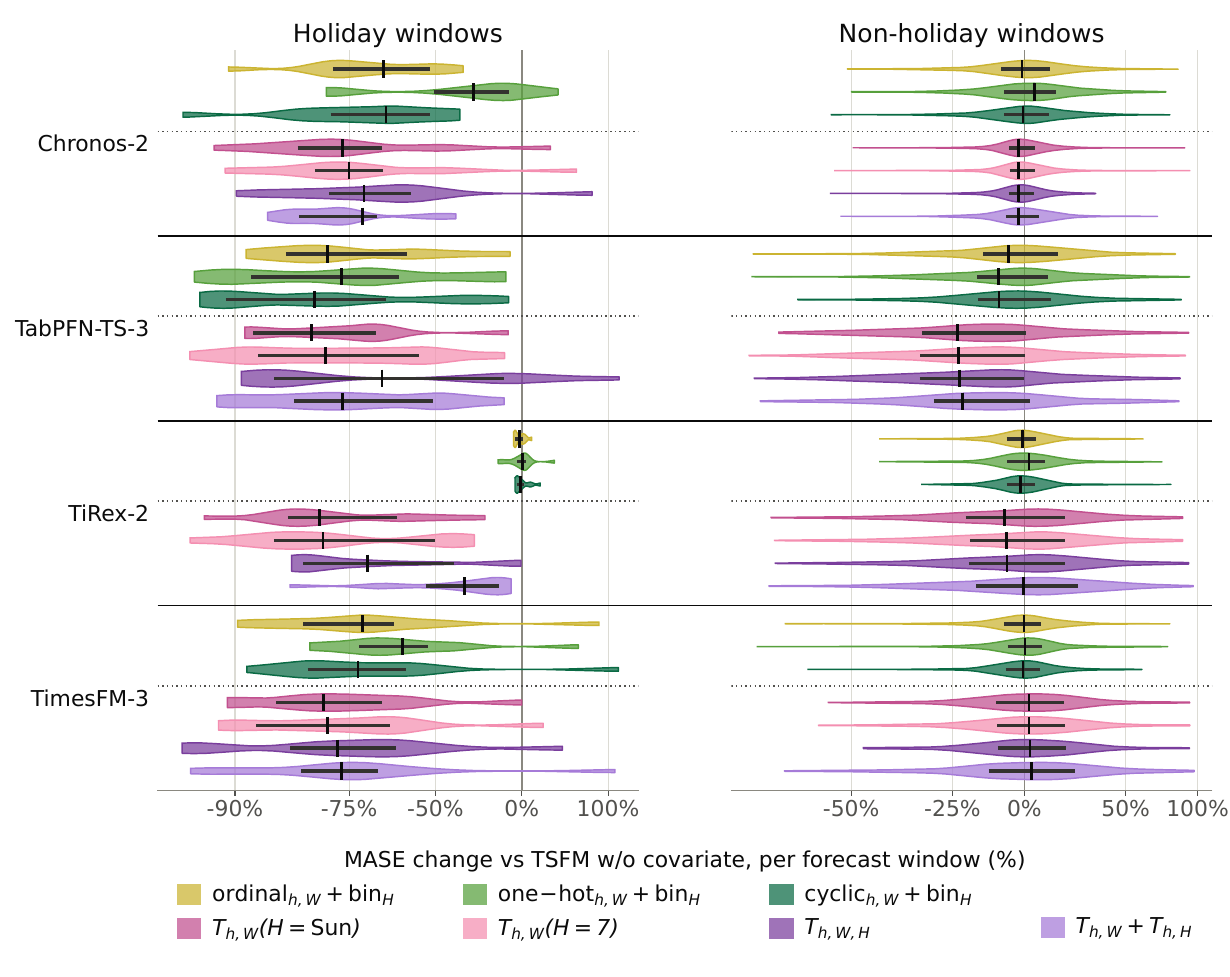}
    \caption{MASE change against the model w/o covariates on the \texttt{TransnetBW} dataset.}
    \label{fig:effect_entsoe_transnetbw_load}
\end{figure}

\begin{figure}[H]
    \centering
    \includegraphics[width=1.0\linewidth]{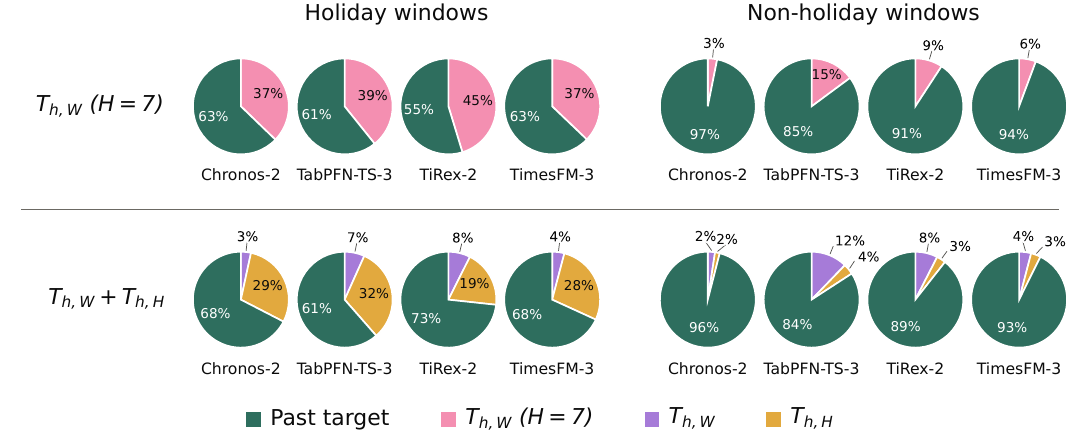}
    \caption{SHAP contributions for all models on the \texttt{TransnetBW} dataset, exemplarily depicted for the \encTholEight and \encTholPartial encodings.}
    \label{fig:placeholder}
\end{figure}

\clearpage
\subsubsection{FeederBW\_33}
\begin{table}[H]
\centering
\caption{Results for the \texttt{FeederBW\_033} dataset, split by date-type.}
\label{tab:eval5-feeder-bw-033}
\resizebox{\textwidth}{!}{%
\begin{tabular}{ll rr rr rr}
\toprule
& & \multicolumn{2}{c}{\textbf{All days}} & \multicolumn{2}{c}{\textbf{Holidays only}} & \multicolumn{2}{c}{\textbf{Non-holidays only}} \\
\cmidrule(lr){3-4} \cmidrule(lr){5-6} \cmidrule(lr){7-8}
\textbf{Model} & \textbf{Encoding} & \textbf{MASE} & \textbf{SQL} & \textbf{MASE} & \textbf{SQL} & \textbf{MASE} & \textbf{SQL}
 \\
\midrule
\multirow{2}{*}{Baselines}
 & seasonal-naive (168h) & \underline{1.0841} & \underline{1.0291} & 2.2640 & 2.0940 & \underline{1.0474} & \underline{0.9960} \\
 & $T_{h,W}(H=Sun)_{naive}$    & 1.1726 & 1.0772 & \underline{1.4250} & \underline{1.2242} & 1.1648 & 1.0727 \\
\midrule
\multirow{8}{*}{Chronos-2}
 & \encNone              & 0.8007 & 0.6230 & 2.0622 & 1.7281 & 0.7615 & 0.5886 \\
 & \encOrdinal           & 0.7736 & 0.6029 & 1.3490 & 1.0991 & 0.7557 & 0.5875 \\
 & \encOnehot            & 0.8060 & 0.6299 & 1.7825 & 1.4414 & 0.7756 & 0.6047 \\
 & \encCyclic            & 0.7800 & 0.6060 & 1.4794 & 1.2052 & 0.7583 & 0.5874 \\
 & \encTcombo            & 0.7519 & 0.5827 & 0.8203 & 0.6677 & \underline{0.7498} & \underline{0.5801} \\
 & \encTholEight         & \underline{0.7514} & \underline{0.5822} & \underline{0.7832} & \underline{0.6388} & 0.7504 & 0.5805 \\
 & \encTholAll           & 0.7532 & 0.5833 & 0.8484 & 0.6679 & 0.7502 & 0.5806 \\
 & \encTholPartial       & 0.7536 & 0.5852 & 0.8603 & 0.7031 & 0.7503 & 0.5815 \\
\midrule
\multirow{8}{*}{TabPFN-TS-3}
 & \encNone              & 0.8464 & 0.6569 & 2.2249 & 1.8579 & 0.8036 & 0.6196 \\
 & \encOrdinal           & \textbf{\underline{0.7316}} & \textbf{\underline{0.5621}} & 0.5745 & 0.4371 & 0.7365 & 0.5660 \\
 & \encOnehot            & 0.7322 & 0.5651 & 0.6186 & 0.4809 & \textbf{\underline{0.7357}} & 0.5677 \\
 & \encCyclic            & 0.7384 & 0.5685 & \textbf{\underline{0.5686}} & \textbf{\underline{0.4297}} & 0.7437 & 0.5728 \\
 & \encTcombo            & 0.7349 & 0.5635 & 0.6398 & 0.4822 & 0.7378 & 0.5661 \\
 & \encTholEight         & 0.7349 & 0.5637 & 0.6502 & 0.4982 & 0.7375 & \textbf{\underline{0.5657}} \\
 & \encTholAll           & 0.7433 & 0.5711 & 0.8454 & 0.6746 & 0.7402 & 0.5679 \\
 & \encTholPartial       & 0.7343 & 0.5649 & 0.6235 & 0.4956 & 0.7377 & 0.5671 \\
\midrule
\multirow{8}{*}{TiRex-2}
 & \encNone              & 0.8333 & 0.6527 & 2.1872 & 1.8520 & 0.7913 & 0.6154 \\
 & \encOrdinal           & 0.8319 & 0.6526 & 2.1681 & 1.8299 & 0.7904 & 0.6160 \\
 & \encOnehot            & 0.8370 & 0.6545 & 2.3092 & 1.9461 & 0.7912 & 0.6143 \\
 & \encCyclic            & 0.8327 & 0.6520 & 2.2293 & 1.8775 & 0.7893 & 0.6139 \\
 & \encTcombo            & \underline{0.7728} & \underline{0.6078} & \underline{0.6535} & \underline{0.5087} & 0.7765 & \underline{0.6109} \\
 & \encTholEight         & 0.7755 & 0.6108 & 0.7539 & 0.5941 & \underline{0.7762} & 0.6113 \\
 & \encTholAll           & 0.7852 & 0.6186 & 1.0662 & 0.8531 & 0.7765 & 0.6113 \\
 & \encTholPartial       & 0.8255 & 0.6517 & 1.9920 & 1.6878 & 0.7892 & 0.6195 \\
\midrule
\multirow{8}{*}{TimesFM-3}
 & \encNone              & 0.7882 & 0.6132 & 2.0202 & 1.6470 & 0.7499 & 0.5811 \\
 & \encOrdinal           & 0.7582 & 0.5876 & 1.1298 & 0.9001 & 0.7466 & 0.5779 \\
 & \encOnehot            & 0.7577 & 0.5876 & 1.1647 & 0.9413 & 0.7451 & 0.5766 \\
 & \encCyclic            & 0.7558 & 0.5889 & 0.9021 & 0.7292 & 0.7512 & 0.5846 \\
 & \encTcombo            & \underline{0.7393} & \underline{0.5723} & 0.7375 & \underline{0.5602} & \underline{0.7393} & \underline{0.5727} \\
 & \encTholEight         & 0.7395 & 0.5726 & \underline{0.7329} & 0.5626 & 0.7397 & 0.5729 \\
 & \encTholAll           & 0.7439 & 0.5773 & 0.8773 & 0.7142 & 0.7398 & 0.5730 \\
 & \encTholPartial       & 0.7448 & 0.5761 & 0.7485 & 0.5670 & 0.7447 & 0.5764 \\
\bottomrule
\end{tabular}}
\end{table}
\clearpage
\begin{figure}[H]
    \centering
    \includegraphics[width=1.0\linewidth]{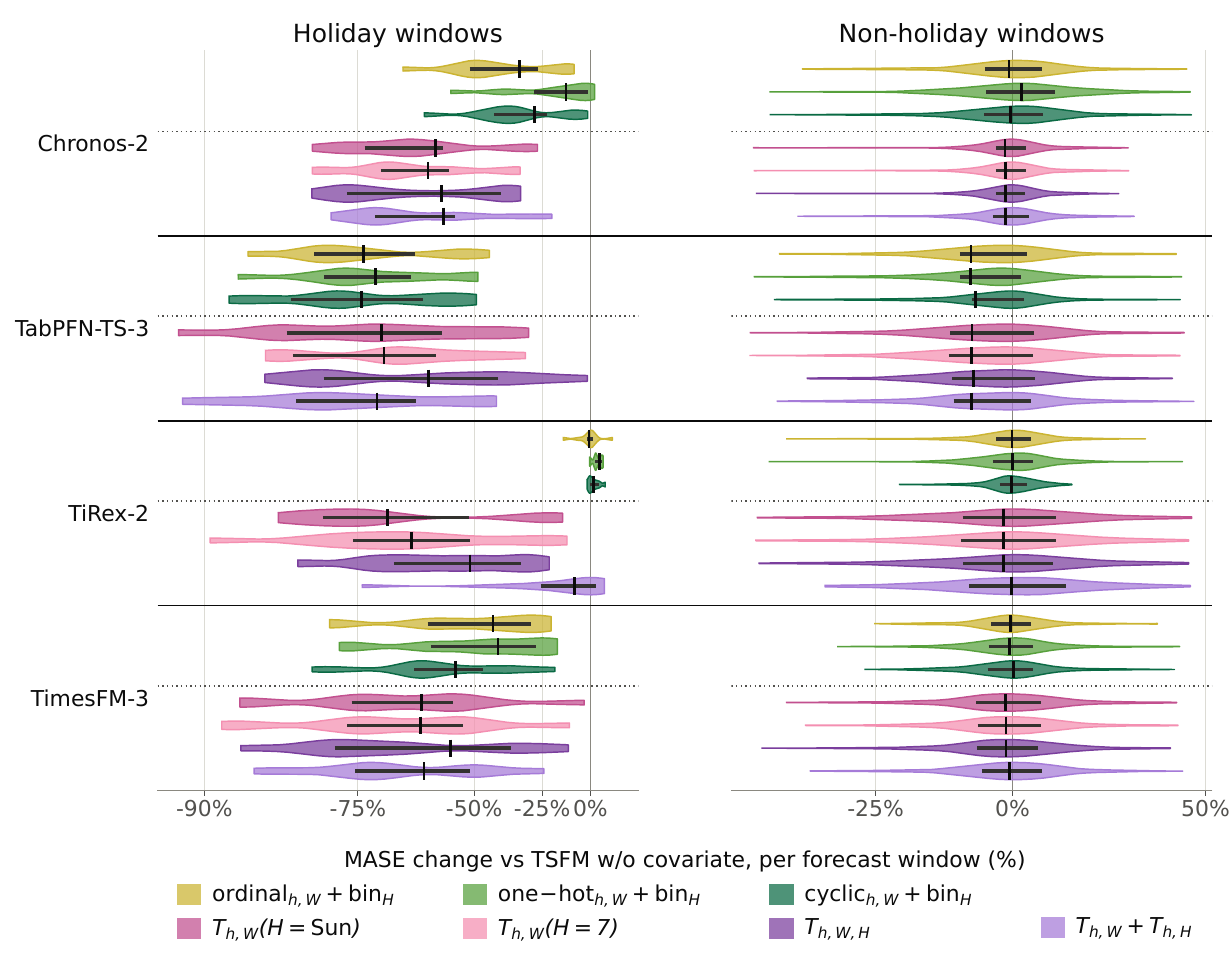}
    \caption{MASE change against the model w/o covariates on the \texttt{FeederBW\_33} dataset.}
    \label{fig:effect_feeder_bw_033}
\end{figure}

\begin{figure}[H]
    \centering
    \includegraphics[width=1.0\linewidth]{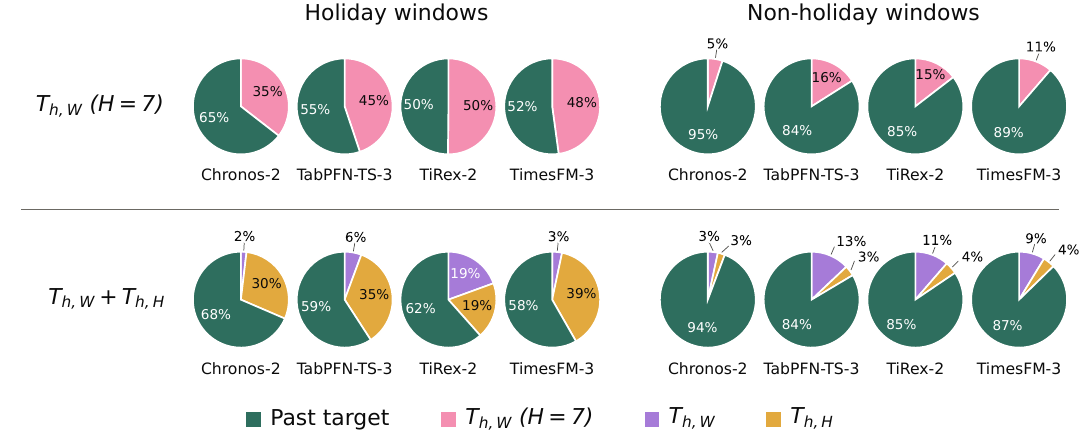}
    \caption{SHAP contributions for all models on the \texttt{FeederBW\_33} dataset, exemplarily depicted for the \encTholEight and \encTholPartial encodings.}
    \label{fig:placeholder}
\end{figure}
\clearpage

\subsubsection{FeederBW\_35}
\begin{table}[H]
\centering
\caption{Results for the \texttt{FeederBW\_035} dataset, split by date-type.}
\label{tab:eval5-feeder-bw-035}
\resizebox{\textwidth}{!}{%
\begin{tabular}{ll rr rr rr}
\toprule
& & \multicolumn{2}{c}{\textbf{All days}} & \multicolumn{2}{c}{\textbf{Holidays only}} & \multicolumn{2}{c}{\textbf{Non-holidays only}} \\
\cmidrule(lr){3-4} \cmidrule(lr){5-6} \cmidrule(lr){7-8}
\textbf{Model} & \textbf{Encoding} & \textbf{MASE} & \textbf{SQL} & \textbf{MASE} & \textbf{SQL} & \textbf{MASE} & \textbf{SQL}
 \\
\midrule
\multirow{2}{*}{Baselines}
 & seasonal-naive (168h) & 1.0562 & 0.9302 & 2.1083 & 1.8575 & 1.0235 & 0.9014 \\
 & $T_{h,W}(H=Sun)_{naive}$    & \underline{0.9036} & \underline{0.7487} & \underline{0.9751} & \underline{0.7881} & \underline{0.9013} & \underline{0.7475} \\
\midrule
\multirow{8}{*}{Chronos-2}
 & \encNone              & 0.7308 & 0.5812 & 1.5818 & 1.3653 & 0.7043 & 0.5569 \\
 & \encOrdinal           & 0.7293 & 0.5775 & 1.4074 & 1.1544 & 0.7082 & 0.5596 \\
 & \encOnehot            & 0.7479 & 0.5948 & 1.5148 & 1.2989 & 0.7241 & 0.5729 \\
 & \encCyclic            & 0.7314 & 0.5800 & 1.4410 & 1.2025 & 0.7094 & 0.5606 \\
 & \encTcombo            & 0.7047 & 0.5567 & 1.0764 & 0.8305 & \underline{0.6932} & \underline{0.5482} \\
 & \encTholEight         & 0.7053 & 0.5572 & 1.0869 & 0.8378 & 0.6934 & 0.5485 \\
 & \encTholAll           & \underline{0.7040} & \underline{0.5564} & \underline{1.0245} & \underline{0.7906} & 0.6940 & 0.5492 \\
 & \encTholPartial       & 0.7070 & 0.5591 & 1.0674 & 0.8446 & 0.6958 & 0.5502 \\
\midrule
\multirow{8}{*}{TabPFN-TS-3}
 & \encNone              & 0.8190 & 0.6549 & 1.8137 & 1.5084 & 0.7881 & 0.6284 \\
 & \encOrdinal           & 0.7487 & 0.5973 & 0.8333 & 0.6456 & 0.7461 & 0.5958 \\
 & \encOnehot            & 0.7362 & 0.5894 & 0.8550 & 0.6661 & 0.7325 & 0.5870 \\
 & \encCyclic            & 0.7499 & 0.5976 & 0.8990 & 0.6856 & 0.7453 & 0.5949 \\
 & \encTcombo            & \underline{0.7178} & \underline{0.5716} & \textbf{\underline{0.7465}} & \textbf{\underline{0.5915}} & 0.7169 & 0.5710 \\
 & \encTholEight         & 0.7212 & 0.5740 & 0.7900 & 0.6315 & 0.7190 & 0.5722 \\
 & \encTholAll           & 0.7220 & 0.5745 & 0.8965 & 0.7240 & 0.7166 & \underline{0.5699} \\
 & \encTholPartial       & 0.7211 & 0.5744 & 0.9167 & 0.7216 & \underline{0.7150} & 0.5699 \\
\midrule
\multirow{8}{*}{TiRex-2}
 & \encNone              & 0.7901 & 0.6290 & 1.5769 & 1.3042 & 0.7656 & 0.6080 \\
 & \encOrdinal           & 0.7948 & 0.6318 & 1.5029 & 1.2081 & 0.7728 & 0.6139 \\
 & \encOnehot            & 0.8038 & 0.6377 & 1.6312 & 1.3389 & 0.7781 & 0.6159 \\
 & \encCyclic            & 0.7939 & 0.6313 & 1.5948 & 1.3083 & 0.7690 & 0.6102 \\
 & \encTcombo            & 0.7269 & \underline{0.5741} & 0.8155 & 0.6436 & 0.7242 & \underline{0.5719} \\
 & \encTholEight         & \underline{0.7268} & 0.5742 & \underline{0.8119} & 0.6462 & \underline{0.7241} & 0.5719 \\
 & \encTholAll           & 0.7280 & 0.5750 & 0.8155 & \underline{0.6414} & 0.7253 & 0.5729 \\
 & \encTholPartial       & 0.7612 & 0.6031 & 1.6265 & 1.3650 & 0.7343 & 0.5794 \\
\midrule
\multirow{8}{*}{TimesFM-3}
 & \encNone              & 0.7254 & 0.5784 & 1.5354 & 1.2934 & 0.7003 & 0.5562 \\
 & \encOrdinal           & 0.7123 & 0.5641 & 1.2526 & 0.9731 & 0.6955 & 0.5514 \\
 & \encOnehot            & 0.7153 & 0.5668 & 1.2714 & 0.9961 & 0.6980 & 0.5534 \\
 & \encCyclic            & 0.7130 & 0.5650 & 1.1922 & 0.9160 & 0.6981 & 0.5541 \\
 & \encTcombo            & 0.6912 & \textbf{\underline{0.5472}} & \underline{0.7847} & \underline{0.6134} & 0.6883 & 0.5452 \\
 & \encTholEight         & \textbf{\underline{0.6912}} & 0.5475 & 0.7976 & 0.6347 & \textbf{\underline{0.6879}} & \textbf{\underline{0.5448}} \\
 & \encTholAll           & 0.6926 & 0.5476 & 0.8190 & 0.6225 & 0.6887 & 0.5453 \\
 & \encTholPartial       & 0.7021 & 0.5568 & 0.8577 & 0.6797 & 0.6973 & 0.5530 \\
\bottomrule
\end{tabular}}
\end{table}

\clearpage

\begin{figure}[H]
    \centering
    \includegraphics[width=1.0\linewidth]{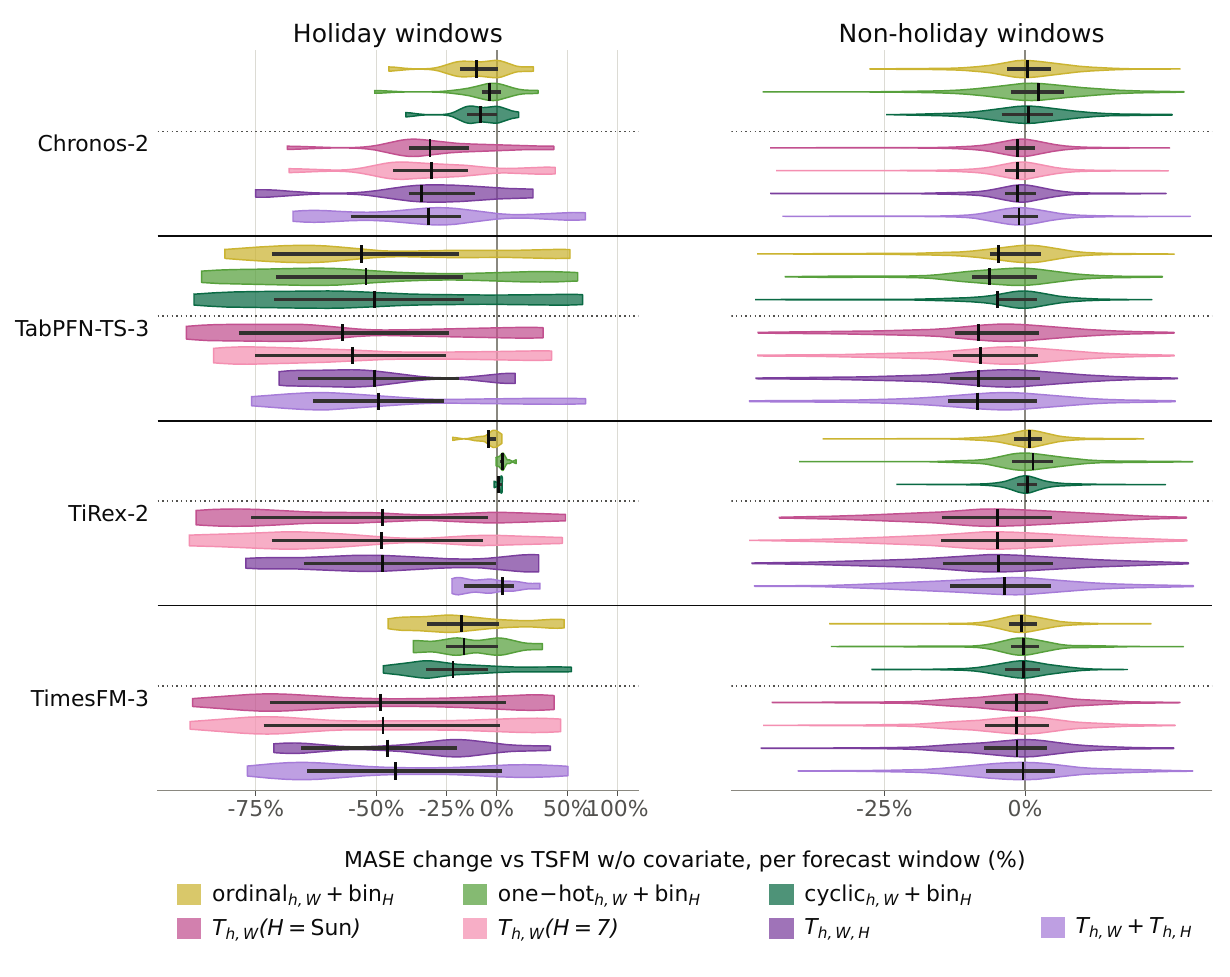}
    \caption{MASE change against the model w/o covariates on the \texttt{FeederBW\_35} dataset.}
    \label{fig:effect_feeder_bw_035}
\end{figure}

\begin{figure}[H]
    \centering
    \includegraphics[width=1.0\linewidth]{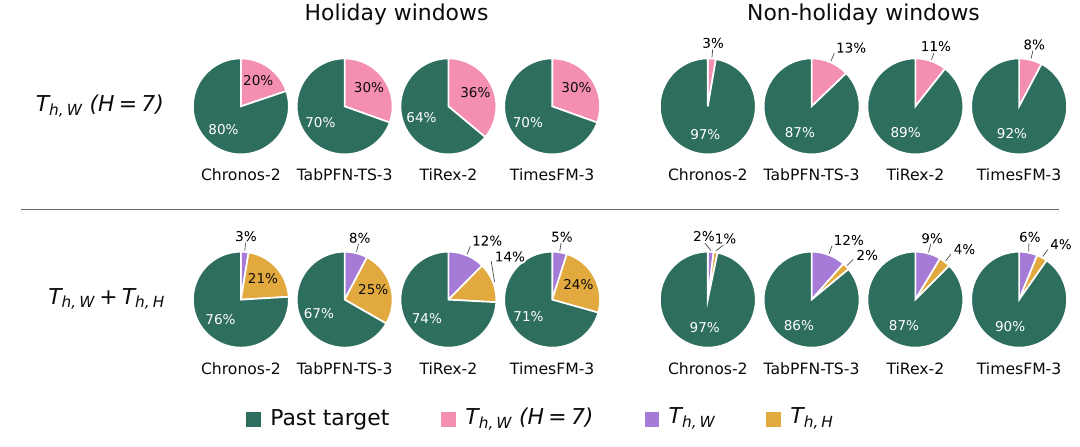}
    \caption{SHAP contributions for all models on the \texttt{FeederBW\_35} dataset, exemplarily depicted for the \encTholEight and \encTholPartial encodings.}
    \label{fig:placeholder}
\end{figure}
\clearpage

\subsubsection{FeederBW\_91}
\begin{table}[H]
\centering
\caption{Results for the \texttt{FeederBW\_091} dataset, split by date-type.}
\label{tab:eval5-feeder-bw-091}
\resizebox{\textwidth}{!}{%
\begin{tabular}{ll rr rr rr}
\toprule
& & \multicolumn{2}{c}{\textbf{All days}} & \multicolumn{2}{c}{\textbf{Holidays only}} & \multicolumn{2}{c}{\textbf{Non-holidays only}} \\
\cmidrule(lr){3-4} \cmidrule(lr){5-6} \cmidrule(lr){7-8}
\textbf{Model} & \textbf{Encoding} & \textbf{MASE} & \textbf{SQL} & \textbf{MASE} & \textbf{SQL} & \textbf{MASE} & \textbf{SQL}
 \\
\midrule
\multirow{2}{*}{Baselines}
 & seasonal-naive (168h) & \underline{0.9918} & \underline{0.8589} & 2.4438 & 2.1058 & \underline{0.9467} & \underline{0.8202} \\
 & $T_{h,W}(H=Sun)_{naive}$    & 1.1284 & 0.8961 & \underline{1.0713} & \underline{0.8874} & 1.1302 & 0.8964 \\
\midrule
\multirow{8}{*}{Chronos-2}
 & \encNone              & 0.6666 & 0.5269 & 1.5522 & 1.3245 & 0.6391 & 0.5021 \\
 & \encOrdinal           & 0.6586 & 0.5193 & 1.1473 & 0.8890 & 0.6434 & 0.5079 \\
 & \encOnehot            & 0.6811 & 0.5404 & 1.4729 & 1.2183 & 0.6565 & 0.5193 \\
 & \encCyclic            & 0.6605 & 0.5220 & 1.2000 & 0.9424 & 0.6437 & 0.5089 \\
 & \encTcombo            & 0.6365 & 0.5016 & 0.8360 & 0.6443 & 0.6303 & 0.4972 \\
 & \encTholEight         & 0.6358 & 0.5013 & 0.8126 & 0.6336 & 0.6303 & 0.4971 \\
 & \encTholAll           & 0.6366 & 0.5017 & 0.8393 & 0.6591 & \textbf{\underline{0.6303}} & \textbf{\underline{0.4968}} \\
 & \encTholPartial       & \textbf{\underline{0.6358}} & \textbf{\underline{0.5007}} & \underline{0.7585} & \underline{0.5982} & 0.6319 & 0.4977 \\
\midrule
\multirow{8}{*}{TabPFN-TS-3}
 & \encNone              & 0.7740 & 0.6144 & 2.0669 & 1.7303 & 0.7339 & 0.5798 \\
 & \encOrdinal           & 0.6876 & 0.5437 & 0.8053 & 0.6304 & 0.6840 & 0.5411 \\
 & \encOnehot            & 0.6796 & 0.5356 & 0.8271 & 0.6451 & 0.6750 & 0.5322 \\
 & \encCyclic            & 0.6821 & 0.5385 & 0.8091 & 0.6320 & 0.6781 & 0.5356 \\
 & \encTcombo            & 0.6569 & 0.5187 & \underline{0.7583} & \textbf{\underline{0.5854}} & 0.6538 & 0.5166 \\
 & \encTholEight         & 0.6589 & 0.5200 & 0.8121 & 0.6160 & 0.6542 & 0.5170 \\
 & \encTholAll           & 0.6628 & 0.5246 & 0.9788 & 0.7778 & 0.6530 & 0.5167 \\
 & \encTholPartial       & \underline{0.6553} & \underline{0.5175} & 0.8017 & 0.6175 & \underline{0.6507} & \underline{0.5144} \\
\midrule
\multirow{8}{*}{TiRex-2}
 & \encNone              & 0.7290 & 0.5788 & 1.8134 & 1.5016 & 0.6953 & 0.5501 \\
 & \encOrdinal           & 0.7343 & 0.5843 & 1.7524 & 1.4342 & 0.7027 & 0.5579 \\
 & \encOnehot            & 0.7343 & 0.5845 & 1.9132 & 1.5778 & 0.6976 & 0.5536 \\
 & \encCyclic            & 0.7307 & 0.5812 & 1.9005 & 1.5618 & 0.6944 & 0.5507 \\
 & \encTcombo            & 0.6959 & 0.5516 & 0.9638 & 0.7519 & 0.6876 & 0.5453 \\
 & \encTholEight         & 0.6957 & 0.5514 & 0.9637 & 0.7566 & 0.6874 & 0.5450 \\
 & \encTholAll           & \underline{0.6930} & \underline{0.5497} & \underline{0.9061} & \underline{0.7187} & \underline{0.6864} & \underline{0.5445} \\
 & \encTholPartial       & 0.7307 & 0.5797 & 1.6295 & 1.3411 & 0.7028 & 0.5560 \\
\midrule
\multirow{8}{*}{TimesFM-3}
 & \encNone              & 0.6652 & 0.5277 & 1.5620 & 1.2881 & 0.6373 & 0.5041 \\
 & \encOrdinal           & 0.6454 & 0.5098 & 0.8840 & 0.7056 & 0.6380 & 0.5037 \\
 & \encOnehot            & 0.6452 & 0.5100 & 0.9129 & 0.7221 & 0.6369 & 0.5034 \\
 & \encCyclic            & \underline{0.6415} & 0.5076 & 0.8899 & 0.7054 & \underline{0.6338} & \underline{0.5015} \\
 & \encTcombo            & 0.6422 & 0.5065 & \textbf{\underline{0.7560}} & 0.5861 & 0.6387 & 0.5040 \\
 & \encTholEight         & 0.6420 & \underline{0.5065} & 0.7588 & \underline{0.5859} & 0.6384 & 0.5040 \\
 & \encTholAll           & 0.6434 & 0.5076 & 0.8080 & 0.6237 & 0.6383 & 0.5040 \\
 & \encTholPartial       & 0.6430 & 0.5066 & 0.7847 & 0.6107 & 0.6386 & 0.5033 \\
\bottomrule
\end{tabular}}
\end{table}
\clearpage

\begin{figure}[H]
    \centering    \includegraphics[width=1.0\linewidth]{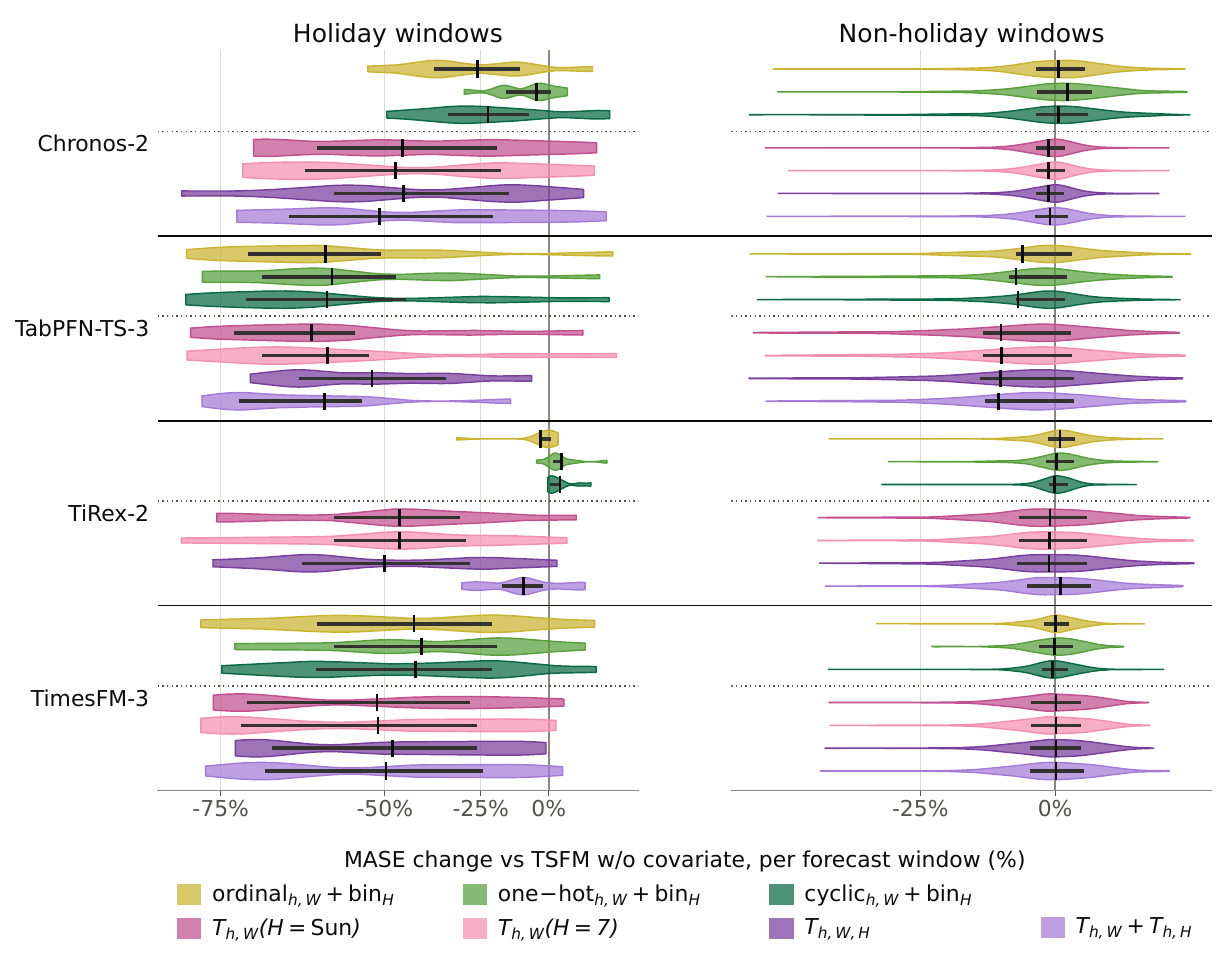}
    \caption{MASE change against the model w/o covariates on the \texttt{FeederBW\_91} dataset.}
    \label{fig:effect_feeder_bw_091}
\end{figure}

\begin{figure}[H]
    \centering
    \includegraphics[width=1.0\linewidth]{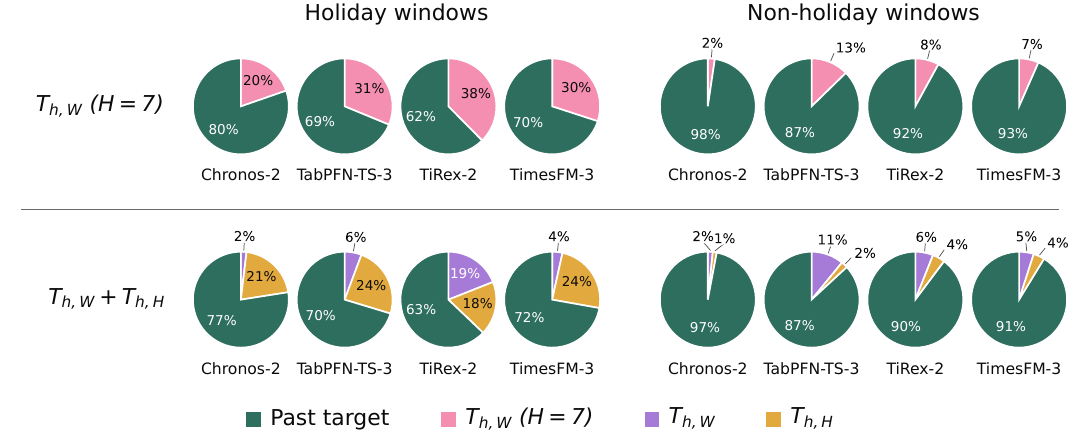}
    \caption{SHAP contributions for all models on the \texttt{FeederBW\_91} dataset, exemplarily depicted for the \encTholEight and \encTholPartial encodings.}
    \label{fig:placeholder}
\end{figure}
\clearpage
\subsubsection{Alicante}
\begin{table}[H]
\centering
\caption{Results for the \texttt{Alicante} dataset, split by date-type.}
\label{tab:eval5-alicante-water}
\resizebox{\textwidth}{!}{%
\begin{tabular}{ll rr rr rr}
\toprule
& & \multicolumn{2}{c}{\textbf{All days}} & \multicolumn{2}{c}{\textbf{Holidays only}} & \multicolumn{2}{c}{\textbf{Non-holidays only}} \\
\cmidrule(lr){3-4} \cmidrule(lr){5-6} \cmidrule(lr){7-8}
\textbf{Model} & \textbf{Encoding} & \textbf{MASE} & \textbf{SQL} & \textbf{MASE} & \textbf{SQL} & \textbf{MASE} & \textbf{SQL}
 \\
\midrule
\multirow{2}{*}{Baselines}
 & seasonal-naive (168h) & 0.9210 & 0.9390 & 1.4840 & 1.4474 & 0.9019 & 0.9217 \\
 & $T_{h,W}(H=Sun)_{naive}$    & \underline{0.7462} & \underline{0.6987} & \underline{0.6702} & \underline{0.6958} & \underline{0.7488} & \underline{0.6988} \\
\midrule
\multirow{8}{*}{Chronos-2}
 & \encNone              & \underline{0.6474} & 0.5266 & 0.9571 & 0.7988 & \underline{0.6369} & \underline{0.5174} \\
 & \encOrdinal           & 0.6624 & 0.5368 & 0.9088 & 0.7497 & 0.6540 & 0.5296 \\
 & \encOnehot            & 0.6710 & 0.5450 & 0.9388 & 0.7900 & 0.6619 & 0.5367 \\
 & \encCyclic            & 0.6625 & 0.5374 & 0.9108 & 0.7532 & 0.6540 & 0.5301 \\
 & \encTcombo            & 0.6504 & 0.5271 & 0.8741 & 0.7016 & 0.6428 & 0.5212 \\
 & \encTholEight         & 0.6481 & \underline{0.5256} & 0.8114 & 0.6514 & 0.6426 & 0.5213 \\
 & \encTholAll           & 0.6521 & 0.5284 & 0.9312 & 0.7360 & 0.6426 & 0.5213 \\
 & \encTholPartial       & 0.6492 & 0.5263 & \underline{0.7609} & \underline{0.6113} & 0.6454 & 0.5234 \\
\midrule
\multirow{8}{*}{TabPFN-TS-3}
 & \encNone              & 0.7050 & 0.5678 & 0.9734 & 0.7468 & 0.6959 & 0.5618 \\
 & \encOrdinal           & 0.6946 & 0.5588 & 0.6517 & 0.5190 & 0.6961 & 0.5602 \\
 & \encOnehot            & 0.6640 & 0.5380 & 0.6619 & 0.5261 & 0.6640 & 0.5384 \\
 & \encCyclic            & 0.6978 & 0.5619 & 0.6498 & 0.5189 & 0.6994 & 0.5633 \\
 & \encTcombo            & 0.6514 & \underline{0.5283} & \textbf{\underline{0.6114}} & \textbf{\underline{0.4854}} & 0.6528 & 0.5297 \\
 & \encTholEight         & 0.6517 & 0.5287 & 0.6381 & 0.5102 & 0.6522 & 0.5293 \\
 & \encTholAll           & 0.6626 & 0.5388 & 0.9830 & 0.8253 & 0.6517 & 0.5291 \\
 & \encTholPartial       & \underline{0.6507} & 0.5284 & 0.6723 & 0.5366 & \underline{0.6499} & \underline{0.5282} \\
\midrule
\multirow{8}{*}{TiRex-2}
 & \encNone              & \underline{0.6694} & \underline{0.5422} & 0.9624 & 0.7909 & \underline{0.6595} & \underline{0.5338} \\
 & \encOrdinal           & 0.6791 & 0.5493 & 0.9685 & 0.7944 & 0.6693 & 0.5410 \\
 & \encOnehot            & 0.6892 & 0.5560 & 0.9844 & 0.8078 & 0.6792 & 0.5475 \\
 & \encCyclic            & 0.6729 & 0.5453 & 0.9716 & 0.7970 & 0.6628 & 0.5367 \\
 & \encTcombo            & 0.6730 & 0.5429 & 0.6890 & 0.5537 & 0.6724 & 0.5425 \\
 & \encTholEight         & 0.6728 & 0.5430 & \underline{0.6850} & \underline{0.5509} & 0.6724 & 0.5427 \\
 & \encTholAll           & 0.6813 & 0.5496 & 0.9290 & 0.7481 & 0.6729 & 0.5429 \\
 & \encTholPartial       & 0.6846 & 0.5529 & 0.9052 & 0.7403 & 0.6771 & 0.5466 \\
\midrule
\multirow{8}{*}{TimesFM-3}
 & \encNone              & 0.6399 & 0.5202 & 0.9515 & 0.7826 & \textbf{\underline{0.6293}} & \textbf{\underline{0.5113}} \\
 & \encOrdinal           & 0.6409 & 0.5210 & 0.9415 & 0.7725 & 0.6307 & 0.5125 \\
 & \encOnehot            & 0.6429 & 0.5224 & 0.9478 & 0.7855 & 0.6326 & 0.5135 \\
 & \encCyclic            & 0.6414 & 0.5210 & 0.9372 & 0.7666 & 0.6314 & 0.5127 \\
 & \encTcombo            & \textbf{\underline{0.6375}} & 0.5186 & \underline{0.6474} & \underline{0.5163} & 0.6371 & 0.5187 \\
 & \encTholEight         & 0.6375 & \textbf{\underline{0.5185}} & 0.6594 & 0.5262 & 0.6367 & 0.5182 \\
 & \encTholAll           & 0.6439 & 0.5234 & 0.8250 & 0.6572 & 0.6377 & 0.5189 \\
 & \encTholPartial       & 0.6470 & 0.5277 & 0.7353 & 0.5853 & 0.6440 & 0.5257 \\
\bottomrule
\end{tabular}}
\end{table}

\clearpage

\begin{figure}[H]
    \centering
    \includegraphics[width=1.0\linewidth]{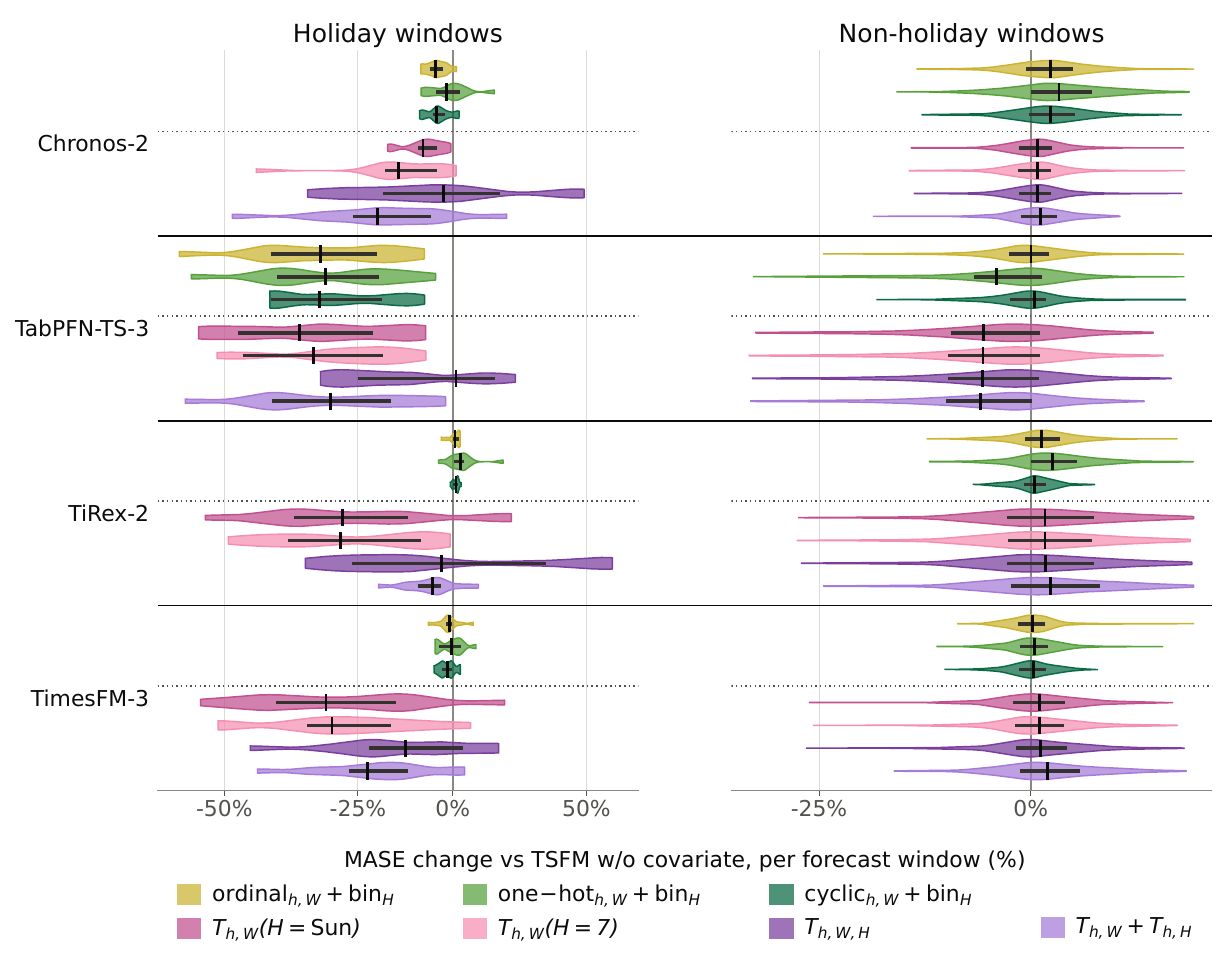}
    \caption{MASE change against the model w/o covariates on the \texttt{Alicante} dataset.}
    \label{fig:effect_alicante_water}
\end{figure}

\begin{figure}[H]
    \centering
    \includegraphics[width=1.0\linewidth]{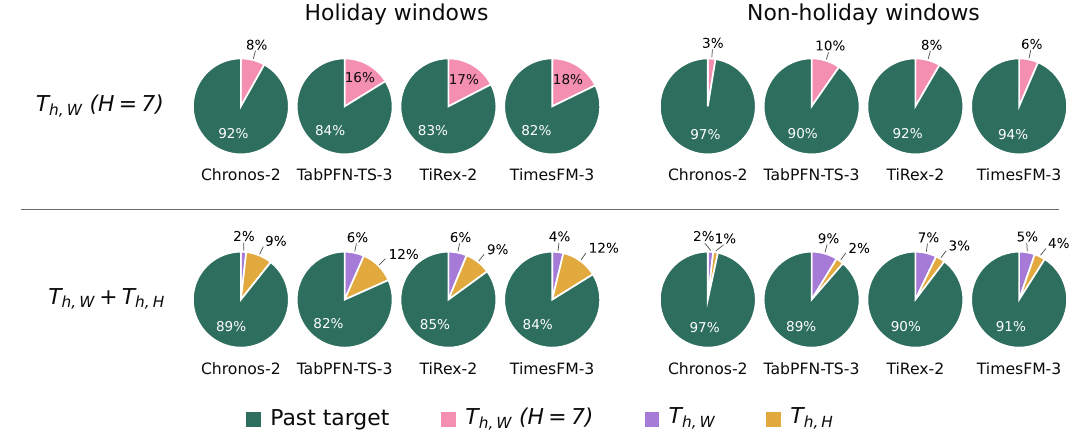}
    \caption{SHAP contributions for all models on the \texttt{Alicante} dataset, exemplarily depicted for the \encTholEight and \encTholPartial encodings.}
    \label{fig:placeholder}
\end{figure}
\clearpage

\subsubsection{uci\_bike}
\label{sec:App_results_uci_bike}
\begin{table}[H]
\centering
\caption{Results for the \texttt{uci\_bike} dataset, split by date-type.}
\label{tab:eval5-bike-sharing-dc}
\resizebox{\textwidth}{!}{%
\begin{tabular}{ll rr rr rr}
\toprule
& & \multicolumn{2}{c}{\textbf{All days}} & \multicolumn{2}{c}{\textbf{Holidays only}} & \multicolumn{2}{c}{\textbf{Non-holidays only}} \\
\cmidrule(lr){3-4} \cmidrule(lr){5-6} \cmidrule(lr){7-8}
\textbf{Model} & \textbf{Encoding} & \textbf{MASE} & \textbf{SQL} & \textbf{MASE} & \textbf{SQL} & \textbf{MASE} & \textbf{SQL}
 \\
\midrule
\multirow{2}{*}{Baselines}
 & seasonal-naive (168h) & \underline{1.1499} & \underline{1.0432} & 1.9203 & 1.6437 & \underline{1.1215} & \underline{1.0211} \\
 & $T_{h,W}(H=Sun)_{naive}$    & 1.4535 & 1.2436 & \underline{1.5922} & \underline{1.3471} & 1.4484 & 1.2398 \\
\midrule
\multirow{8}{*}{Chronos-2}
 & \encNone              & 0.7803 & 0.6308 & 1.4783 & 1.2061 & 0.7546 & 0.6096 \\
 & \encOrdinal           & 0.7748 & 0.6241 & 1.4739 & 1.1732 & 0.7490 & 0.6039 \\
 & \encOnehot            & 0.7989 & 0.6413 & 1.4869 & 1.2040 & 0.7736 & 0.6206 \\
 & \encCyclic            & 0.7818 & 0.6301 & 1.4849 & 1.1910 & 0.7559 & 0.6094 \\
 & \encTcombo            & 0.7607 & 0.6147 & 1.3000 & 1.0404 & 0.7409 & 0.5990 \\
 & \encTholEight         & \underline{0.7576} & \underline{0.6116} & \underline{1.2767} & \underline{1.0008} & 0.7384 & 0.5973 \\
 & \encTholAll           & 0.7621 & 0.6163 & 1.3325 & 1.0896 & 0.7411 & 0.5989 \\
 & \encTholPartial       & 0.7590 & 0.6127 & 1.3353 & 1.0587 & \textbf{\underline{0.7378}} & \underline{0.5963} \\
\midrule
\multirow{8}{*}{TabPFN-TS-3}
 & \encNone              & 0.9259 & 0.7702 & 1.6849 & 1.2865 & 0.8979 & 0.7512 \\
 & \encOrdinal           & 0.9239 & 0.7513 & 1.4144 & 1.0940 & 0.9058 & 0.7386 \\
 & \encOnehot            & 0.8685 & 0.6996 & 1.4456 & 1.1139 & 0.8472 & 0.6843 \\
 & \encCyclic            & 0.9743 & 0.7981 & 1.3298 & 1.0522 & 0.9612 & 0.7887 \\
 & \encTcombo            & \underline{0.8192} & \underline{0.6624} & \underline{1.1761} & \underline{0.9849} & 0.8061 & 0.6506 \\
 & \encTholEight         & 0.8256 & 0.6686 & 1.3651 & 1.1565 & 0.8057 & 0.6507 \\
 & \encTholAll           & 0.8450 & 0.6855 & 1.8066 & 1.5370 & 0.8096 & 0.6541 \\
 & \encTholPartial       & 0.8221 & 0.6668 & 1.5101 & 1.2462 & \underline{0.7967} & \underline{0.6455} \\
\midrule
\multirow{8}{*}{TiRex-2}
 & \encNone              & 0.8644 & 0.6976 & 1.5801 & 1.3287 & 0.8380 & 0.6744 \\
 & \encOrdinal           & 0.8865 & 0.7114 & 1.5462 & 1.2830 & 0.8622 & 0.6903 \\
 & \encOnehot            & 0.8669 & 0.7014 & 1.6341 & 1.3721 & 0.8386 & 0.6767 \\
 & \encCyclic            & 0.8622 & 0.6951 & 1.5776 & 1.3270 & 0.8358 & 0.6718 \\
 & \encTcombo            & \underline{0.8373} & \underline{0.6749} & \underline{1.0495} & \underline{0.8223} & \underline{0.8295} & \underline{0.6694} \\
 & \encTholEight         & 0.8453 & 0.6819 & 1.2130 & 0.9839 & 0.8318 & 0.6708 \\
 & \encTholAll           & 0.8481 & 0.6853 & 1.3146 & 1.0976 & 0.8309 & 0.6701 \\
 & \encTholPartial       & 0.8729 & 0.7036 & 1.5773 & 1.3078 & 0.8470 & 0.6813 \\
\midrule
\multirow{8}{*}{TimesFM-3}
 & \encNone              & 0.7660 & 0.6193 & 1.2742 & 1.0338 & 0.7473 & 0.6040 \\
 & \encOrdinal           & 0.7706 & 0.6215 & 1.2339 & 1.0032 & 0.7535 & 0.6074 \\
 & \encOnehot            & 0.7700 & 0.6224 & 1.2945 & 1.0737 & 0.7507 & 0.6058 \\
 & \encCyclic            & 0.7715 & 0.6218 & 1.2208 & 0.9951 & 0.7550 & 0.6081 \\
 & \encTcombo            & \textbf{\underline{0.7460}} & \textbf{\underline{0.6027}} & \textbf{\underline{0.9445}} & \textbf{\underline{0.7724}} & 0.7387 & 0.5965 \\
 & \encTholEight         & 0.7488 & 0.6040 & 1.0296 & 0.8341 & 0.7384 & 0.5955 \\
 & \encTholAll           & 0.7479 & 0.6037 & 1.0146 & 0.8311 & \underline{0.7381} & \textbf{\underline{0.5953}} \\
 & \encTholPartial       & 0.7656 & 0.6170 & 1.0873 & 0.8750 & 0.7538 & 0.6075 \\
\bottomrule
\end{tabular}}
\end{table}
\clearpage

\begin{figure}[H]
    \centering
    \includegraphics[width=1.0\linewidth]{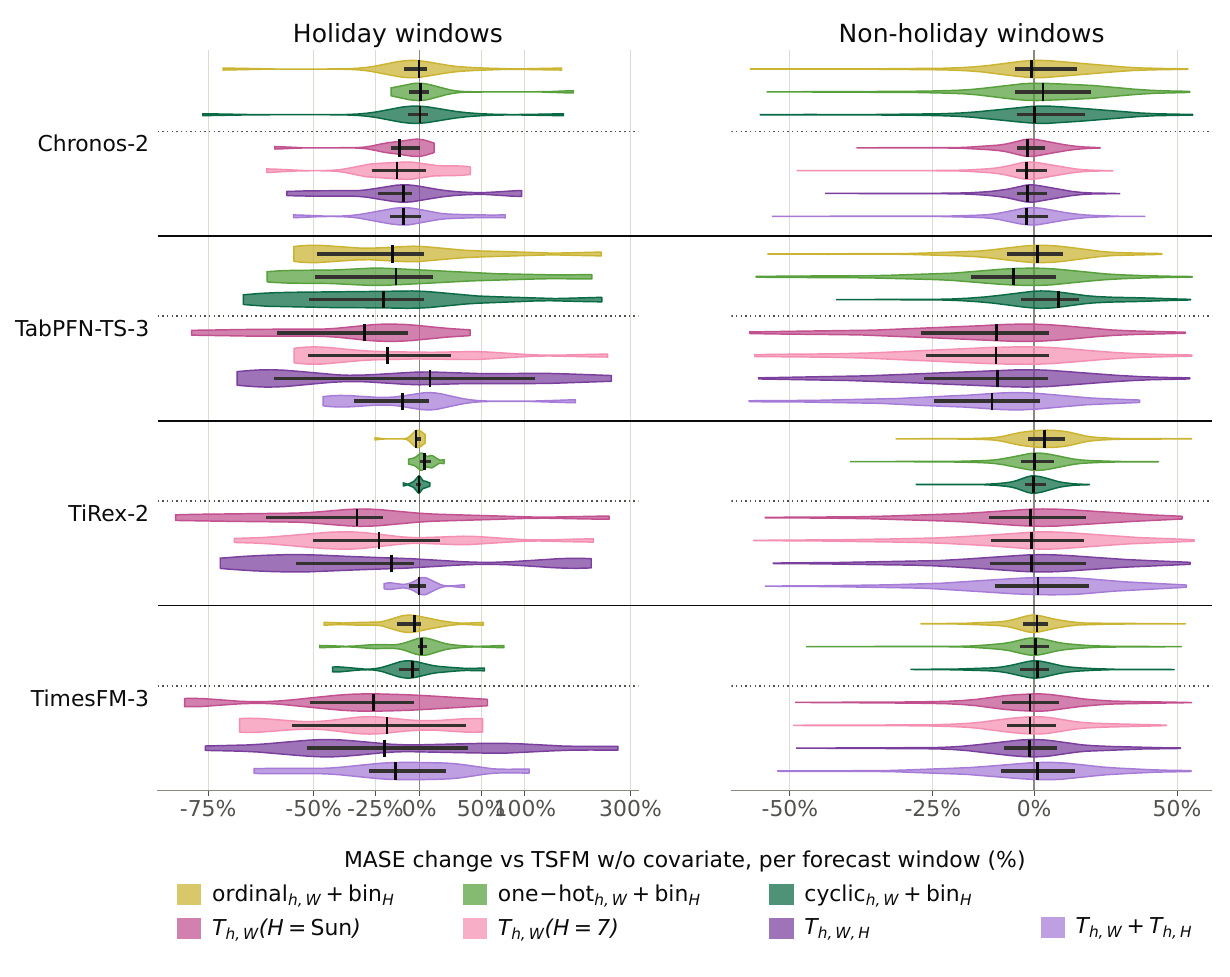}
    \caption{MASE change against the model w/o covariates on the \texttt{uci\_bike} dataset.}
    \label{fig:effect_bike_sharing_dc}
\end{figure}

\begin{figure}[H]
    \centering
    \includegraphics[width=1.0\linewidth]{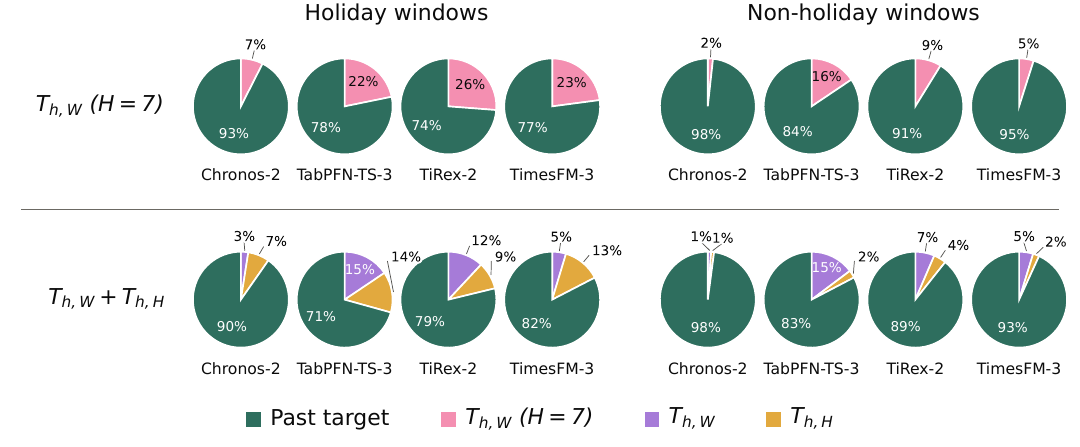}
    \caption{SHAP contributions for all models on the \texttt{uci\_bike} dataset, exemplarily depicted for the \encTholEight and \encTholPartial encodings.}
    \label{fig:placeholder}
\end{figure}
\clearpage

\subsubsection{Overall Results with Ground Truth as Covariate}
\label{sec:oracle_results}
\begin{table}[H]
\centering
\small
\caption{Oracle encoding (target fed in as a known covariate): overall results across all datasets, split by day type (mean over 9 datasets).}
\label{tab:eval5-oracle}
\begin{tabular}{l rr rr rr}
\toprule
& \multicolumn{2}{c}{All days} & \multicolumn{2}{c}{Holidays only} & \multicolumn{2}{c}{Non-holidays only} \\
\cmidrule(lr){2-3} \cmidrule(lr){4-5} \cmidrule(lr){6-7}
Model & MASE & SQL & MASE & SQL & MASE & SQL \\
\midrule
Chronos-2   & 0.2994 & 0.2348 & 0.5473 & 0.4363 & 0.2902 & 0.2274 \\
TabPFN-TS-3 & \textbf{0.0277} & \textbf{0.0208} & \textbf{0.0267} & \textbf{0.0194} & \textbf{0.0277} & \textbf{0.0208} \\
TiRex-2     & 0.3033 & 0.2480 & 0.2631 & 0.2121 & 0.3048 & 0.2493 \\
TimesFM-3   & 0.0824 & 0.0645 & 0.0774 & 0.0612 & 0.0826 & 0.0646 \\
\bottomrule
\end{tabular}
\end{table}
Table~\ref{tab:eval5-oracle} shows a setting in which the ground-truth target is provided as a known covariate. The results therefore represent the maximum performance each model can reach by exploiting covariate information. TabPFN-TS-3 and TimesFM-3 nearly reproduce the target. Chronos-2 and TiRex-2 remain at a MASE of about 0.30 despite perfect information, and Chronos-2 performs notably worse on holidays. The benefit of holiday covariates is thus limited not only by the information they contain but also by each model's ability to use it.
\clearpage
\subsubsection{Overall Results with Context Length 2048}
\label{sec:App_Results2048}
\begin{table}[H]
\centering
\small
\caption{Overall results across all datasets at context 2048, split by day type (mean over 9 datasets).}
\label{tab:eval5-overall-split-ctx2048}
\resizebox{\textwidth}{!}{%
\begin{tabular}{ll rr rr rr}
\toprule
& & \multicolumn{2}{c}{All days} & \multicolumn{2}{c}{Holidays only} & \multicolumn{2}{c}{Non-holidays only} \\
\cmidrule(lr){3-4} \cmidrule(lr){5-6} \cmidrule(lr){7-8}
Model & Encoding & MASE & SQL & MASE & SQL & MASE & SQL \\
\midrule
\multirow{2}{*}{Baselines}
 & seasonal-naive (168h)    & 1.0437 & 0.9788 & 3.0978 & 2.7752 & 0.9703 & 0.9147 \\
 & $T_{h,W}(H=Sun)_{naive}$    & \underline{0.9804} & \underline{0.8290} & \underline{1.4629} & \underline{1.2491} & \underline{0.9617} & \underline{0.8127} \\
\midrule
\multirow{8}{*}{Chronos-2}
 & \encNone            & 0.6970 & 0.5607 & 2.3688 & 2.0581 & 0.6373 & 0.5072 \\
 & \encOrdinal         & 0.6883 & 0.5507 & 1.8060 & 1.5326 & 0.6483 & 0.5156 \\
 & \encOnehot          & 0.7198 & 0.5788 & 2.2567 & 1.9568 & 0.6646 & 0.5293 \\
 & \encCyclic          & 0.6965 & 0.5579 & 1.8765 & 1.6026 & 0.6544 & 0.5206 \\
 & \encTcombo          & \textbf{\underline{0.6572}} & \textbf{\underline{0.5229}} & \underline{1.5185} & \underline{1.1842} & \textbf{\underline{0.6256}} & \textbf{\underline{0.4987}} \\
 & \encTholEight       & 0.6588 & 0.5242 & 1.5406 & 1.2036 & 0.6262 & 0.4991 \\
 & \encTholAll         & 0.6743 & 0.5367 & 1.9640 & 1.5397 & 0.6273 & 0.5001 \\
 & \encTholPartial     & 0.6636 & 0.5287 & 1.6463 & 1.3022 & 0.6275 & 0.5003 \\
\midrule
\multirow{8}{*}{TabPFN-TS-3}
 & \encNone            & 0.8855 & 0.7096 & 2.6146 & 2.1610 & 0.8236 & 0.6577 \\
 & \encOrdinal         & 0.7570 & 0.6074 & 1.6378 & 1.3859 & 0.7241 & 0.5783 \\
 & \encOnehot          & 0.7609 & 0.6094 & 1.8312 & 1.5212 & 0.7214 & 0.5757 \\
 & \encCyclic          & 0.7666 & 0.6152 & 1.6463 & 1.3937 & 0.7337 & 0.5861 \\
 & \encTcombo          & \underline{0.6929} & \underline{0.5572} & \underline{1.2646} & \underline{1.0609} & \underline{0.6710} & \underline{0.5377} \\
 & \encTholEight       & 0.7077 & 0.5719 & 1.6273 & 1.4201 & 0.6730 & 0.5398 \\
 & \encTholAll         & 0.7407 & 0.6036 & 2.5777 & 2.3122 & 0.6736 & 0.5410 \\
 & \encTholPartial     & 0.7100 & 0.5727 & 1.7297 & 1.4741 & 0.6720 & 0.5391 \\
\midrule
\multirow{8}{*}{TiRex-2}
 & \encNone            & 0.8008 & 0.6447 & 2.2826 & 1.9248 & 0.7485 & 0.5996 \\
 & \encOrdinal         & 0.8278 & 0.6618 & 2.1759 & 1.8019 & 0.7803 & 0.6218 \\
 & \encOnehot          & 0.8062 & 0.6515 & 2.3829 & 2.0031 & 0.7503 & 0.6037 \\
 & \encCyclic          & 0.8023 & 0.6463 & 2.3449 & 1.9797 & \underline{0.7476} & \underline{0.5992} \\
 & \encTcombo          & \underline{0.7754} & \underline{0.6201} & \underline{1.2231} & \underline{1.0103} & 0.7582 & 0.6050 \\
 & \encTholEight       & 0.7861 & 0.6303 & 1.5636 & 1.3235 & 0.7564 & 0.6037 \\
 & \encTholAll         & 0.8077 & 0.6495 & 2.1241 & 1.8312 & 0.7594 & 0.6059 \\
 & \encTholPartial     & 0.8387 & 0.6717 & 2.3494 & 2.0074 & 0.7842 & 0.6236 \\
\midrule
\multirow{8}{*}{TimesFM-3}
 & \encNone            & 0.6861 & 0.5540 & 2.3042 & 1.9828 & 0.6288 & 0.5033 \\
 & \encOrdinal         & 0.6727 & 0.5411 & 1.9447 & 1.6532 & 0.6273 & 0.5013 \\
 & \encOnehot          & 0.6815 & 0.5489 & 2.1395 & 1.8313 & 0.6293 & 0.5030 \\
 & \encCyclic          & 0.6721 & 0.5410 & 1.9632 & 1.6607 & \underline{0.6260} & \underline{0.5009} \\
 & \encTcombo          & \underline{0.6604} & \underline{0.5294} & \textbf{\underline{1.1522}} & \textbf{\underline{0.9532}} & 0.6416 & 0.5132 \\
 & \encTholEight       & 0.6737 & 0.5411 & 1.4471 & 1.2205 & 0.6445 & 0.5154 \\
 & \encTholAll         & 0.6894 & 0.5551 & 1.8623 & 1.5684 & 0.6465 & 0.5180 \\
 & \encTholPartial     & 0.6958 & 0.5571 & 1.6234 & 1.3394 & 0.6614 & 0.5281 \\
\bottomrule
\end{tabular}}
\end{table}

\clearpage

\end{document}